\documentclass[acmsmall,screen,nonacm]{acmart}

\AtBeginDocument{}

\usepackage{ifthen}
\usepackage{hyperref}
\usepackage{amsmath}

\usepackage{amssymb}
\usepackage{mdframed}
\usepackage{caption}
\usepackage{adjustbox}
\usepackage[flushleft]{threeparttable}

\usepackage{booktabs}   \usepackage[labelformat=simple]{subcaption} 

\usepackage[section]{placeins}
\usepackage{multirow}
\usepackage{multicol}

\usepackage{crossreftools}
\usepackage{graphicx}
\usepackage{wrapfig}

\newcommand{\subsubsubsection}[1]{\noindent\textbf{\emph{#1}}\enspace}

\usepackage{xcolor}
\definecolor{darkgreen}{RGB}{25, 116, 25}
\definecolor{persianblue}{rgb}{0.11, 0.22, 0.73}
\definecolor{lapislazuli}{rgb}{0.15, 0.38, 0.61}
\definecolor{light-gray}{gray}{0.9}
\definecolor{light-pink}{rgb}{0.858, 0.188, 0.478}
\definecolor{maroon}{rgb}{0.5, 0.0, 0.0}
\colorlet{pink}{red!40}
\colorlet{cyan}{cyan!60}
\colorlet{gray}{gray!60}
 \usepackage{prettyref}
\newcommand{\pref}{\prettyref}
\newrefformat{thm}{Theorem~\ref{#1}}
\newrefformat{cor}{Corollary~\ref{#1}}
\newrefformat{lem}{Lemma~\ref{#1}}
\newrefformat{cha}{Chapter~\ref{#1}}
\newrefformat{sec}{\S\!~\ref{#1}}
\newrefformat{app}{Appendix~\ref{#1}}
\newrefformat{tab}{Table~\ref{#1}}
\newrefformat{fig}{Figure~\ref{#1}}
\newrefformat{alg}{Algorithm~\ref{#1}}
\newrefformat{exa}{Example~\ref{#1}}
\newrefformat{def}{Definition~\ref{#1}}
\newrefformat{li}{Line~\ref{#1}}
\newrefformat{eq}{Equation~\ref{#1}}
\newrefformat{cond}{Condition~\ref{#1}}
\newrefformat{step}{Step~\ref{#1}} \usepackage[inline]{enumitem}
\newlist{rqs}{enumerate}{1}
\setlist[rqs,1]{label={\bfseries RQ\arabic*},align=left,leftmargin=*}

\newlist{ps}{enumerate}{1}
\setlist[ps,1]{label={\bfseries P\arabic*},align=left,leftmargin=*}

\newlist{fs}{enumerate}{1}
\setlist[fs,1]{label={\bfseries F\arabic*},align=left,leftmargin=*}

\newlist{cs}{enumerate}{1}
\setlist[cs,1]{
    label={\textup{\bfseries C\arabic*}},
    ref={\bfseries C\arabic*},
    align=left,leftmargin=*
}

\newlist{rs}{enumerate}{1}
\setlist[rs,1]{
    label={\textup{\bfseries AC\arabic*}},
    ref={\bfseries AC\arabic*},
    align=left,leftmargin=*
}

\newlist{steps}{enumerate}{1}
\setlist[steps,1]{
    label={\bfseries S\arabic*},
    align=left,
    leftmargin=*,
    wide=0pt
}
 \newcommand{\eqdef}{\buildrel \mbox{\tiny\textrm{def}} \over =}

\usepackage{mathtools}

\DeclarePairedDelimiterX{\norm}[1]{\lVert}{\rVert}{#1}

\DeclareMathOperator*{\argmax}{argmax}
\DeclareMathOperator*{\argmin}{argmin}

\renewcommand{\hat}[1]{\widehat{#1}}

\newcommand{\sy}[1]{{\color{persianblue}\hat{\boldsymbol{#1}}}}

\newcommand{\Ly}[2]{#1^{(#2)}}
\newcommand{\idx}[1]{\prescript{}{#1}{}}

\usepackage{array,mathtools}
\newcolumntype{M}[1]{>{\hbox to #1\bgroup\hss$}l<{$\egroup}}

\makeatletter
\newcommand\@brcolwidth{0.67em}

\def\@brarray[#1]{\array{r*\c@MaxMatrixCols {M{#1}}}}
\makeatother

\newcommand{\vbmatrix}[2]{
    \setcounter{MaxMatrixCols}{20}
    \setlength{\arraycolsep}{#1}
    \begin{bmatrix}
        #2
    \end{bmatrix}
}

\newcommand{\vbscalematrix}[3]{
    \scalebox{#1}{$\vbmatrix{#2}{#3}$}
}

\newcommand{\vbinlinematrixScale}{1}
\newcommand{\vbinlinematrix}[2]{
    \vbscalematrix{\vbinlinematrixScale}{#1}{#2}
}
\newcommand{\bpoint}[1]{\vbmatrix{0pt}{#1}}

\newcommand{\myvmatrix}[2]{
    \setcounter{MaxMatrixCols}{20}
    \setlength{\arraycolsep}{#1}
    \begin{matrix}
        #2
    \end{matrix}
}

\newcommand{\buildVector}[1]{
    \left\langle
        #1
    \right\rangle
}
\newcommand{\buildVectorTwo}[2]{
    \left\langle
        \;#1\;\,\big\vert\;\,#2\;
    \right\rangle
}

\newcommand{\SliceDNN}[3]{#1\ly{{#2}:{#3}}}

\newcommand{\tyReal}{\mathbb{R}}

\newcommand{\stheta}{\sy{\theta}}

\newcommand{\bigargs}[1]{\bigl({#1}\bigr)}

\newcommand{\eval}[1]{\llbracket{#1}\rrbracket}

\newcommand{\withRef}[1]{\bigl[{#1}\bigr]}
\newcommand{\condsy}[1]{\widetilde{#1}}

\newcommand{\relu}{\texttt{ReLU}}
\newcommand{\scalarReLU}{\textrm{relu}}
\newcommand{\identity}{\texttt{Id}}

\newcommand{\scalarHardswish}{\textrm{hardswish}}
\newcommand{\hardswish}{\texttt{Hardswish}}

\newcommand{\ConcreteExec}[1]{#1}

\newcommand{\convexhull}{\texttt{ConvexHull}}

\newcommand{\toolname}{\textsc{APRNN}}

\newcommand{\PolyRepair}{\texttt{VPolytopeRepair}}
\newcommand{\RefKW}{\texttt{ref}}
\newcommand{\OriginalKW}{\texttt{og}}
\newcommand{\PreKW}{\texttt{pre}}
\newcommand{\foo}{\texttt{Shift\&Assert}}

\newcommand{\zip}{\texttt{zip}}
\newcommand{\spec}{\Psi}

\newcommand{\Minimize}[2]{\texttt{Minimize}\:\,{#1}\;\texttt{Subject\!\! To}\;\,{#2}}
\newcommand{\vecNorm}[2]{\big\lVert{#1}\big\rVert_{#2}}

\newcommand{\Update}{\texttt{Update}}

\newcommand{\append}{\texttt{append}}

\newcommand{\len}{\texttt{len}}

\newcommand{\centroid}{\texttt{calc\_ref}}

\newcommand{\polytope}{P}

\newcommand{\polytopeSet}{\mathcal{\polytope}}

\newcommand{\point}{X}
\newcommand{\spoint}{\sy{\point}}
\newcommand{\pointAlt}{Y}
\newcommand{\pointSet}{\mathcal{\point}}

\newcommand{\ly}[1]{^{(#1)}}

\newcommand{\expr}{\mathcal{E}}
\newcommand{\formula}{\mathcal{F}}
\newcommand{\vars}{{\sy V}}
\newcommand{\accuracy}{\text{accuracy}}

\newcommand{\ite}{\mathtt{ite}}
 \newcommand{\sW}{\sy{W}}
\newcommand{\sB}{\sy{B}}

\newcommand{\locallylinear}{locally linear}

\newcommand{\dnn}{\mathcal{N}}

\newcommand{\OverviewDNN}{\dnn_1}
\newcommand{\OverviewDNNPointwiseRepaired}{\dnn_2}
\newcommand{\OverviewDNNPolytopeWrong}{\dnn_3}
\newcommand{\OverviewDNNShifted}{\dnn_4}
\newcommand{\OverviewDNNPolytopeRepaired}{\dnn_5}
 
\usepackage{tikz}
\usetikzlibrary{positioning,snakes,arrows,shapes}
\usepackage{pgfplots}

\usetikzlibrary{fit,calc}
\newcommand*{\tikzmk}[1]{\tikz[remember picture,overlay,] \node (#1) {};\ignorespaces}
\newcommand{\boxit}[2]{\tikz[remember picture,overlay]{\node[yshift=3pt,xshift=3pt,fill=#1,opacity=.15,fit={(A)($(B)+(#2\linewidth,.8\baselineskip)$)}] {};}\ignorespaces}

\newcommand{\VarParam}[1]{\textbf{\small{#1}}}
\newcommand{\NewParam}[1]{\textbf{\small{\colorbox{green!70}{#1}}}}
\newcommand{\ParamDNNFigureScale}{1}
\newcommand{\ParamDNNPlotScale}{0.5}
\newcommand{\ParamDNNFigureWidth}{0.55\linewidth}
\newcommand{\ParamDNNPlotWidth}{0.4\linewidth}
\newcommand{\ParamDNNFigureSep}{\hspace{-6ex}}

\newcommand{\neworrenewcommand}[1]{\providecommand{#1}{}\renewcommand{#1}}

\newcommand{\overviewdnnfig}[9]{
\neworrenewcommand{\ooverviewdnn}[1]{
    \tikzstyle{dnnnode} = [circle, draw, inner sep=0pt,minimum size=3.5ex]
    \tikzstyle{relunode} = [circle, draw, double, double distance=1pt, inner sep=0pt,minimum size=3.5ex]
    \draw node[dnnnode] (x00) at (0, 0) {\footnotesize $\Ly{X}{0}_0$};
    \draw node[relunode,label={[label distance=-0.5ex]above:{\small $#4$}}] (x10) at (1.8, 1.2) {\footnotesize $\Ly{X}{1}_0$};
    \draw node[relunode,label={[label distance=-0.5ex]above:{\small $#5$}}] (x11) at (1.8, 0) {\footnotesize $\Ly{X}{1}_1$};
    \draw node[relunode,label={[label distance=-0.5ex]above:{\small $#6$}}] (x12) at (1.8, -1.2) {\footnotesize $\Ly{X}{1}_2$};
    \draw node[dnnnode,label={[label distance=-0.5ex]above:{\small $##1$}}] (x20) at (3.6, 0) {\footnotesize $\Ly{X}{2}_0$};

    \draw[->] (x00) -- (x10) node[midway,above] {\small $#1$};
    \draw[->] (x00) -- (x11) node[midway,above] {\small $#2$};
    \draw[->] (x00) -- (x12) node[midway,below] {\small $#3$};
    \draw[->] (x10) -- (x20) node[midway,above] {\small $#7$};
    \draw[->] (x11) -- (x20) node[midway,above] {\small $#8$};
    \draw[->] (x12) -- (x20) node[midway,below] {\small $#9$};
}
\ooverviewdnn
}

\newcommand{\overviewdnnplotrich}[8]{
    \begin{axis}[
ymin=-0.8,ymax=1,
        xlabel={Input $\Ly{X}{0}_0$},
        ylabel={Output $\Ly{X}{2}_0$},
    font=\huge]
\addplot[ultra thick,red,domain=#1:#3,samples=2] {#2};
        \addplot[ultra thick,blue,domain=#3:#5,samples=2] {#4};
        \addplot[ultra thick,green,domain=#5:#7,samples=2] {#6};
        \draw[fill=red,opacity=.7] (axis cs:#1,-2) rectangle (axis cs:#3,-0.75);
        \draw[fill=blue,opacity=.7] (axis cs:#3,-2) rectangle (axis cs:#5,-0.75);
        \draw[fill=green,opacity=.7] (axis cs:#5,-2) rectangle (axis cs:#7,-0.75);
#8
    \end{axis}
}
 \usepackage[ruled,vlined,linesnumbered]{algorithm2e}
\SetKwProg{Def}{def}{:}{end}

\SetKw{returnKw}{return}

\SetAlgoNlRelativeSize{-1}

\SetCommentSty{CommentFont}
\SetKwComment{Comment}{$\triangleright$ }{}

\newcommand{\Omit}[1]{}

\newcommand{\HightlightDef}[4]{
    \tikzmk{A}
    \Def{#3}{\tikzmk{B} \boxit{#1}{#2}
        #4
    }
}

\newcommand{\DefHelperFunc}[2]{
    \HightlightDef{cyan}{.8}{#1}{#2}
}
\newcommand{\DefAlgBody}[1]{
#1
}

\usepackage{listings}

 \usepackage{amsthm}
\newenvironment{proofsketch}{\begin{proof}[Proof Sketch]}{\end{proof}}

\makeatletter
\newtheorem*{rep@theorem}{\rep@title}
\newcommand{\newreptheorem}[2]{\newenvironment{rep#1}[1]{\def\rep@title{#2 \ref{##1}}\begin{rep@theorem}}{\end{rep@theorem}}}
\makeatother

\newreptheorem{theorem}{Theorem}

\newcommand{\defProof}[1]{\crtcrossreflabel{}[#1]}
\newcommand{\refProof}[2]{The \hyperref[#2]{full proof of \pref{#1}} is in~\pref{app:Appendix-Proof}~(page~\pageref{#2}).}
 
\newcommand\target{arxiv}
\newcommand\onlyfor[3]{\ifthenelse{\equal{#1}{\target}}{#2}{#3}}

\newcommand{\pldivspace}[1]{\onlyfor{pldi}{\vspace{#1}}{}}

\newcommand\inMainText{true}
\newcommand{\labelInMainText}[1]{\ifthenelse{\equal{true}{\inMainText}}{\label{#1}}{}}
 
\onlyfor{pldi}{
\setcopyright{rightsretained}
\acmPrice{}
\acmDOI{10.1145/3591238}
\acmYear{2023}
\copyrightyear{2023}
\acmSubmissionID{pldi23main-p113-p}
\acmJournal{PACMPL}
\acmVolume{7}
\acmNumber{PLDI}
\acmArticle{124}
\acmMonth{6}
\received{2022-11-10}
\received[accepted]{2023-03-31}
}{
\setcopyright{none}
\settopmatter{printacmref=false}
\renewcommand\footnotetextcopyrightpermission[1]{}
\fancyfoot{}
}

\begin{document}

\title[Architecture-Preserving Provable Repair of DNNs]{Architecture-Preserving Provable Repair of Deep Neural Networks}

\author{Zhe Tao}
\affiliation{
    \department{Computer Science}
    \institution{University of California, Davis}
    \city{Davis}
    \state{California}
    \postcode{95616}
    \country{U.S.A.}
}
\email{zhetao@ucdavis.edu}
\orcid{0000-0002-4047-699X}
\author{Stephanie Nawas}
\affiliation{
    \department{Computer Science}
    \institution{University of California, Davis}
    \city{Davis}
    \state{California}
    \postcode{95616}
    \country{U.S.A.}
}
\email{snawas@ucdavis.edu}
\orcid{0009-0003-1506-2853}
\author{Jacqueline Mitchell}
\affiliation{
    \department{Computer Science}
    \institution{University of California, Davis}
    \city{Davis}
    \state{California}
    \postcode{95616}
    \country{U.S.A.}
}
\email{jlmitchell@ucdavis.edu}
\orcid{0009-0007-8593-2972}
\author{Aditya V. Thakur}
\affiliation{
  \position{Associate Professor}
  \department{Computer Science}
  \institution{University of California, Davis}
  \city{Davis}
  \state{California}
  \postcode{95616}
  \country{U.S.A.}
}
\email{avthakur@ucdavis.edu}
\orcid{0000-0003-3166-1517}

\begin{abstract}
Deep neural networks (DNNs) are becoming increasingly important components
of software, and are considered the state-of-the-art solution for a number
of problems, such as image recognition. However, DNNs are far from
infallible, and incorrect behavior of DNNs can have disastrous real-world
consequences. This paper addresses the problem of architecture-preserving
V-polytope provable repair of DNNs.
A V-polytope defines a convex bounded polytope using its vertex representation.
V-polytope provable repair guarantees that the repaired DNN
satisfies the given specification on the infinite set of points in the given V-polytope.
An architecture-preserving repair only modifies the parameters of the DNN, without
modifying its architecture. The repair has the flexibility to
modify multiple layers of the DNN, and runs in polynomial time.
It~supports DNNs with activation functions that have some linear pieces,
as well as fully-connected, convolutional, pooling and residual layers.
To the best our knowledge, this is the first provable repair approach that
has all of these features.
We implement our approach in a tool called APRNN. Using
MNIST, ImageNet, and ACAS Xu DNNs, we show that
it has better efficiency, scalability, and generalization
compared to PRDNN and REASSURE, prior provable repair methods that are
not architecture preserving.
\end{abstract}



\maketitle

\section{Introduction}
\label{sec:Introduction}

Deep Neural Networks~(DNNs)~\cite{Goodfellow-et-al-2016},
which learn by generalizing from a finite set of examples,
are increasingly becoming critical
components of software.
They have emerged as the state-of-the-art approach for
solving problems such as image recognition~\cite{DBLP:journals/corr/IandolaMAHDK16,DBLP:conf/nips/KrizhevskySH12,DBLP:conf/iclr/DosovitskiyB0WZ21} and natural-language
processing~\cite{DBLP:conf/naacl/DevlinCLT19,DBLP:journals/corr/RoBERTa,DBLP:journals/corr/abs-2203-02155}. They are being applied in diverse problem domains such as
scientific computing~\cite{stevens2020ai}, medicine~\cite{Goodfellow-et-al-2016,soenksenDLlesion},
and economics~\cite{MALIAR202176}. Importantly, they are being deployed in
safety-critical applications such as aircraft
controllers~\cite{DBLP:journals/corr/abs-1810-04240}. However, DNNs are far from
infallible, and incorrect behavior of DNNs can have disastrous real-world
consequences.
Thus, there has been
extensive research on formal methods for
the analysis and verification of
DNNs~\cite{DBLP:conf/cav/KatzBDJK17,DBLP:conf/fmcad/LahavK21,10.1145/3498704,DBLP:conf/iclr/XuZ0WJLH21,DBLP:conf/iclr/FerrariMJV22,DBLP:conf/nips/SinghGMPV18}.
In contrast,
this paper addresses the next step of this process: what to do once faulty
behavior is identified.

\newcommand{\imagenetcorrect}{dragonfly}
\newcommand{\imagenetincorrect}{manhole~cover}
\newcommand{\mnistincorrect}{``5''}
\newcommand{\mnistincorrecttwo}{``8''}

Consider the scenario in which we are given a (finite) set of inputs for which a
DNN gives an incorrect result. For instance,
\pref{fig:Intro-ImageNet-NAE-Repair} shows an image of
\imagenetcorrect{} that an ImageNet DNN misclassifies as \imagenetincorrect{};
\pref{fig:Intro-MNIST-Repair} shows an image of handwritten ``7''
that is misclassified as \mnistincorrect{}. \emph{Pointwise repair of DNNs}
takes as input a finite set of inputs along with the specification of their
corresponding behavior, and returns a repaired DNN satisfying this repair
specification.

\clearpage
A~DNN repair approach should satisfy the following properties:
\begin{figure}[t]
  \centering
\begin{minipage}[t]{0.233\textwidth}
      \centering
      \includegraphics[width=0.95\textwidth]{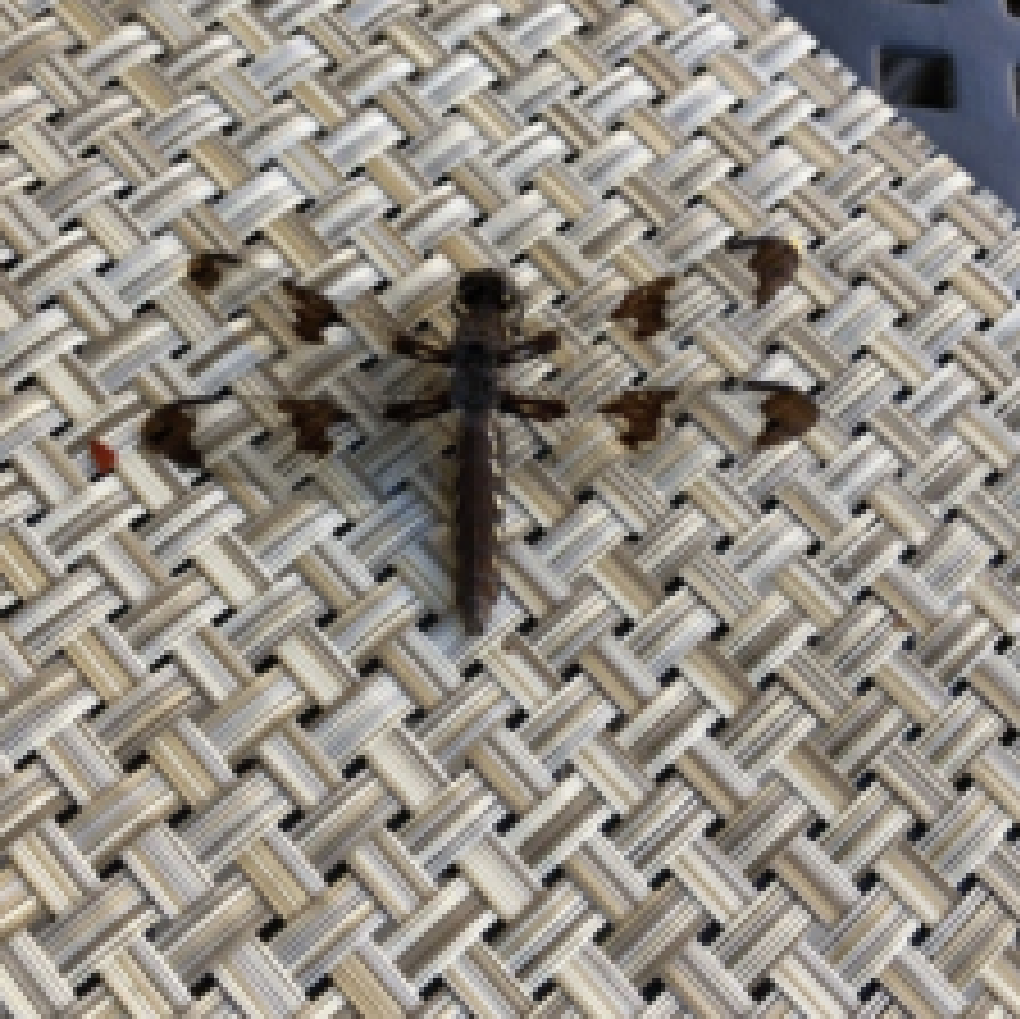}
      \caption{
Dragonfly
misclassified as \imagenetincorrect.}
      \label{fig:Intro-ImageNet-NAE-Repair}
  \end{minipage}
  \hfill
  \begin{minipage}[t]{0.233\textwidth}
      \centering
      \includegraphics[width=0.95\textwidth]{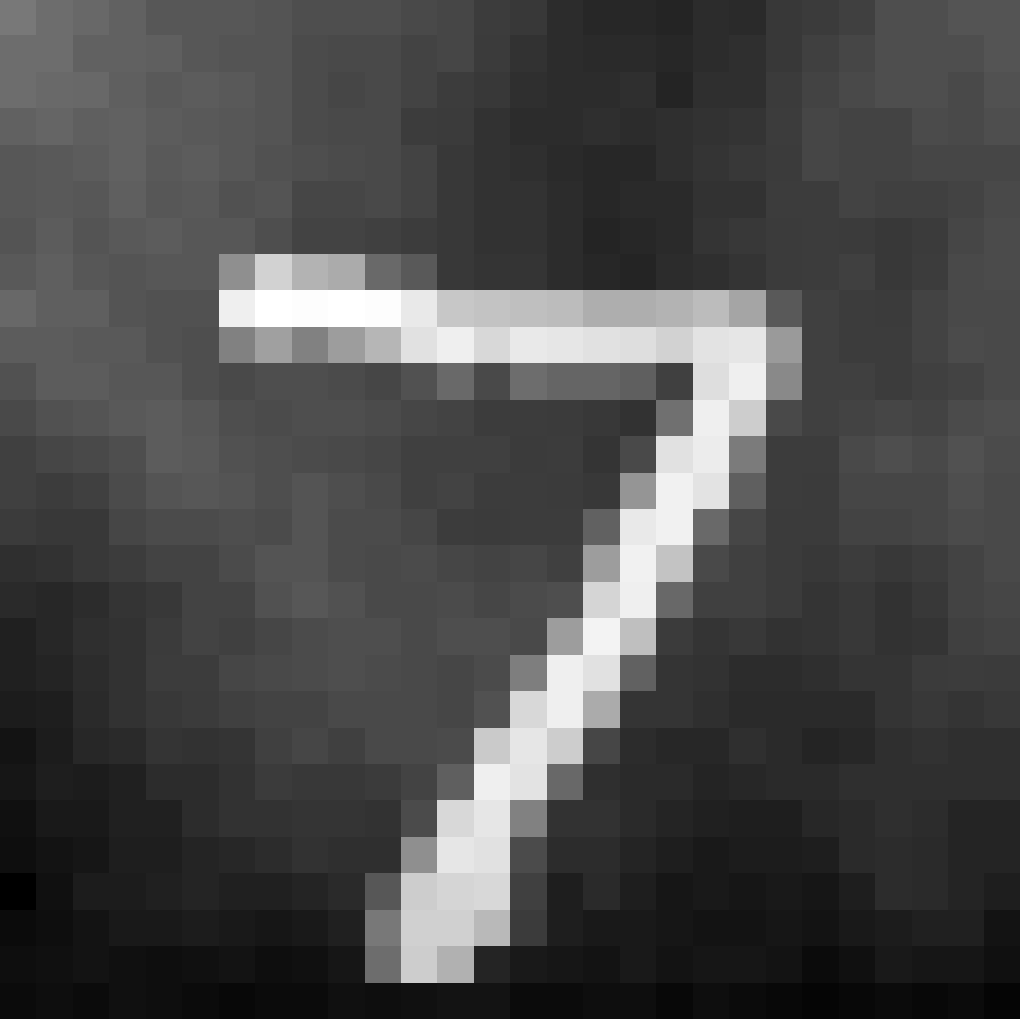}
      \caption{Digit ``7'' misclassified as digit \mnistincorrect{}.}
\label{fig:Intro-MNIST-Repair}
  \end{minipage}
  \hfill
  \begin{minipage}[t]{0.233\textwidth}
      \centering
      \includegraphics[width=0.95\textwidth]{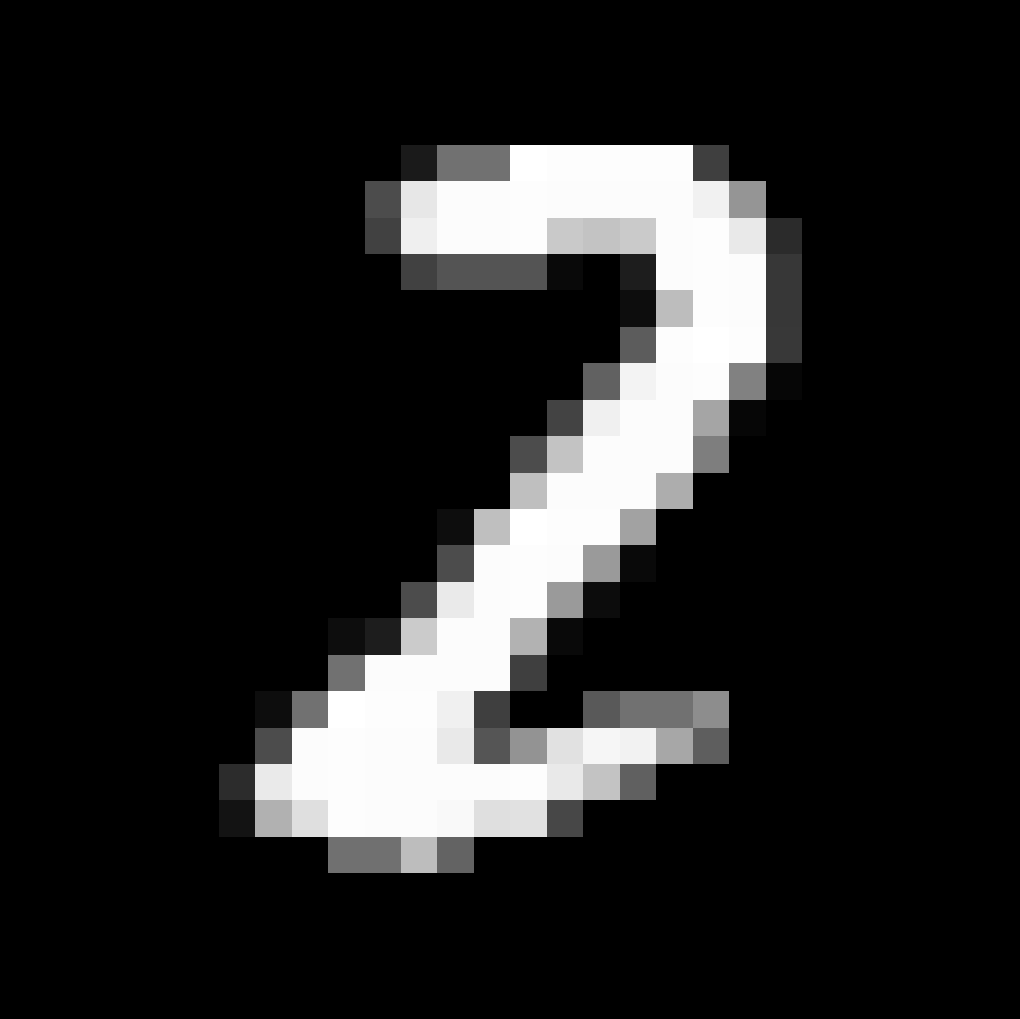}
      \caption{Correctly classified digit ``2''.}
      \label{fig:Intro-MNIST-Drawdown}
  \end{minipage}
  \hfill
  \begin{minipage}[t]{0.233\textwidth}
      \centering
      \includegraphics[width=0.95\textwidth]{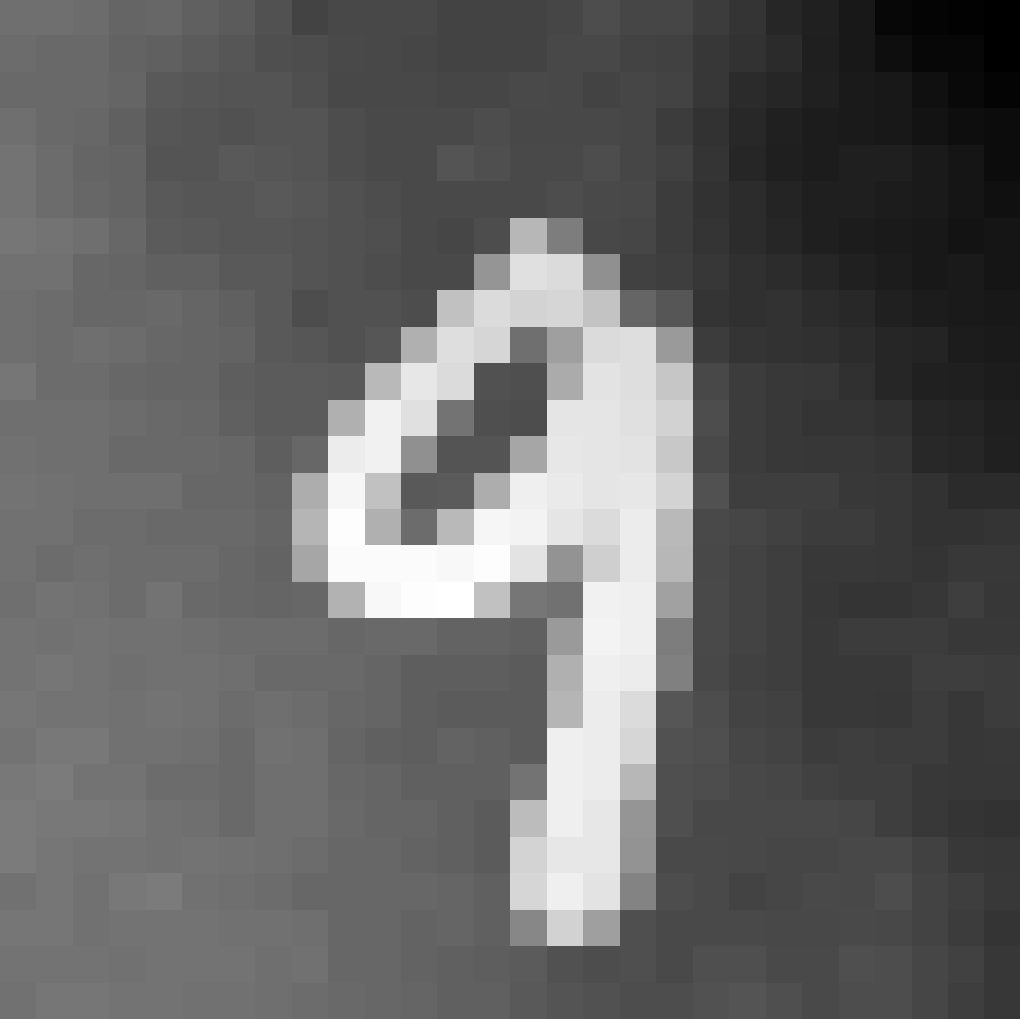}
      \caption{Digit ``9'' misclassified as digit \mnistincorrecttwo{}.}
\label{fig:Intro-MNIST-Generalization}
  \end{minipage}
\end{figure}

\begin{ps}
  \item \label{item:P1} \emph{Low repair drawdown}. Repairs to the DNN should
  not cause significant \emph{drawdown}, which is when the DNN ``forgets'' its
  previous correct behavior on other inputs. For example, suppose a DNN
  correctly classifies the image in~\pref{fig:Intro-MNIST-Drawdown} as a ``2''.
  Repairing the DNN to correctly classify the image in
  \pref{fig:Intro-MNIST-Repair} should not result in the incorrect classification of
  the image in~\pref{fig:Intro-MNIST-Drawdown}.

  \item \label{item:P2} \emph{High repair generalization}.
  Repairing the DNN on one set of inputs should result in the repair of a similar set of inputs.
  For example, repairing the DNN for the fog-corrupted image in~\pref{fig:Intro-MNIST-Repair}
  should also fix the classification for the fog-corrupted image in~\pref{fig:Intro-MNIST-Generalization}.
\item \label{item:P3}\label{property:P3} \emph{Efficacy.} The repair should
  \emph{guarantee} that the changes to the DNN result in correct outputs for the
  inputs according to the repair specification. For example, a repair of a faulty
  input to an aircraft collision avoidance DNN will guarantee that the output of
  the repaired DNN satisfies the specified safety properties.
  \item \label{item:P4} \emph{Efficient.} The repair process should be efficient and scale
  to real-world networks.
 \end{ps}

One approach to pointwise repair could be to \emph{retrain} the
DNN by adding
the faulty inputs with their corrected output to the original training set.
Unfortunately, this method is not only
incredibly time-consuming for large DNNs, but also infeasible in cases where access
to the original training set is restricted. Additionally, retraining the DNN does not guarantee
correctness on the faulty inputs added to the new training set.
Another pointwise repair approach is to \emph{fine tune}
the DNN on the identified faulty inputs using gradient descent. This strategy, however,
significantly increases the risk of high drawdown~\cite{DBLP:conf/aaai/KemkerMAHK18}.
A less desirable option is to simply accept the faulty behavior of the DNN as
a natural consequence of deep learning. This can be incredibly dangerous in
safety-critical applications of DNNs, and the faulty
behavior can even be exploited by bad actors if left unpatched.

There are many scenarios in which we might want to specify how the DNN should
behave for an \emph{infinite} set of inputs. One approach to represent such an
infinite set of points is via \emph{V-polytopes}, where a vertex representation
of a convex bounded polytope is used. \emph{V-polytope repair of DNNs} takes as input a
finite set of V-polytopes and the specification of the corresponding behavior
constraining the (infinite) set of points in each of the V-polytopes. For
example, a V-polytope repair specification can be used to express the constraint that certain
input regions of an aircraft collision avoidance DNN should satisfy some safety
properties. Retraining and fine tuning only take finite sets of input points, so
they cannot be classified as V-polytope repair methods. On the other hand,
V-polytope repair can be reduced to pointwise repair if the V-polytope input is
defined as a single vertex.

Recently, there have been a number of approaches addressing \emph{Provable
Repair} of
DNNs~\cite{SRDM2019,DBLP:conf/lpar/GoldbergerKAK20,DBLP:conf/pldi/SotoudehT21,DBLP:conf/iclr/Fu22}.
Provable Repair ensures that the resulting repaired DNN is guaranteed to satisfy
the given repair specification (property~\ref{property:P3}).

In this paper, we present a new Provable Repair approach that, to the best of our knowledge,
is the first method that includes \emph{all} of the following features:
\begin{fs}
  \item \label{item:F1} \emph{Architecture-Preserving}: Our method preserves the architecture of the DNN
  throughout the repair process. The repaired DNN
  will have an identical structure to the DNN pre-repair.
\item \label{item:F2} \emph{Arbitrary V-Polytopes}: Our method supports the repair of arbitrary
    V-polytopes.
  \item \label{item:F3}\label{feature:3} \emph{Multi-Layer Repair}: Our approach supports the
    modification of the weights of multiple layers of a DNN rather than
    restricting edits to a single layer.
\item \label{item:F4}\label{feature:4} \emph{Non-Piecewise-Linear Activation Function Support}: Our approach is applicable to DNNs that use
    activation functions that have some linear pieces such as ReLU and Hardswish.
\item \label{item:F5} \emph{Efficiency}: Our repair algorithm runs in polynomial time.
  \item \label{item:F6} \emph{Scalability}: Our method repairs large real-world
  DNNs, including ImageNet DNNs like VGG19 and ResNet152,
  and handles complex global properties, such as for the ACAS Xu networks.

\end{fs}

\noindent \textbf{Organization.}
\pref{sec:Preliminaries} presents key definitions and notation.
\pref{sec:Overview} illustrates the key insights of
our architecture-preserving provable V-polytope repair method on a simple DNN.
\pref{sec:VPolytopeRepair} describes our algorithm in detail.
\pref{sec:Comparison} presents a qualitative comparison between our approach and prior provable-repair techniques
such as PRDNN~\cite{DBLP:conf/pldi/SotoudehT21} and REASSURE~\cite{DBLP:conf/iclr/Fu22}.
\pref{sec:Implementation} outlines the implementation of our approach in a tool called
\toolname{}.
We evaluate \toolname{} using MNIST, ImageNet, and ACAS~Xu networks and compare against PRDNN, REASSURE~\cite{REASSURE}, as well as lookup-based approaches.
\pref{sec:ExperimentalEvalation} shows
that \toolname{} outperforms these prior techniques in terms of efficiency,
scalability, and generalization,
all while preserving the original network architecture.
\pref{sec:RelatedWork} describes related work and \pref{sec:Conclusion} concludes.

\section{Preliminaries}
\label{sec:Preliminaries}

\begin{figure}[t]
    \begin{subfigure}{\ParamDNNFigureWidth}
        \centering
        \begin{tikzpicture}[scale=\ParamDNNFigureScale, transform shape]
    \overviewdnnfig
        {\VarParam{-1}}{\VarParam{1}}{\VarParam{0.5}}
        {\VarParam{+0}}{\VarParam{-2}}{\VarParam{+0}}
        {0.5}{-0.5}{1}
        {\VarParam{-0.5}}
\end{tikzpicture}
         \vspace{1ex}
        \caption{DNN $\OverviewDNN$.}
        \label{fig:Overview-DNN}
    \end{subfigure}
    \ParamDNNFigureSep{}
    \begin{subfigure}{\ParamDNNPlotWidth}
        \centering\begin{tikzpicture}[scale=\ParamDNNPlotScale,
    refdashed/.style={dashed,black,line width=0.1em},
        point/.style={circle,draw,fill=black,minimum size=0.1em,scale=0.6},
        relu/.style={blue,line width=0.3em},
        linearized/.style={dashed,red,line width=0.3em}]
    \overviewdnnplotrich
        {-2}{(-0.5)*x-0.5}
        {0}{(0.5)*x-0.5}
        {2}{0.5}
        {4}{
    \node[point,red] at (axis cs: -1.5, 0.25) (p1) {};
    \node[above right = 0em and 0em of p1, scale=0.64] {$(-1.5, 0.25)$};
\node[point,red] at (axis cs: -0.5, -0.25) (p2) {};
    \node[below left = 0em and -.6em of p2, scale=0.64] {$(-0.5, -0.25)$};
\draw[refdashed] (axis cs: -2.55, -0.1) -- (axis cs: 4.55, -0.1);
    \draw[refdashed] (axis cs: -2.55,  0.1) -- (axis cs: 4.55,  0.1);
    }
\end{tikzpicture}
         \vspace{-0.8ex}
        \caption{Input-output plot of $\dnn_1$.}
        \label{fig:Overview-DNN-Plot}
    \end{subfigure}
    \vspace{-1ex}
    \caption{DNN $\OverviewDNN$ and its input-output plot. The nodes $\Ly{X}{1}_i$ have ReLU activation functions.}
    \vspace{-1ex}
\end{figure}
 
This section lists concepts and notations used in the rest of the paper.

\subsection{Deep Neural Networks}

We restrict ourselves to fully-connected DNNs when describing our approach.
However, our approach works for other DNN architectures as demonstrated by our
experimental evaluation~(\pref{sec:ExperimentalEvalation}). We use $x$, $y$, $w$
to denote scalar values, and $X$, $Y$, $W$, and $\theta$ to denote matrix
values. 

\begin{definition}
A \emph{deep neural network (DNN)}~$\dnn$ with $L$ layers and parameters $\theta$ is a
sequence of tuples
$(W\ly{0}, B\ly{0}, \sigma\ly{0}), \cdots, (W\ly{L-1}, B\ly{L-1},
\sigma\ly{L-1})$ with $W\ly{\ell}, B\ly{\ell} \in \theta$.
For the $\ell$th layer,
$n\ly{\ell}$ and $n\ly{\ell+1}$ are the number
of input and output neurons, respectively,
$W\ly{\ell} \in \mathbb{R}^{n\ly{\ell}\times n\ly{\ell+1}}$
is the \emph{weight} matrix,
and
$B\ly{\ell}\in \mathbb{R}^{n\ly{\ell+1}}$
is the \emph{bias} matrix,
and
$\sigma\ly{\ell}: \mathbb{R}^{n\ly{\ell+1}} \!\!\rightarrow \mathbb{R}^{n\ly{\ell+1}}$
is an \emph{activation~function}.
\end{definition}

\begin{definition}
\label{def:identity}
The \emph{identity activation function is a vector-valued function $\identity:
\mathbb{R}^n \rightarrow \mathbb{R}^n$ defined component-wise as}
$    \identity(X)_i = X_i$.
\end{definition}

\begin{definition}
\label{def:relu}
The \emph{ReLU activation function is a vector-valued function $\relu:
\mathbb{R}^n \rightarrow \mathbb{R}^n$ defined component-wise as}
$
\relu(X)_i = \scalarReLU(X_i)
$
where $\scalarReLU(x) = x$ if $x \geq 0$, otherwise $\scalarReLU(x) = 0$.
\end{definition}

\begin{definition}
\label{def:hardswish}
The \emph{Hardswish activation function} is a vector-valued function
$\hardswish: \mathbb{R}^n \rightarrow \mathbb{R}^n$ defined component-wise as
$
    \hardswish(X)_i = \scalarHardswish(X_i)$
where
$\scalarHardswish(x) = 0$ if $x \leq -3$,
$\scalarHardswish(x) = x$ if $x \geq 3$,
otherwise $\scalarHardswish(x) = x\cdot (x+3) / 6$.
\end{definition}

\begin{example}
\label{exa:Overview-DNN}
\pref{fig:Overview-DNN} shows a graphical representation of the DNN $\OverviewDNN$;
weights are shown
on arrows, biases are shown above nodes, and double nodes denote the ReLU
activation function.
The DNN $\OverviewDNN$ has two layers; i.e., $\len(\OverviewDNN) = 2$.
The first layer is $(W\ly{0}, B\ly{0}, \sigma\ly{0})$, where
$W\ly{0} = \vbmatrix{1.5pt}{-1 & 1 & 0.5}$, $B\ly{0} = \vbmatrix{1.5pt}{0 & -2
& 0}$, $\sigma\ly{0} = \relu$. The second layer is $(W\ly{1},
B\ly{1}, \sigma\ly{1})$, where $W\ly{1} = \vbmatrix{1.5pt}{0.5 & -0.5 &
1}^\intercal$, $B\ly{1} = \vbmatrix{1.5pt}{-0.5}$, $\sigma\ly{1} =
\identity$.
\end{example}

\begin{definition}
Given a DNN $\dnn$ with $L$ layers and parameters $\theta$
and an input $X \in \mathbb{R}^{n\ly{0}}$,
the network output $\Ly{X}{L} \eqdef \dnn^\theta(X\ly{0})$, where
$ \Ly{X}{0} \eqdef X$,
and $\Ly{X}{\ell+1} \eqdef \Ly{\sigma}{\ell}(\Ly{X}{\ell} \Ly{W}{\ell} + \Ly{B}{\ell})$.
We use $\dnn(X)$ if the parameters $\theta$ are clear from the context.
\end{definition}

\begin{example}
\label{exa:Overview-DNN-Forward}
Consider the DNN $\OverviewDNN$ in~\pref{fig:Overview-DNN} discussed in \pref{exa:Overview-DNN}.
Given an input point $\Ly{X}{0} =
\vbmatrix{0pt}{4}$, the output $\Ly{X}{1}$ of layer 0 is
$
\Ly{X}{1}
    =  \vbmatrix{0pt}{4}
    \vbmatrix{2pt}{-1 & 1 & 0.5} +
    \vbmatrix{2pt}{0 & -2 & 0}
    =  \vbmatrix{2pt}{-4 & 2 & 2}
$
and the output $\Ly{X}{2}$ of layer 1 is
$
\Ly{X}{2}
    =  \vbmatrix{2pt}{0 & 2 & 2}
    \vbmatrix{2pt}{0.5 & -0.5 & 1}^\intercal +
    \vbmatrix{2pt}{-0.5}
    =  \vbmatrix{2pt}{0.5}
$.
\pref{fig:Overview-DNN-Plot} plots of the output of $\OverviewDNN$
for the input range $[-2, 4]$.
\end{example}

We use $\accuracy(\dnn, S)$ to denote the accuracy of the DNN $\dnn$
on a set $S$.

\begin{definition}
    Given a DNN $\dnn$ with $L$ layers,
    $\SliceDNN{\dnn}{\ell_0}{\ell_1}$ denotes
    the slice of $\dnn$ from $\ell_0$th to the $\ell_1\!-\!1$th layer,
    where
    $\SliceDNN{\dnn}{\ell_0}{\ell_1} \eqdef \big(W\ly{\ell_0}, B\ly{\ell_0},
    \sigma\ly{\ell_0}),
    \big(W\ly{\ell_0+1}, B\ly{\ell_0+1},
    \sigma\ly{\ell_0+1}),\ldots, (W\ly{\ell_1-1}, B\ly{\ell_1-1}, \sigma\ly{\ell_1-1}\big)$.
\end{definition}

The \emph{vertex representation (or V-representation)} of a bounded convex polytope
defines the polytope as the convex hull of a finite set of vertices of the polytope.
We use $\convexhull(P)$ to denote the convex
hull of the finite set of points $P$, and \emph{V-polytope} to denote this
finite set of points defining a bounded convex polytope.

\begin{definition}[V-Polytope]
    \label{def:VPolytope}
    A \emph{V-polytope} $P$ is defined as a finite set of points $\{ \idx{0}X, \idx{1}X, \ldots, \idx{n}X\}$,
    $\idx{i}X \in \mathbb{R}^n$. The V-polytope $P$ represents the bounded convex polytope
    $\convexhull(P)$.
\end{definition}

An alternative representation of convex polytopes is by using an
\emph{H-representation}, which represents a polytope as a set of
linear constraints.
Converting a polytope from V-representation to
H-representation is called the facet enumeration problem, and from
H-representation to V-representation is called the vertex enumeration problem.
It is not known whether there exists a polynomial time algorithm in general for
either of these problems~\cite{bremner1997primal}. In the rest of the paper, we
use ``polytope'' to mean V-polytope, and use ``affine'' and ``linear''
interchangeably.

$\sy x$, $\sy y$, $\sy w$ denotes symbolic variables,
$\sy X$, $\sy Y$, $\sy W$, $\sy \theta$ denotes vectors of symbolic variables.
$\expr_\vars$ and $\formula_\vars$
denote the set of symbolic expressions and formulas over variables $\vars$,
respectively.

\begin{definition}
 A \emph{linear expression} is of the form $\Sigma_i w_i \sy x_i + \sy y$.
 A \emph{linear formula} is of the form
$\bigwedge_i \Sigma_j w_j \sy x_j + \sy y_i \bowtie 0$, where $\bowtie \in \{<, \leq, =, \geq, > \}$.
\end{definition}

$\psi(\sy x_1, \sy x_2, \ldots, \sy x_n)$ denotes a formula over variables $\sy x_i$.
We overload this notation and use $\psi(\sy E_1, \sy E_2,$ $\ldots, \sy E_n)$ to denote a formula
$\psi(\sy x_1, \sy x_2, \ldots, \sy x_n) \wedge \bigwedge_i \sy x_i = \sy E_i$.
Given $\sy E \in \expr_\vars$,
$\eval{\sy E}_V$ denotes the valuation of
the expression $\sy E$ using the value $V$.
We define a \emph{symbolic V-polytope}
$\sy\polytope  = \bigl\{ \idx{0}\sy\point, \cdots, \idx{n}\sy\point \bigr\}$
as a set of symbolic points $\idx{i}\sy\point \in \expr_\vars^m$.
We use $\eval{ \sy\polytopeSet }_V$ to denote
$
\left\{
    \eval{ \idx{0}\sy\point }_V, \cdots,
    \eval{ \idx{n}\sy\point }_V
\right\}
$.

\subsection{Provable Repair}
\label{sec:Preliminaries-ProvableRepair}

Given a DNN $\dnn$ and a repair specification, the goal of provable repair
is to find a DNN~$\dnn'$ that satisfies the given repair
specification. Provable Repair approaches can be classified along two axes:
(a)~the type of repair specifications (pointwise vs.\ polytope), and
(b)~whether or not the approach modifies the original architecture of the DNN
$\dnn$ (architecture modifying vs.\ architecture preserving). Before defining
these four variants of the problem, we define two quantitative metrics used to
evaluate a repair technique: \emph{drawdown} and \emph{generalization}.

The \emph{Drawdown set} is a set of points that are disjoint from the repair
specification and are representative of the existing knowledge in the DNN.
The \emph{Generalization set} is a set of points that are disjoint from but similar
to those in the repair specification.

\begin{definition}
    \label{def:drawdown}
 Given a drawdown set $S$ and DNNs $\dnn$ and $\dnn'$, the \emph{drawdown}
 of $\dnn'$ with respect to $\dnn$ is defined as $\accuracy(\dnn, S) -
 \accuracy(\dnn', S)$.
 Lower drawdown is better.
\end{definition}

\begin{definition}
    \label{def:generalization}
 Given a generalization set $S$ and DNNs $\dnn$ and $\dnn'$, the \emph{generalization}
 of $\dnn'$ with respect to $\dnn$ is defined as $\accuracy(\dnn', S) -
 \accuracy(\dnn, S)$.
 Higher generalization is better.
\end{definition}

\begin{definition}
    \label{def:Preliminaries-PointwiseSpecification}
    A \emph{pointwise repair specification} is a tuple
    $(\pointSet, \spec)$, where $\pointSet \eqdef \{\idx{1}X, \idx{2}X, \ldots, \idx{n}X \}$
    and $\spec \eqdef \{\psi_1, \psi_2, \ldots, \psi_n \}$.
    A \emph{DNN $\dnn^\theta$ satisfies
    $(\pointSet, \spec)$}
    iff the following formula is true:
    $\bigwedge_{1\leq i\leq n} \psi_i\big(\dnn^\theta(\idx{i}X)\big)$.
\end{definition}

\begin{definition}
    \label{def:Preliminaries-PointwiseRepair}
    Given a DNN $\dnn$ and a pointwise repair specification $(\pointSet, \spec)$,
    the \emph{provable pointwise repair problem} is to find a DNN $\dnn'$ that
    satisfies $(\pointSet, \spec)$.
The architecture of $\dnn'$ need not be the same as that of $\dnn$.
\end{definition}

\begin{definition}
    \label{def:Preliminaries-APPointwiseRepair}
    Given a DNN $\dnn^\theta$ and a pointwise repair specification $(\pointSet, \spec)$,
    the \emph{architecture-preserving provable pointwise repair problem} is to
    find parameters $\theta'$ such that  $\dnn^{\theta'}$ satisfies $(\pointSet, \spec)$.
\end{definition}

\begin{definition}
    \label{def:Preliminaries-PolytopeSpecification}
    A \emph{V-polytope repair specification} is a tuple
    $(\polytopeSet, \spec)$, where
$\polytopeSet \eqdef \{\polytope_1, \polytope_2, \ldots, \polytope_n\}$ with
    $\polytope_i$ being a V-polytope, and
$\spec \eqdef \{\psi_1, \psi_2, \ldots, \psi_n \}$
where $\psi_i(X)$ is a linear formula.
A \emph{DNN $\dnn^\theta$  satisfies
    $(\polytopeSet, \spec)$} iff the following
    formula is true:
    $\bigwedge_{1\leq i\leq n} \forall X \in \convexhull(\polytope_i). \psi_i\big(\dnn^\theta(X)\big)$.
\end{definition}

\begin{definition}
    \label{def:Preliminaries-PolytopeRepair}
    Given a DNN $\dnn$ and a V-polytope repair specification $(\polytopeSet, \spec)$,
    the \emph{provable V-polytope repair problem} is to find a DNN $\dnn'$ that
    satisfies $(\polytopeSet, \spec)$.
The architecture of $\dnn'$ need not be the same as that of $\dnn$.
\end{definition}

\begin{definition}
    \label{def:Preliminaries-APPolytopeRepair}
    Given a DNN $\dnn^\theta$ and a V-polytope repair specification $(\polytopeSet, \spec)$,
    the \emph{architecture-preserving provable V-polytope repair problem} is to
    find parameters $\theta'$ such that  $\dnn^{\theta'}$ satisfies $(\polytopeSet, \spec)$.
\end{definition}

\section{Overview}
\label{sec:Overview}

This section illustrates our architecture-preserving provable V-polytope repair
algorithm using a simple DNN $\OverviewDNN$ (\pref{fig:Overview-DNN}). We
highlight the key insights of our approach using a pointwise repair
specification~(\pref{sec:Overview-Pointwise}), before moving onto a V-polytope
repair specification (\pref{sec:Overview-Polytope}). We elide details in favor
of exposition, deferring them to \pref{sec:VPolytopeRepair}. The examples in
this section only use fully-connected linear layers and the ReLU activation
function, which is piecewise-linear.
However, our formalism described in \pref{sec:VPolytopeRepair} handles any
activation functions that have some linear pieces, such as
Hardswish~(\pref{def:hardswish}).
Furthermore, our implementation~(\pref{sec:Implementation})
also handles
convolutional layers,
pooling layers, and residual layers, as
demonstrated by our experimental evaluation~(\pref{sec:ExperimentalEvalation}).

\subsection{Provable Pointwise Repair}
\label{sec:Overview-Pointwise}

Consider the pointwise repair specification $(\pointSet, \spec)$, where
$
    \pointSet \eqdef \big\{ \bpoint{-1.5}, \bpoint{-0.5} \big\}
$, $\spec \eqdef \{ \psi_1,
\psi_1 \}$ and  $\psi_1(\sy Y) \eqdef -0.1 \leq \sy{Y}_0 \leq 0.1$. The DNN
$\OverviewDNN$ does not satisfy $(\pointSet, \spec)$:
$\psi_1\left(\OverviewDNN\bigl(\bpoint{-1.5}\bigr)\right)$
and
$\psi_1\left(\OverviewDNN\bigl(\bpoint{-0.5}\bigr)\right)$
are both false, because
$\OverviewDNN\bigl(\bpoint{-1.5}\bigr)_0 =  0.25 > 0.1$,
$\OverviewDNN\bigl(\bpoint{-0.5}\bigr)_0 = -0.25 < -0.1$.

Let $\sy \theta$ represent the symbolic variables corresponding to each of the
parameters $\theta$ in the DNN
$\dnn: \tyReal^m \rightarrow \tyReal^n$.
Let $\dnn^{\sy \theta}(X) \in \expr_{\sy \theta}^n$ denote the symbolic expression
representing the output of $\dnn$ for input $X \in \tyReal^m$.
Using this notation, we can phrase our architecture-preserving provable
pointwise repair problem for $\OverviewDNN$ as finding a satisfying assignment
$\theta'$ for the following formula $\phi_{\text{spec}} \in \formula_{\sy \theta}$:
\begin{equation}
\label{eq:Overview-pointwisespec}
\phi_{\text{spec}}
\eqdef
-0.1 \leq \OverviewDNN^{\sy \theta}\bigl(\bpoint{-1.5}\bigr)_0 \leq 0.1
\wedge
-0.1 \leq \OverviewDNN^{\sy \theta}\bigl(\bpoint{-0.5}\bigr)_0 \leq 0.1
\end{equation}

In general, the expression $\dnn^{\sy \theta}(X)$ will be highly complex and
non-linear, involving quadratic terms (if the DNN has more than a single layer),
disjunctions (to model piecewise-linear functions such as ReLU), or other
non-linear functions (due to activation functions such as Hardswish). Thus,
directly solving \pref{eq:Overview-pointwisespec} is impractical in most cases.

One observation made by prior approaches~\cite{DBLP:conf/lpar/GoldbergerKAK20,DBLP:conf/pldi/SotoudehT21}
is that if the repair is
restricted to only modify parameters in a single layer, then the quadratic terms
can be avoided in $\dnn^{\sy \theta}(X)$.
In our approach,
we \textbf{extend this observation} as follows:
the quadratic terms can be avoided if we
restrict the repair to only modify weights in some single layer $k$ but allow
modification of \emph{biases in all layers~$\ell \geq k$}.
Though simple in hindsight, allowing more biases to be modified
by the repair improves the quality of the repair in practice.
In our example, we
restrict the repair to only modify first layer weights and all layers' biases;
that is, $\Ly{\sy{W}}{0}$, $\Ly{\sy{B}}{0}$, and $\Ly{\sy{B}}{1}$ are the
symbolic variables corresponding to their respective weights and biases in
$\OverviewDNN$. However, even this restriction is not sufficient to make the
repair problem tractable. The problem remains NP-hard even when the
repair is restricted to only modify the parameters of a single layer and the DNN contains
only ReLU activation functions~\cite{DBLP:conf/lpar/GoldbergerKAK20}.

The \textbf{first key insight} of our approach is to find a \emph{linear formula
$\phi_{\text{lin}}$ that implies $\phi_{\text{spec}}$}.
Any satisfying assignment $\theta'$ of $\phi_{\text{lin}}$
will imply that $\OverviewDNN^{\theta'}$ satisfies the repair specification.
Checking satisfiability of the linear formula can be done in polynomial time~\cite{khachiyanpolylp}
making our repair approach efficient and scalable to large DNNs.
Furthermore, a linear programming (LP) solver can be used to find a $\theta'$ that minimizes, for instance,
the difference between $\theta'$ and the original parameters~$\theta$~\cite{gurobi},
which lowers repair drawdown (\pref{def:drawdown}).

The \textbf{key primitive} we define is a \emph{conditional symbolic function $\condsy
\sigma$} corresponding to an activation function
$\sigma$~(\pref{def:Conditional-Activation-Function}). We illustrate this concept
for the ReLU activation function using the DNN $\OverviewDNN$.
Consider the input point $\Ly{X}{0} = \bpoint{-1.5}$, then
$\Ly{\sy{X}}{1}_0$, the first symbolic output of layer~0,  is
$\ite(
    -1.5 \Ly{\sy{W}}{0}_{0,0} + \Ly{\sy{B}}{0}_0 \geq 0,
    -1.5\Ly{\sy{W}}{0}_{0,0} + \Ly{\sy{B}}{0}_0,
    0
)$.
Notice that in DNN $\OverviewDNN$, we have
$-1.5\Ly{{W}}{0}_{0,0} + \Ly{{B}}{0}_0 \geq 0$.
If we constrain the values of $\Ly{\sy{W}}{0}_{0,0}$ and $\Ly{\sy{B}}{0}_0$
such that
$\varphi_1 \eqdef -1.5\Ly{\sy{W}}{0}_{0,0} + \Ly{\sy{B}}{0}_0 \geq 0$
 is true, then
$\Ly{\sy{X}}{1}_0 = -1.5\Ly{\sy{W}}{0}_{0,0} + \Ly{\sy{B}}{0}_0$.
Similarly,
if we constrain the values of $\Ly{\sy{W}}{0}_{0,1}$ and $\Ly{\sy{B}}{0}_1$
such that $\varphi_2 \eqdef -1.5 \Ly{\sy{W}}{0}_{0,1} + \Ly{\sy{B}}{0}_1 < 0$ is true, then
$\Ly{\sy{X}}{1}_1 = 0$.
If we constrain the values of $\Ly{\sy{W}}{0}_{0,2}$ and $\Ly{\sy{B}}{0}_2$
such that $\varphi_3 \eqdef -1.5 \Ly{\sy{W}}{0}_{0,2} + \Ly{\sy{B}}{0}_2 < 0$ is true, then
$\Ly{\sy{X}}{1}_2 = 0$.
Using this symbolic value for $\Ly{\sy{X}}{1}$, we
get $\Ly{\sy{X}}{2} = \vbmatrix{0pt}{
    - 0.75 \Ly{\sy{W}}{0}_{0,0}
    + 0.5  \Ly{\sy{B}}{0}_0
    +      \Ly{\sy{B}}{1}_0
}$.
We can confirm that the linear formula
$$
\varphi_1 \wedge \varphi_2 \wedge \varphi_3
\wedge -0.1 \leq - 0.75 \Ly{\sy{W}}{0}_{0,0}
+ 0.5  \Ly{\sy{B}}{0}_0
+      \Ly{\sy{B}}{1}_0 \leq 0.1
$$
implies
$
-0.1 \leq \OverviewDNN^{\sy \theta}
\bigl(
    \bpoint{-1.5}
\bigr)_0 \leq 0.1
$.
Similarly, for input $\Ly{X}{0} = \vbmatrix{0pt}{-0.5}$,
we get the linear formula
{\small\[
- 0.5 \Ly{\sy{W}}{0}_{0,0} + \Ly{\sy{B}}{0}_0 \!\geq\! 0 \wedge
- 0.5 \Ly{\sy{W}}{0}_{0,1} + \Ly{\sy{B}}{0}_1 \!<\! 0 \wedge
- 0.5 \Ly{\sy{W}}{0}_{0,2} + \Ly{\sy{B}}{0}_2 \!<\! 0 \wedge
-0.1 \!\leq\!
    - 0.25 \Ly{\sy{W}}{0}_{0,0}
    + 0.5  \Ly{\sy{B}}{0}_0
    +      \Ly{\sy{B}}{1}_0
\!\leq\! 0.1
\]}
which implies
$-0.1 \leq \OverviewDNN^{\sy \theta}\bigl(\bpoint{-0.5}\bigr)_0 \leq 0.1$.

Those constraints corresponding to each of the points in the repair specification
formulates an LP problem. To find the minimal modification of $\OverviewDNN$, we
minimize the $L^\infty$ and normalized $L^1$ norm of the delta of both the parameters and
outputs. The DNN $\OverviewDNNPointwiseRepaired$
(\pref{fig:Overview-Pointwise-RepairedDNN}) shows this minimal repair of
$\OverviewDNN$ that satisfies the given pointwise repair specification $(\pointSet, \spec)$.

\begin{figure}[t]
    \begin{subfigure}{\ParamDNNFigureWidth}
        \centering
        \begin{tikzpicture}[scale=\ParamDNNFigureScale, transform shape]
    \overviewdnnfig
        {\NewParam{-0.4}}{\VarParam{1}}{\VarParam{0.5}}
        {\VarParam{+0}}{\VarParam{-2}}{\VarParam{+0}}
        {0.5}{-0.5}{1}
        {\NewParam{-0.2}}
\end{tikzpicture}
         \vspace{1ex}
        \caption{DNN $\OverviewDNNPointwiseRepaired$. }
        \label{fig:Overview-Pointwise-RepairedDNN}
    \end{subfigure}
\ParamDNNFigureSep{}
    \begin{subfigure}{\ParamDNNPlotWidth}
        \centering\begin{tikzpicture}[scale=\ParamDNNPlotScale,
    refdashed/.style={dashed,black,line width=0.1em},
        point/.style={circle,draw,fill=black,minimum size=0.1em,scale=0.6},
        relu/.style={blue,line width=0.3em},
        linearized/.style={dashed,red,line width=0.3em}]
    \overviewdnnplotrich
        {-2}{(-0.2)*x-0.2}
        {0}{(0.5)*x-0.2}
        {2}{0.8}
        {4}{
    \node[point,red] at (axis cs: -1.5, 0.1) (p1) {};
    \node[above right = 0em and 0em of p1, scale=0.64] {$(-1.5, 0.1)$};
\node[point,red] at (axis cs: -0.5, -0.1) (p2) {};
    \node[below left = 0em and -.6em of p2, scale=0.64] {$(-0.5, -0.1)$};
\draw[refdashed] (axis cs: -2.55, -0.1) -- (axis cs: 4.55, -0.1);
    \draw[refdashed] (axis cs: -2.55,  0.1) -- (axis cs: 4.55,  0.1);
    }
\end{tikzpicture}
         \vspace{-0.8ex}
        \caption{Input-output plot of $\OverviewDNNPointwiseRepaired$.}
        \label{fig:Overview-Pointwise-RepairedDNN-Plot}
    \end{subfigure}
    \vspace{-1ex}
    \caption{DNN $\OverviewDNNPointwiseRepaired$
    that satisfies
    the pointwise repair specification~$(\pointSet, \spec)$ in~\pref{sec:Overview-Pointwise}.}
    \vspace{-1ex}
\end{figure}
 
\subsection{Provable V-Polytope Repair}
\label{sec:Overview-Polytope}

Consider the V-polytope repair specification $(\polytopeSet_1, \spec_1)$
where $\polytopeSet_1 \eqdef \{\polytope_1\}$
with the V-polytope $\polytope_1 \eqdef \bigl\{\bpoint{-1.5}, \bpoint{-0.5}\bigr\}$,
and $\spec_1 \eqdef \{\psi_1\}$
where $\psi_1(\sy Y) \eqdef -0.1 \leq \sy Y_0 \leq 0.1$.
The DNN $\OverviewDNN$ does not satisfy the V-polytope specification $(\polytopeSet_1, \spec_1)$;
viz., the following formula is not true:
\begin{equation}
    \forall X \in
        \convexhull\left(
            \bigl\{\bpoint{-1.5}, \bpoint{-0.5}\bigr\}
        \right).
    -0.1 \leq \OverviewDNN(X)_0 \leq 0.1
\end{equation}
This V-polytope repair specification is an extension of the
pointwise repair specification $(\pointSet, \spec)$ we discussed in~\pref{sec:Overview-Pointwise}:
the pointwise repair specification constrained the behavior
of only the two points
$\bpoint{-1.5}$ and $\bpoint{-0.5}$, while the V-polytope
repair specification constrains the
behavior of $\OverviewDNN$ on all (infinite) points
in
$
\convexhull\left(
    \bigl\{\bpoint{-1.5}, \bpoint{-0.5}\bigr\}
\right)$.
However, notice that the DNN
$\OverviewDNNPointwiseRepaired$~(\pref{fig:Overview-Pointwise-RepairedDNN})
repaired using the pointwise repair specification $(\pointSet, \spec)$ also
satisfies the V-polytope repair specification  $(\polytopeSet_1, \spec_1)$.

\begin{figure}[t]
\begin{subfigure}{\ParamDNNFigureWidth}
        \centering
        \begin{tikzpicture}[scale=\ParamDNNFigureScale, transform shape]
    \overviewdnnfig
        {\NewParam{-0.4}}{\VarParam{1}}{\NewParam{0.366}}
        {\VarParam{+0}}{\VarParam{-2}}{\VarParam{+0}}
        {0.5}{-0.5}{1}
        {\NewParam{-0.2}}
\end{tikzpicture}
         \vspace{1ex}
        \caption{DNN $\OverviewDNNPolytopeWrong$.}
        \label{fig:Overview-Polytope-Wrong}
    \end{subfigure}
\ParamDNNFigureSep{}
    \begin{subfigure}{\ParamDNNPlotWidth}
        \centering
        \begin{tikzpicture}[scale=\ParamDNNPlotScale,
    refdashed/.style={dashed,black,line width=0.1em},
        point/.style={circle,draw,fill=black,minimum size=0.1em,scale=0.6},
        relu/.style={blue,line width=0.3em},
        linearized/.style={dashed,red,line width=0.3em}]
    \overviewdnnplotrich
        {-2}{(-0.2)*x-0.2}
        {0}{(0.3666666666666667)*x-0.2}
        {2}{-(0.1333333333333333)*x + 0.8}
        {4}{
    \node[point,red] at (axis cs: -1.5, 0.1) (p1) {};
    \node[above right = 0em and 0em of p1, scale=0.64] {$(-1.5, 0.1)$};
\node[point,red] at (axis cs: -0.5, -0.1) (p2) {};
    \node[below left = 0em and -.6em of p2, scale=0.64] {$(-0.5, -0.1)$};
\draw[refdashed] (axis cs: -1.5, -0.1) -- (axis cs: -0.5, -0.1);
    \draw[refdashed] (axis cs: -1.5,  0.1) -- (axis cs: -0.5,  0.1);
\node[point,blue] at (axis cs: 1.5, 0.35) (p3) {};
    \node[above left =  0em and 0em of p3, scale=0.64] {$(1.5, 0.35)$};
\node[point,green] at (axis cs: 3, 0.4) (p4) {};
    \node[above right = 0em and 0em of p4, scale=0.64] {$(3, 0.4)$};
\draw[refdashed] (axis cs: 1.5, -0) -- (axis cs: 3, -0);
    \draw[refdashed] (axis cs: 1.5,  0.4) -- (axis cs: 3,  0.4);
}
\end{tikzpicture}
         \vspace{-0.8ex}
        \caption{Input-output plot of DNN $\OverviewDNNPolytopeWrong$.}
        \label{fig:Overview-Polytope-Wrong-Plot}
    \end{subfigure}
\vspace{-1ex}
    \caption{
    DNN
    $\OverviewDNNPolytopeWrong$ that does not satisfy the V-polytope repair specification
    $(\polytopeSet_2, \spec_2)$ in~\pref{sec:Overview-Polytope}.}
    \vspace{-1ex}
\end{figure}
 
Based on this observation, one might be tempted to conclude that
merely repairing the vertices of a V-polytope is sufficient to
repair all points in the convex hull of the vertices.
The next example shows this not to be true.
Consider the V-polytope repair
specification
$(\polytopeSet_2, \spec_2)$
where $\polytopeSet_2 \eqdef \{\polytope_1, \polytope_2\}$
with the V-polytope
$\polytope_1 \eqdef \bigl\{\bpoint{-1.5}, \bpoint{-0.5}\bigr\}$,
$\polytope_2 \eqdef \bigl\{\bpoint{1.5}, \bpoint{3}\bigr\}$,
and $\spec_2 \eqdef \{\psi_1, \psi_2\}$ where
$\psi_1(\sy Y) \eqdef -0.1 \leq \sy Y_0 \leq 0.1$,
$\psi_2(\sy Y) \eqdef 0 \leq \sy Y_0 \leq 0.4$.
The DNN $\OverviewDNN$ does not satisfy this repair specification,
because the following formula is not true:
\begin{align*}
\myvmatrix{2pt}{
    \forall \Ly{X}{0}
        &\hspace{-1pt}\in&\hspace{-1pt}
    \convexhull\bigl( \bigl\{
        &\hspace{-4pt}
    \bpoint{-1.5}
        &\hspace{-4pt},&\hspace{-4pt}
    \bpoint{-0.5}
        &\hspace{-4pt}
    \bigr\} \bigr)
        .&\hspace{-1pt}
    -0.1
        &\hspace{-1pt}\leq&\hspace{-1pt}
    \OverviewDNN(\Ly{X}{0})_0
        &\hspace{-1pt}\leq&\hspace{-1pt}
    0.1
        &\wedge
\\
    \forall \Ly{X}{0}
        &\hspace{-1pt}\in&\hspace{-1pt}
    \convexhull\bigl( \bigl\{
        &\hspace{-4pt}
    \bpoint{1.5}
        &\hspace{-4pt},&\hspace{-4pt}
    \bpoint{3}
        &\hspace{-4pt}
    \bigr\} \bigr)
    \hfill
    \hfill
    \hfill
        .&\hspace{-1pt}
    0
        &\hspace{-1pt}\leq&\hspace{-1pt}
    \OverviewDNN(\Ly{X}{0})_0
        &\hspace{-1pt}\leq&\hspace{-1pt}
    0.4
}
\end{align*}
Suppose we just repair the vertices of the polytopes so that
the following formula is true:
\begin{align*}
\myvmatrix{2pt}{
    -0.1
        &\hspace{-1pt}\leq&\hspace{-1pt}
    \OverviewDNN\bigl(&\hspace{-4pt}\bpoint{-1.5}&\hspace{-4pt}\bigr)_0
        &\hspace{-1pt}\leq&\hspace{-1pt}
    0.1
        &\wedge&
    -0.1
        &\hspace{-1pt}\leq&\hspace{-1pt}
    \OverviewDNN\bigl(&\hspace{-4pt}\bpoint{-0.5}&\hspace{-4pt}\bigr)_0
        &\hspace{-1pt}\leq&\hspace{-1pt}
    0.1
        &\wedge&
\\
\hfill
    0
        &\hspace{-1pt}\leq&\hspace{-1pt}
\hfill
    \OverviewDNN\bigl(&\hspace{-4pt}\bpoint{1.5}&\hspace{-4pt}\bigr)_0
        &\hspace{-1pt}\leq&\hspace{-1pt}
    0.4
        &\wedge&
\hfill
    0
        &\hspace{-1pt}\leq&\hspace{-1pt}
\hfill
    \OverviewDNN\bigl(&\hspace{-4pt}\bpoint{   3}&\hspace{-4pt}\bigr)_0
        &\hspace{-1pt}\leq&\hspace{-1pt}
    0.4
}
\end{align*}
The DNN $\OverviewDNNPolytopeWrong$ in~\pref{fig:Overview-Polytope-Wrong}
shows the corresponding repaired DNN.
As we can see in \pref{fig:Overview-Polytope-Wrong-Plot},
DNN $\OverviewDNNPolytopeWrong$ does not satisfy
the V-polytope specification $(\polytopeSet_2, \spec_2)$.
In particular, though the DNN $\OverviewDNNPolytopeWrong$ satisfies
the specification $\psi_1$ corresponding to the V-polytope
$\polytope_1 \eqdef \bigl\{\bpoint{-1.5}, \bpoint{-0.5}\bigr\}$,
it does not satisfy the specification $\psi_2$ corresponding to
the V-polytope
$\polytope_2 \eqdef \bigl\{\bpoint{1.5}, \bpoint{3}\bigr\}$.
For instance, $\OverviewDNNPolytopeWrong\bigl(\bpoint{2}\bigr)_0 = 0.532 > 0.4$,
which violates the specification $\psi_2$.

What property distinguishes $\polytope_1$ and $\polytope_2$ for the DNNs
$\OverviewDNN$ and $\OverviewDNNPolytopeWrong$?
$\OverviewDNN$
and $\OverviewDNNPolytopeWrong$ are
\emph{\locallylinear{}} (\pref{def:locallylinear}) for $\polytope_1$;
that is, there exists linear functions $f_1$ and $f_2$ such that
$\OverviewDNN(X)  = f_1(X)$  and $\OverviewDNNPolytopeWrong(X) = f_2(X)$ for all $X \in \convexhull(\polytope_1)$.
This leads us to the \textbf{second key insight} of
our approach: if a DNN $\dnn$ is \locallylinear{} for a V-polytope $\polytope$,
then repairing the behavior of $\dnn$ on the vertices of $\polytope$
using our approach
is sufficient to guarantee that $\dnn$ satisfies a V-polytope specification over
$\polytope$. This insight explains the behavior for $\polytope_1$ in the above
example.
However, this insight does not directly indicate how to tackle
the V-polytope $\polytope_2$ for which
$\OverviewDNN$ and $\OverviewDNNPolytopeWrong$ are not \locallylinear.

\begin{figure}[t]
    \begin{subfigure}{\ParamDNNFigureWidth}
        \centering
        \begin{tikzpicture}[scale=\ParamDNNFigureScale, transform shape]
    \overviewdnnfig
        {\VarParam{-1}}{\NewParam{0.75}}{\NewParam{0.333}}
        {\VarParam{+0}}{\NewParam{-2.25}}{\VarParam{+0}}
        {0.5}{-0.5}{1}
        {\VarParam{-0.5}}
\end{tikzpicture}
         \caption{DNN $\OverviewDNNShifted$}
        \label{fig:Overview-Polytope-ShiftedDNN}
    \end{subfigure}
\ParamDNNFigureSep{}
    \begin{subfigure}{\ParamDNNPlotWidth}
        \centering
        \begin{tikzpicture}[scale=\ParamDNNPlotScale,
    refdashed/.style={dashed,black,line width=0.1em},
        point/.style={circle,draw,fill=black,minimum size=0.1em,scale=0.6},
        relu/.style={blue,line width=0.3em},
        linearized/.style={dashed,red,line width=0.3em}]
    \overviewdnnplotrich
        {-2}{(-0.5)*x-0.5}
        {0}{(0.3333333333333333)*x-0.5}
        {3}{-(0.04166666666666674)*x + 0.6250000000000002}
        {4}{
    \node[point,red] at (axis cs: -1.5, 0.25) (p1) {};
    \node[above right = 0em and 0em of p1, scale=0.64] {$(-1.5, 0.25)$};
\node[point,red] at (axis cs: -0.5, -0.25) (p2) {};
    \node[below left = 0em and -.6em of p2, scale=0.64] {$(-0.5, -0.25)$};
\draw[refdashed] (axis cs: -1.5, -0.1) -- (axis cs: -0.5, -0.1);
    \draw[refdashed] (axis cs: -1.5,  0.1) -- (axis cs: -0.5,  0.1);
\node[point,blue] at (axis cs: 1.5, 0) (p3) {};
    \node[above left =  0em and -.2em of p3, scale=0.64] {$(1.5, 0)$};
\node[point,blue] at (axis cs: 3, 0.5) (p4) {};
    \node[above right = 0em and 0em of p4, scale=0.64] {$(3, 0.5)$};
\draw[refdashed] (axis cs: 1.5, -0) -- (axis cs: 3, -0);
    \draw[refdashed] (axis cs: 1.5,  0.4) -- (axis cs: 3,  0.4);
    }
\end{tikzpicture}         \vspace{-1.5ex}
        \caption{Input-output plot of $\OverviewDNNShifted$.}
        \label{fig:Overview-Polytope-ShiftedDNN-Plot}
    \end{subfigure}
    \\
    \begin{subfigure}{\ParamDNNFigureWidth}
        \centering
        \begin{tikzpicture}[scale=\ParamDNNFigureScale, transform shape]
    \overviewdnnfig
        {\VarParam{-1}}{\NewParam{0.75}}{\NewParam{0.333}}
        {\VarParam{+0}}{\NewParam{-2.25}}{\VarParam{+0}}
        {\NewParam{0.2}}{-0.5}{\NewParam{0.6}}
        {\NewParam{-0.2}}
\end{tikzpicture}
 \caption{DNN $\OverviewDNNPolytopeRepaired$.}
        \label{fig:Overview-Polytope-RepairedDNN}
    \end{subfigure}
\ParamDNNFigureSep{}
    \begin{subfigure}{\ParamDNNPlotWidth}
        \centering
        \begin{tikzpicture}[scale=\ParamDNNPlotScale,
    refdashed/.style={dashed,black,line width=0.1em},
        point/.style={circle,draw,fill=black,minimum size=0.1em,scale=0.6},
        relu/.style={blue,line width=0.3em},
        linearized/.style={dashed,red,line width=0.3em}]
    \overviewdnnplotrich
        {-2}{(-0.2)*x-0.2}
        {0}{(0.2)*x-0.2}
        {3}{-(0.1749999999999999)*x + 0.9249999999999997}
        {4}{
    \node[point,red] at (axis cs: -1.5, 0.1) (p1) {};
    \node[above right = 0em and -1.2em of p1, scale=0.64] {$(-1.5, 0.1)$};
\node[point,red] at (axis cs: -0.5, -0.1) (p2) {};
    \node[below left = 0em and -.6em of p2, scale=0.64] {$(-0.5, -0.1)$};
\draw[refdashed] (axis cs: -1.5, -0.1) -- (axis cs: -0.5, -0.1);
    \draw[refdashed] (axis cs: -1.5,  0.1) -- (axis cs: -0.5,  0.1);
\node[point,blue] at (axis cs: 1.5, 0.1) (p3) {};
    \node[above left =  0em and -.7em of p3, scale=0.64] {$(1.5, 0.1)$};
\node[point,blue] at (axis cs: 3, 0.4) (p4) {};
    \node[above right = 0em and 0em of p4, scale=0.64] {$(3, 0.4)$};
\draw[refdashed] (axis cs: 1.5, -0) -- (axis cs: 3, -0);
    \draw[refdashed] (axis cs: 1.5,  0.4) -- (axis cs: 3,  0.4);
    }
\end{tikzpicture}
         \vspace{-1.5ex}
        \caption{Input-output plot of $\OverviewDNNPolytopeRepaired$.}
        \label{fig:Overview-Polytope-RepairedDNN-Plot}
    \end{subfigure}
    \vspace{-1ex}
    \caption{DNN $\OverviewDNNShifted$ that is \locallylinear\ for
    V-polytopes $P_1$ and $P_2$ from $\polytopeSet_2$ in~\pref{sec:Overview-Polytope},
    as well as DNN $\OverviewDNNPolytopeRepaired$
    that satisfies the V-polytope repair specification~$(\polytopeSet_2, \spec_2)$ in~\pref{sec:Overview-Polytope}.}
    \vspace{-1ex}
\end{figure} 
The \textbf{third key insight} of our approach is that
the parameters of
a DNN $\dnn$ can be modified so that $\dnn$ is \locallylinear{} for
a given V-polytope $\polytope$. Consider again the DNN $\OverviewDNN$
and the V-polytope repair specification $(\polytopeSet_2, \spec_2)$.
Consider the following constraints:
\begin{align}
\begin{split}
\myvmatrix{2pt}{
    - 1.5 \Ly{\sy{W}}{0}_{0,0} + \Ly{\sy{B}}{0}_0 \geq 0 &\wedge&
    - 1.5 \Ly{\sy{W}}{0}_{0,1} + \Ly{\sy{B}}{0}_1 < 0 &\wedge&
    - 1.5 \Ly{\sy{W}}{0}_{0,2} + \Ly{\sy{B}}{0}_2 < 0 &\wedge&  \\
- 0.5 \Ly{\sy{W}}{0}_{0,0} + \Ly{\sy{B}}{0}_0 \geq 0 &\wedge&
    - 0.5 \Ly{\sy{W}}{0}_{0,1} + \Ly{\sy{B}}{0}_1 < 0 &\wedge&
    - 0.5 \Ly{\sy{W}}{0}_{0,2} + \Ly{\sy{B}}{0}_2 < 0 &\wedge&  \\
\hfill 1.5 \Ly{\sy{W}}{0}_{0,0} + \Ly{\sy{B}}{0}_0 < 0 &\wedge&
    \hfill 1.5 \Ly{\sy{W}}{0}_{0,1} + \Ly{\sy{B}}{0}_1 < 0 &\wedge&
    \hfill 1.5 \Ly{\sy{W}}{0}_{0,2} + \Ly{\sy{B}}{0}_2 \geq 0 &\wedge&  \\
\hfill 3 \Ly{\sy{W}}{0}_{0,0} + \Ly{\sy{B}}{0}_0 < 0 &\wedge&
    \hfill 3 \Ly{\sy{W}}{0}_{0,1} + \Ly{\sy{B}}{0}_1 < 0 &\wedge&
    \hfill 3 \Ly{\sy{W}}{0}_{0,2} + \Ly{\sy{B}}{0}_2 \geq 0
}
\end{split}
\label{eq:Overview-Polytope-Shift-Activation}
\end{align}
By solving the above linear formula we get the corresponding DNN
$\OverviewDNNShifted$ shown in \pref{fig:Overview-Polytope-ShiftedDNN}. As
shown in its input-output plot~\pref{fig:Overview-Polytope-ShiftedDNN-Plot}, the
DNN $\OverviewDNNShifted$ is \locallylinear{} for both $\polytope_1$
and~$\polytope_2$.

Now we can modify the parameters of $\OverviewDNNShifted$ to repair the behavior
on just the vertices of $\polytope_1$ and $\polytope_2$. We could again choose to
modify the first-layer weights and all biases. However, for illustrative
purposes, we will choose to modify the weights and biases of only the second
layer of $\OverviewDNNShifted$. This leads us to the following linear formula:
\begin{align*}
\myvmatrix{2pt}{
-0.1 &\hspace{-1pt}\leq&\hspace{-1pt}
1.5  \Ly{\sy{W}}{1}_{0,0}
    + \Ly{\sy{B}}{1}_0
&\hspace{-1pt}\leq&\hspace{-1pt} 0.1 &\wedge&
-0.1 &\hspace{-1pt}\leq&\hspace{-1pt}
0.5  \Ly{\sy{W}}{1}_{0,0}
    + \Ly{\sy{B}}{1}_0
&\hspace{-1pt}\leq&\hspace{-1pt} 0.1 &\wedge& \\
\hfill
0 &\hspace{-1pt}\leq&\hspace{-1pt}
\hfill
0.5  \Ly{\sy{W}}{1}_{2,0}
    + \Ly{\sy{B}}{1}_0
&\hspace{-1pt}\leq&\hspace{-1pt} 0.4 &\wedge&
\hfill
0 &\hspace{-1pt}\leq&\hspace{-1pt}
\hfill
\Ly{\sy{W}}{1}_{2,0}
    + \Ly{\sy{B}}{1}_0
&\hspace{-1pt}\leq&\hspace{-1pt} 0.4
}
\end{align*}
\pref{fig:Overview-Polytope-RepairedDNN} presents the resulting repaired DNN
$\OverviewDNNPolytopeRepaired$. As can be seen in its input-output
plot~\pref{fig:Overview-Polytope-RepairedDNN-Plot}, $\OverviewDNNPolytopeRepaired$
satisfies the V-polytope repair specification $(\polytopeSet_2, \spec_2)$.
Comparing $\OverviewDNN$ and $\OverviewDNNPolytopeRepaired$,
we see that our repair approach modified
\emph{weights across
multiple layers and biases across
multiple layers}.

\section{Approach}
\label{sec:VPolytopeRepair}

\begin{figure}[t]
\begin{algorithm}[H]
    \SetInd{.3em}{0.5em}
    \small
    \DontPrintSemicolon
    \caption{$\PolyRepair\bigl(\dnn, \polytopeSet\ly{0}, \spec, s, k\bigr)$}
    \label{alg:VPolytopeRepair}
    \KwIn{
A DNN $\dnn$,
        a set of V-polytopes $\polytopeSet$,
        a repair specification $\spec$,
        a partition $s$,
        and a layer index $k$.
}
    \KwOut{
Repaired DNN $\dnn_\texttt{ret}$ that satisfies $(\polytopeSet, \spec)$, or $\bot$.
}
    \vspace{0.5ex}
    {
\begin{multicols}{2}
\DefAlgBody{
    $
        \dnn_\texttt{ret} \gets \texttt{copy}(\dnn)
    $ \; \label{li:PolyRepair-Copy-Network}
\ForEach(
\label{li:PolyRepair-ForEachSlice-Begin}
    ){$k_i, \ell_i \in s$}{
        $
            \SliceDNN{\dnn_\texttt{ret}}{0}{\ell_i} \gets
                \foo\big(
                    \SliceDNN{\dnn_\texttt{ret}}{0}{\ell_i},
                    \SliceDNN{\dnn}{0}{\ell_i},
                     \polytopeSet\ly{0},
\top,
                     k_i
                \big)
        $ \;
        \label{li:PolyRepair-ShiftAndAssert}
\lIf{$\SliceDNN{\dnn_\texttt{ret}}{0}{\ell_i} = \bot$}{
            \Return $\bot$
                \label{li:PolyRepair-ShiftLinearRegion-Infeasible}
        }
    } \label{li:PolyRepair-ForEachSlice-End}
\Return $
\foo\big(
                \dnn_\texttt{ret},
                \dnn,
                \polytopeSet\ly{0},
                \spec,
                k
            \big)
    $ \;
    \label{li:PolyRepair-Spec}
}

     \vspace{1ex}
\DefHelperFunc{
$\foo\bigl(
    \dnn,
    \dnn_\OriginalKW,
    \polytopeSet\ly{0},
    \spec,
    k
\bigr)$
\label{li:Shift-And-Assert}
}{
    $
        \dnn_\texttt{ret},\, \varphi,\, L
        \gets
        \texttt{copy}(\dnn),\, \top,\, \len(\dnn)
    $ \;
    \label{li:ShiftAndRepair-Init}
\ForEach(
        \label{li:ShiftAndRepair-ForEachPolytope-Begin}
    ){$
\polytope\ly{k} \in \SliceDNN{\dnn}{0}{k}(\polytopeSet\ly{0})
    $}{
$\sy\polytope\ly{L}, \varphi_\sigma
            \gets \SliceDNN{\condsy\dnn}{k}{L}
\big(
                    \polytope\ly{k}
                \big)
        $ \;
        \label{li:ShiftAndRepair-ForEachPolytope-Call}
$\sy\polytopeSet\ly{L}, \varphi \gets
            \append(\sy\polytopeSet\ly{L}, \sy\polytope\ly{L}),
            \varphi \wedge \varphi_\sigma
        $\;
        \label{li:ShiftAndRepair-ForEachPolytope-Add}
    }
    \label{li:ShiftAndRepair-ForEachPolytope-End}
$
        \varphi \gets \varphi \wedge \spec\big( \sy\polytopeSet\ly{L} \big)
    $ \;
    \label{li:ShiftAndRepair-Spec}
$
        \vec\Delta \gets \buildVectorTwo{
            \sy\polytopeSet\ly{L} - \dnn_\OriginalKW\big( \polytopeSet\ly{0} \big)
        }{
            \sy\theta_{\SliceDNN{\dnn}{k}{L}} - \theta_{\SliceDNN{\dnn}{k}{L}}
        }
    $ \;
    \label{li:ShiftAndRepair-Delta}
$
    \theta'_{\SliceDNN{\dnn}{k}{L}} \hspace{-.7ex}\gets
        \Minimize{\vecNorm{\vec\Delta}{p}}{\varphi}
    $ \;
    \label{li:ShiftAndRepair-Minimize}
\lIf(
        \label{li:ShiftAndRepair-Infeasible}
    ){$\theta'_{\SliceDNN{\dnn}{k}{L}} = \bot$}{
        \Return $\bot$
    }
$
        \SliceDNN{\dnn_\texttt{ret}}{k}{L} \gets \Update(
            \SliceDNN{\dnn_\texttt{ret}}{k}{L},
            \theta'_{\SliceDNN{\dnn}{k}{L}}
        )
    $ \;
    \label{li:ShiftAndRepair-Update}
\Return $\dnn_\texttt{ret}$
    \label{li:ShiftAndRepair-Return}
} 
\DefHelperFunc{
$
    \condsy\dnn
\big(
        \polytope\ly{0}
    \big)
$
\label{li:Conditional-Forward-Polytope}
}{
    $\varphi,\, L,\,\point\ly{0}_\RefKW \gets \top,\, \len(\dnn),\, \centroid\big( \polytope\ly{0} \big)$ \;
\ForEach(
        \label{li:Conditional-Forward-Polytope-Each-Begin}
    ){$\idx{j}\point\ly{0} \in \polytope\ly{0}$}{
        $
            \idx{j}\sy\point\ly{L}, \idx{j}\varphi_\sigma \gets
                \condsy\dnn\withRef{
                    \point\ly{0}_\RefKW
}\big(
                    \idx{j}\point\ly{0}
                \big)
        $ \;
        \label{li:Conditional-Forward-Polytope-Each}
}
    \label{li:Conditional-Forward-Polytope-Each-End}
\Return $
        \bigcup_{j}{\bigl\{\idx{j}\sy\point\ly{L}\bigr\}}
        ,\,
        \bigwedge_{j}{\idx{j}\varphi_\sigma}
    $
    \label{li:Conditional-Forward-Polytope-Return}
}
     \DefHelperFunc{
    $\condsy\dnn\withRef{\point\ly{0}_\RefKW}\big(\point\ly{0}\big)$
    \label{li:Conditional-Forward-Point}
}{
    $L \gets \len(\dnn)$
        \\ \label{li:Conditional-Forward-Point-Init}
\ForEach(
\label{li:Conditional-Forward-Point-ForEachLayer-Begin}
    ){$\ell \in \{0, \cdots, L-1\}$}{
        \lIf(
\label{li:Conditional-Forward-Point-Symbolic-Pre-First}
        ){$\ell = 0$}
        {
            $\sy{X}\ly{\ell+1}_\PreKW \gets X\ly{\ell}\sy{W}\ly{\ell} + \sy{B}\ly{\ell}$
        }
\hspace{7.6ex}\lElse(
\label{li:Conditional-Forward-Point-Symbolic-Pre-Other}
        ){
            \hspace{0.8ex}$\sy{X}\ly{\ell+1}_\PreKW \gets \sy{X}\ly{\ell}W\ly{\ell} + \sy{B}\ly{\ell}$
        }
$\ConcreteExec{
            X_{\RefKW}\ly{\ell+1} \gets X_{\RefKW}\ly{\ell}W\ly{\ell} + B\ly{\ell}
        }$  \\ \label{li:Conditional-Forward-Point-Concrete-Pre}
$
            \sy{X}\ly{\ell+1}, \varphi\ly{\ell+1}_\sigma
            \gets \condsy{{\sigma}}\ly{\ell}\withRef{{X_{\RefKW}\ly{\ell+1}}}\bigl(
                \sy{X}\ly{\ell+1}_\PreKW
            \bigr)
        $
            \\ \label{li:Conditional-Forward-Point-Symbolic-Post}
$\ConcreteExec{
            X_{\RefKW}\ly{\ell+1} \gets \sigma\ly{\ell}\bigl(X_{\RefKW}\ly{\ell+1}\bigr)
        }$  \\ \label{li:Conditional-Forward-Point-Concrete-Post}
    }
    \label{li:Conditional-Forward-Point-ForEachLayer-End}
\Return $\sy{X}\ly{L}, \bigwedge_{1 \leq \ell \leq L}{\varphi\ly{\ell}_\sigma}$ \;
    \label{li:Conditional-Forward-Point-Return}
}     \DefHelperFunc{
$
    \condsy\relu\withRef{
        \point_\RefKW
    }\bigl(
        \sy\point
    \bigr)
$
\label{li:SymbolicReLU}
}{
    $\sy{Y},\, \varphi = \buildVector{},\, \top$ \\
\ForEach(
\label{li:SymbolicReLU-ForEach-Begin}
    ){
        $\sy{X}_i, {X}^\RefKW_i \in \zip(\sy{X}, {X}_{\RefKW})$
    }{
        \uIf{${X}^{\RefKW}_i \geq 0$}
        {\hspace{1.95ex}
            $\sy{Y},\, \varphi \gets \append(\sy{Y}, \sy{X}_i),\, \varphi \wedge \sy{X}_i \geq 0$
            \label{li:SymbolicReLU-On}
        }\lElse(
\label{li:SymbolicReLU-Off}
        ){$\sy{Y},\, \varphi \gets \append(\sy{Y}, 0),\, \varphi \wedge \sy{X}_i < 0$
        }
    }
    \label{li:SymbolicReLU-ForEach-End}
    \Return $\sy{Y},\, \varphi$
    \label{li:SymbolicReLU-Return}
}
     \DefHelperFunc{
$
    \condsy\hardswish\withRef{
        \point_\RefKW
    }\bigl(
        \sy\point
    \bigr)
$
\label{li:SymbolicHardSwish}
}{
    $\sy{Y},\, \varphi = \buildVector{},\, \top$ \\
\ForEach(
\label{li:SymbolicHardSwish-ForEach-Begin}
    ){
        $\sy{X}_i, {X}^\RefKW_i \in \zip(\sy{X}, {X}_{\RefKW})$
    }{
        \uIf(
){${X}^{\RefKW}_i \geq 0$}{
            \hspace{2.4ex}
            $\sy{Y},\, \varphi \gets \append(\sy{Y}, \sy{X}_i),\,
            \varphi \wedge \sy{X}_i \geq 3$
            \label{li:SymbolicHardSwish-On}
        }
        \lElse(
\label{li:SymbolicHardSwish-Off}
        )
{
            $\sy{Y},\, \varphi \gets \append(\sy{Y}, 0),\,
            \varphi \wedge \sy{X}_i \leq -3$
        }
    }
    \label{li:SymbolicHardSwish-ForEach-End}
    \Return $\sy{Y},\, \varphi$
    \label{li:SymbolicHardSwish-Return}
}
     \end{multicols}
    }
\vspace{-1.5ex}
\end{algorithm}
\vspace{-.5ex}
\end{figure}

This section describes our architecture-preserving provable V-polytope repair
algorithm~(\pref{alg:VPolytopeRepair}). A pointwise repair
specification~(\pref{def:Preliminaries-PointwiseSpecification}) can be expressed
as a V-polytope repair
specification~(\pref{def:Preliminaries-PolytopeSpecification}) in which each
polytope is a single point. Consequently, an approach for provable V-polytope
repair subsumes provable pointwise repair. Thus, we only discuss our algorithm
for architecture-preserving provable V-polytope repair. Our
experimental evaluation demonstrates that our tool handles both provable
pointwise repair and provable V-polytope repair.

The function
$
\PolyRepair\bigl(
        \dnn,
        \polytopeSet\ly{0},
        \spec,
        s,
        k
    \bigr)
$ in \pref{alg:VPolytopeRepair}
takes as input a DNN $\dnn$,
a V-polytope repair specification
$\bigl( \polytopeSet\ly{0}, \spec \bigr)$~(\pref{def:Preliminaries-PolytopeSpecification}),
a network partition $s$~(\pref{def:Network-Partition}),
and a layer index $k$.
The function returns either $\bot$, indicating the algorithm was unable to
repair the DNN $\dnn$ with the given arguments, or returns a repaired DNN
$\dnn_\texttt{ret}$ that satisfies $\bigl( \polytopeSet\ly{0}, \spec \bigr)$.
The algorithm modifies parameters in multiple layers as determined by $s$ and
$k$.

In the rest of the section, we describe the working of
\pref{alg:VPolytopeRepair} using examples,
as well as a series of theorems and their proof sketches related
to its correctness and efficiency. Detailed proofs can be found in
\onlyfor{pldi}{the extended version~\cite{APRNNExtended}}{\pref{app:Appendix-Proof}}.
We first introduce the concept of a function being \emph{locally linear}
for a polytope $P$,
which is key to our approach.

\begin{definition}
    \label{def:locallylinear}
Given a function $g$ and a V-polytope $P$,
$g$ is \emph{\locallylinear} for the polytope $P$ iff
there exists a linear function $f$ such that
$g(X)  = f(X)$ for all $X \in \convexhull(P)$.
\end{definition}

Intuitively, suppose a function $g$ has linear pieces, then $g$ being
\emph{\locallylinear} for the polytope $P$ implies that $P$ is entirely in a
linear piece of $g$. DNNs that use piecewise-linear activation functions, such
as ReLU, or activation functions that have linear pieces, such as Hardswish,
have linear pieces.

\begin{example}
\label{exa:locallylinear}
  Consider the DNN $\OverviewDNN$ shown in \pref{fig:Overview-DNN}
  and the corresponding input-output plot of \pref{fig:Overview-DNN-Plot}.
  For example, $\OverviewDNN$ is \locallylinear\ for V-polytopes
  $\{[-1.5], [-0.5]\}$, $\{[-2], [0]\}$, $\{[0], [2]\}$, and $\{ [2], [4]\}$.
  $\OverviewDNN$ is not \locallylinear\ for V-polytopes $\{[-1], [1]\}$ and $\{[1.5], [3]\}$.
\end{example}

\begin{definition}
\label{def:Conditional-Activation-Function}
For a given activation function
$\sigma: \tyReal^m \rightarrow \tyReal^n$,
a \emph{conditional symbolic activation function}
$
    \condsy\sigma\withRef{\point_\RefKW}:
        \expr_{\stheta}^m \rightarrow
        \expr_{\stheta}^n \times \formula_{\sy{\theta}}
$
with $\point_\RefKW \in \tyReal^m$
 satisfies the following conditions:
\begin{cs}
\item\labelInMainText{cond:Conditional-Activation-Function-C1}
If
        $
            \sy\pointAlt, \varphi \eqdef \condsy\sigma\withRef{\point_\RefKW}\bigl(
                \sy\point
            \bigr)
        $,
        then $\theta \vDash \varphi$ implies
        $
            \eval{\sy\pointAlt}_{\theta} = \sigma\bigl(
                \eval{\sy\point}_{\theta}
            \bigr)
        $.
\item\labelInMainText{cond:Conditional-Activation-Function-C2}
    If
        $
            \sy\pointAlt, \varphi \eqdef \condsy\sigma\withRef{\point_\RefKW}\bigl(
                \sy\point
            \bigr)
        $
        and
        $\sy\point$ is linear,
        then
        $\sy\pointAlt$ is a linear expression and
        $\varphi$ is a linear~formula.
\item\labelInMainText{cond:Conditional-Activation-Function-C3}
    Let the symbolic polytope
    $
        \sy\polytope \eqdef
            \left\{
                \idx{0}\sy\point,
                \idx{1}\sy\point,
                \ldots,
                \idx{p-1}\sy\point,
            \right\}
    $,
    $
        \idx{j}\sy\pointAlt, \idx{j}\varphi \eqdef
            \condsy\sigma\withRef{\point_\RefKW}
            \bigl( \idx{j}\sy\point \bigr)
    $ for $0 \leq j < p$, and
    $\varphi_{\sy\polytope} \eqdef \bigwedge_j \idx{j}\varphi$.
    If $\theta \vDash \varphi_{\sy\polytope}$,
    then $\sigma$ is \locallylinear{} for the polytope $\eval{ \sy\polytope }_\theta$.
\end{cs}
\end{definition}

Let
$
    \sy\pointAlt, \varphi \eqdef \condsy\sigma\withRef{\point_\RefKW}\bigl(
        \sy\point
    \bigr)
$;
the intuition behind a conditional symbolic activation function $\condsy\sigma$
with a reference point $\point_\RefKW$
is to \emph{deterministically} constrain the symbolic input $\sy\point$ to a
linear piece of $\sigma$ using a linear formula $\varphi$ such that the symbolic
output $\sy\pointAlt$ exactly encodes $\sigma\bigargs{\sy\point}$.
\pref{cond:Conditional-Activation-Function-C1} states that
$
    \eval{\sy\pointAlt}_{\theta} = \sigma\bigl(
        \eval{\sy\point}_{\theta}
    \bigr)
$ for any assignment $\theta$ that satisfies $\varphi$,
which ensures the correctness of this conditional encoding of
$\sigma\bigargs{\sy\point}$.
\pref{cond:Conditional-Activation-Function-C2} is necessary for formulating the
repair problem as an LP problem; \pref{cond:Conditional-Activation-Function-C1} and
 \pref{cond:Conditional-Activation-Function-C2} imply that $\varphi$ constrains
$\sy\point$ to a linear piece of $\sigma$.
\pref{cond:Conditional-Activation-Function-C3} is necessary for provable
polytope
repair~(\pref{thm:Conditional-Forward-Polytope}-\pref{cond:Conditional-Forward-Polytope-C3}).
It implies that $\condsy\sigma\withRef{\point_\RefKW}$ constrains any
input~$\sy\point$ to the same linear piece as determined by $\point_\RefKW$.

The following example demonstrates the conditional symbolic activation function
$\condsy\relu$~(\pref{li:SymbolicReLU} in~\pref{alg:VPolytopeRepair})
for $\relu$.

\begin{example}
\label{exa:cond-relu}
Let
$
\sy\pointAlt, \varphi \eqdef \condsy\relu\withRef{\point_\RefKW}\bigl(
    \sy\point
\bigr)
$
where
$
\sy\point \eqdef \vbinlinematrix{2pt}{
\sy\point_0 & \sy\point_1
}
$ and
$
\point_\RefKW \eqdef \vbinlinematrix{2pt}{
    5 &
    -2
}
$. We will show that
$
\sy\pointAlt =
\vbinlinematrix{2pt}{
    \sy\point_0 &
    0
}
$ and
$
\varphi =
\sy\point_0 \geq 0 \wedge
\sy\point_1 < 0
$.
On~\pref{li:SymbolicReLU-ForEach-Begin}--\ref{li:SymbolicReLU-ForEach-End},
$\condsy\relu$ loops over each component~$i$ and constrains $\spoint_i$ to the
linear piece containing $\point^\RefKW_i$.
Recall that $\relu(X)_i = \scalarReLU(X_i)$ is a piecewise linear function
where $\scalarReLU(x) = x$ if $x \geq 0$, otherwise $\scalarReLU(x) = 0$.
\begin{steps}
\item[{\crtcrossreflabel{{\bfseries S1a}}[step:ReLU-On]}]
    If $\point^\RefKW_i \geq 0$, \pref{li:SymbolicReLU-On} constrains $\spoint_i$ to
    the linear piece $\scalarReLU(x) = x$ where $x \geq 0$ with $\spoint_i \geq 0$,
    which implies $\sy\pointAlt_i \eqdef \spoint_i$.
In this example, because $\point^\RefKW_0 = 5 \geq 0$, $\sy\point_0
    \geq 0$ is in $\varphi$ and~$\sy\pointAlt_0 \eqdef \sy\point_0$.
\item[{\crtcrossreflabel{{\bfseries S1b}}[step:ReLU-Off]}]
    If $\point^\RefKW_i < 0$, \pref{li:SymbolicReLU-Off} constrains $\spoint_i$ to
    the linear piece $\scalarReLU(x) = 0$ where $x < 0$ with $\spoint_i < 0$,
    which implies $\sy\pointAlt_i \eqdef 0$.
In this example, because $\point^\RefKW_1 = -2 < 0$, $\sy\point_1
    < 0$ is in $\varphi$ and~$\sy\pointAlt_1 \eqdef 0$.
\end{steps}
\end{example}

\begin{theorem}
\label{thm:Conditional-ReLU}
$\condsy\relu$
is a conditional symbolic activation function for \relu. \end{theorem}

\begin{proofsketch}
\pref{cond:Conditional-Activation-Function-C1} is satisfied because $\spoint_i
\geq 0$ implies $\sy\pointAlt_i = \spoint_i$ and $\spoint_i < 0$ implies
$\sy\pointAlt_i = 0$ using the definition of ReLU.
\pref{cond:Conditional-Activation-Function-C2} is satisfied because $\spoint_i
\geq 0$, $\spoint_i < 0$ are linear formulas and $\sy\pointAlt_i = \spoint_i$,
$\sy\pointAlt_i = 0$ are linear~expressions if $\spoint_i$ is a linear
expression.
\pref{cond:Conditional-Activation-Function-C3} is satisfied because
$\condsy\relu\withRef{\point_\RefKW}$ constrains any $\spoint_i$ to the linear
piece that contains $\point_\RefKW$.
Given $\theta\vDash\varphi_{\sy\polytope}$, $\eval{\sy\polytope}_\theta$ is
entirely in the same linear piece, hence $\relu$ is
\locallylinear\ for~$\eval{\sy\polytope}_\theta$.
\onlyfor{arxiv}{\refProof{thm:Conditional-ReLU}{proof:Conditional-ReLU}}{}
\end{proofsketch}

Moreover, for a non-piecewise-linear activation function $\sigma$, there exists
$\condsy\sigma$ if $\sigma$ has linear pieces.
The following example demonstrates $\condsy\hardswish$~(\pref{li:SymbolicHardSwish} in
\pref{alg:VPolytopeRepair}).
\onlyfor{pldi}{The proof that $\condsy\hardswish$ is a conditional symbolic activation function can be found in the extended version~\cite{APRNNExtended}.}{}

\begin{example}
\label{exa:cond-hardswish}
Let
$
\sy\pointAlt, \varphi \eqdef \condsy\hardswish\withRef{\point_\RefKW}\bigl(
    \sy\point
\bigr)
$
where
$
\sy\point \eqdef \vbinlinematrix{2pt}{
\sy\point_0 & \sy\point_1
}
$ and
$
\point_\RefKW \eqdef \vbinlinematrix{2pt}{
    5 &
    -2
}
$. We will show that
$
\sy\pointAlt =
\vbinlinematrix{2pt}{
    \sy\point_0 &
    0
}
$ and
$
\varphi =
\sy\point_0 \geq 3 \wedge
\sy\point_1 \leq -3
$.
On~\pref{li:SymbolicHardSwish-ForEach-Begin}--\ref{li:SymbolicHardSwish-ForEach-End},
$\condsy\hardswish$ loops over each component $i$ and constrains $\spoint_i$ to
the linear piece closest to $\point^\RefKW_i$.
Recall that $\hardswish(X)_i = \scalarHardswish(X_i)$ has two linear pieces:
$\scalarHardswish(x) = x$ if $x \geq 3$ and $\scalarHardswish(x) = 0$ if $x \leq
-3$.
\begin{steps}
\item[{\crtcrossreflabel{{\bfseries S1a}}[step:Hardswish-On]}]
    If $\point^\RefKW_i \geq 0$, \pref{li:SymbolicHardSwish-On} constrains
    $\spoint_i$ to the closest linear piece $\scalarHardswish(x) = x$ where $x \geq 3$
    with $\spoint_i \geq 3$, which implies
    $\sy\pointAlt_i \eqdef \spoint_i$.
Here
    because $\point^\RefKW_0 = 5 \geq 0$, $\sy\point_0 \geq 3$ is in $\varphi$
    and~$\sy\pointAlt_0 \eqdef \sy\point_0$.
\item[{\crtcrossreflabel{{\bfseries S1b}}[step:Hardswish-Off]}]
    If $\point^\RefKW_i < 0$, \pref{li:SymbolicHardSwish-Off} constrains
    $\spoint_i$ to the closest linear piece $\scalarHardswish(x) = 0$ where $x \leq -3$
    with $\spoint_i \leq -3$, which implies $\sy\pointAlt_i \eqdef 0$.
Here
because $\point^\RefKW_1 = -2 < 0$, $\sy\point_1
    \leq -3$ is in $\varphi$ and~$\sy\pointAlt_1 \eqdef 0$.
\end{steps}
\end{example}

\onlyfor{arxiv}{
\begin{theorem}
    \label{thm:Conditional-Hardswish}
    $\condsy\hardswish$
is a conditional symbolic activation function for \hardswish.     \end{theorem}

    \begin{proofsketch}
\pref{cond:Conditional-Activation-Function-C1} is satisfied because $\spoint_i
    \geq 3$ implies $\sy\pointAlt_i = \spoint_i$ and $\spoint_i \leq -3$ implies
    $\sy\pointAlt_i = 0$ using the definition of Hardswish.
\pref{cond:Conditional-Activation-Function-C2} is satisfied because $\spoint_i
    \geq 3$, $\spoint_i \leq -3$ are linear formulas and $\sy\pointAlt_i = \spoint_i$,
    $\sy\pointAlt_i = 0$ are linear~expressions if $\spoint_i$ is a linear
    expression.
\pref{cond:Conditional-Activation-Function-C3} is satisfied because
    $\condsy\hardswish\withRef{\point_\RefKW}$ constrains any $\spoint_i$ to the linear
    piece that contains $\point_\RefKW$.
Given $\theta\vDash\varphi_{\sy\polytope}$, $\eval{\sy\polytope}_\theta$ is
    entirely in the same linear piece, hence $\hardswish$ is
    \locallylinear\ for~$\eval{\sy\polytope}_\theta$.
\onlyfor{arxiv}{\refProof{thm:Conditional-Hardswish}{proof:Conditional-Hardswish}}{}
\end{proofsketch}
}{}

Next we present
the conditional symbolic forward execution
$
\condsy\dnn\withRef{\point\ly{0}_\RefKW}
$ \emph{for a concrete point} $\point\ly{0}$
~(\pref{li:Conditional-Forward-Polytope}
in~\pref{alg:VPolytopeRepair})
via a walk-through example.
Let
$
\sy\point\ly{L}, \varphi
\eqdef
\condsy\dnn\withRef{\point\ly{0}_\RefKW}\bigargs{\point\ly{0}}
$.
$\condsy\dnn\withRef{\point\ly{0}_\RefKW}$
takes a concrete point $\point\ly{0}$,
makes the first layer weight $\sW\ly{0}$ and all layers' bias
$\sB\ly{\ell}$ symbolic, then
uses conditional symbolic activation functions to forward execute
$\point\ly{0}$ with the reference~point~$\point\ly{0}_\RefKW$.
It returns the symbolic output point $\sy\point\ly{L}$ with constraint $\varphi$.

\begin{example}
\label{exa:Conditional-Forward-Point}
Consider the DNN $\OverviewDNN$~(\pref{fig:Overview-DNN}) and input point
$\point\ly{0} \eqdef \bpoint{-1.5}$ from~\pref{sec:Overview-Pointwise}. Let
$
    \sy\point\ly{2}, \varphi
    \eqdef
    \condsy\OverviewDNN
    \withRef{
\point\ly{0}_\RefKW
}
    \big(
        \point\ly{0}
\big)
$ where $\point\ly{0}_\RefKW \eqdef \point\ly{0}$.
We will show that
$
    \sy\point\ly{2}
    \!=\! \vbscalematrix{0.85}{0pt}{
        - 0.75 \Ly{\sy{W}}{0}_{0,0}
        + 0.5  \Ly{\sy{B}}{0}_0
        +      \Ly{\sy{B}}{1}_0
    }
$
and
$
    \varphi
    =
    \varphi\ly{1}_\sigma
$
where
$
    \varphi\ly{1}_\sigma
    \eqdef
    -1.5 \Ly{\sy{W}}{0}_{0,0}\!+\!\Ly{\sy{B}}{0}_0\!\geq\!0
    \,\wedge\,
    -1.5 \Ly{\sy{W}}{0}_{0,1}\!+\!\Ly{\sy{B}}{0}_1 \!<\! 0
    \,\wedge\,
    -1.5 \Ly{\sy{W}}{0}_{0,2}\!+\!\Ly{\sy{B}}{0}_2 \!<\! 0
$.

Lines \ref{li:Conditional-Forward-Point-ForEachLayer-Begin}--\ref{li:Conditional-Forward-Point-ForEachLayer-End} runs
$\OverviewDNN$
on $\point\ly{0}$
layer by layer.
The following steps are used for the first layer~$\OverviewDNN\ly{0}$:
\begin{steps}[start=2]
\item[{\crtcrossreflabel{{\bfseries S1a}}[step:Conditional-Forward-Point-Symbolic-Pre-First]}]
    \pref{li:Conditional-Forward-Point-Symbolic-Pre-First} applies the layer
    affine transformation
    $
    \sy\point\ly{1}_\PreKW \gets X\ly{0} \sy{W}\ly{0} + \sy{B}\ly{0}
    $
    with the symbolic layer weight $\sW\ly{0}$ and bias $\sB\ly{0}$ to the layer
    input $X\ly{0}$.
$\sy\point\ly{1}_\PreKW \eqdef\! \vbscalematrix{0.85}{2pt}{
    - 1.5 \Ly{\sy{W}}{0}_{0,0}\!+\!\Ly{\sy{B}}{0}_0 &
    - 1.5 \Ly{\sy{W}}{0}_{0,1}\!+\!\Ly{\sy{B}}{0}_1 &
    - 1.5 \Ly{\sy{W}}{0}_{0,2}\!+\!\Ly{\sy{B}}{0}_2
    }$.

\item\label{step:Conditional-Forward-Point-Concrete-Pre}
    \pref{li:Conditional-Forward-Point-Concrete-Pre} applies the layer affine
    transformation
    $
    X_{\RefKW}\ly{1} \gets X_{\RefKW}\ly{0}W\ly{0} + B\ly{0}
    $
    with the original layer weight $W\ly{0}$ and bias $B\ly{0}$ to the reference
    point $X_{\RefKW}\ly{0}$,
    resulting in
    $X_{\RefKW}\ly{1} \eqdef \vbinlinematrix{3pt}{1.5 & -3.5 & -0.75}$.

\item\label{step:Conditional-Forward-Point-Symbolic-Post}
    \pref{li:Conditional-Forward-Point-Symbolic-Post} applies the conditional activation function
    $
        \sy\point\ly{1}, \varphi\ly{1}_\sigma
\gets
        \condsy{{\sigma}}\ly{0}\withRef{
            X_{\RefKW}\ly{1}
        }\bigl( \sy{X}\ly{1}_\PreKW \bigr)
    $
    with the transformed reference point $X_{\RefKW}\ly{1}$
    to $\sy\point\ly{1}_\PreKW$,
    where
    $
    \sy\point\ly{1} \eqdef \vbscalematrix{0.85}{3pt}{
        -1.5 \sW\ly{0}_{0,0}\!+\!\sB\ly{0}_0 &
        0 &
        0
    }
    $
    and $\varphi\ly{1}_\sigma$ is shown~above.

\item\label{step:Conditional-Forward-Point-Concrete-Post}
    \pref{li:Conditional-Forward-Point-Concrete-Post} applies the activation
    function
    $
    X_{\RefKW}\ly{1} \gets \sigma\ly{0}\bigargs{X_{\RefKW}\ly{1}}
    $
    to
$X_{\RefKW}\ly{1}$,
    resulting in
    $X_{\RefKW}\ly{1} \eqdef \vbinlinematrix{3pt}{1.5 & 0 & 0}$.
\end{steps}

For the second layer
$
\OverviewDNN\ly{1}
$,
a different
\pref{step:Conditional-Forward-Point-Symbolic-Pre-Other}
is used to avoid quadratic terms:
\begin{steps}
\item[{\crtcrossreflabel{{\bfseries S1b}}[step:Conditional-Forward-Point-Symbolic-Pre-Other]}]
    \pref{li:Conditional-Forward-Point-Symbolic-Pre-Other}
    applies the layer affine transformation
    $
    \sy\point\ly{2}_\PreKW \gets \spoint\ly{1} W\ly{1} + \sy{B}\ly{1}
    $
    with the original layer weight $W\ly{1}$ and symbolic bias $\sB\ly{1}$
    to the symbolic layer input $\spoint\ly{1}$.
$\sy\point\ly{2}_\PreKW \eqdef \vbscalematrix{0.85}{0pt}{
        - 0.75 \Ly{\sy{W}}{0}_{0,0}
        \!+\! 0.5  \Ly{\sy{B}}{0}_0
        \!+\!      \Ly{\sy{B}}{1}_0
    }$.
\end{steps}

Then steps
\ref{step:Conditional-Forward-Point-Concrete-Pre},
\ref{step:Conditional-Forward-Point-Symbolic-Post}
and
\ref{step:Conditional-Forward-Point-Concrete-Post}
are repeated.
\pref{step:Conditional-Forward-Point-Concrete-Pre}
results in
$X_{\RefKW}\ly{2} \eqdef \vbinlinematrix{0pt}{0.25}$.
\pref{step:Conditional-Forward-Point-Symbolic-Post}
results in
$\sy\point\ly{2} \eqdef \sy\point\ly{2}_\PreKW$
and
$\varphi\ly{2}_\sigma \eqdef \top$
because $\OverviewDNN\ly{1}$ is the last layer
with an identity activation function.
\pref{step:Conditional-Forward-Point-Concrete-Post}
results in
$X_{\RefKW}\ly{2} \eqdef \vbinlinematrix{0pt}{0.25}$.
Finally, on~\pref{li:Conditional-Forward-Point-Return}, $\sy\point\ly{2}$ and
$\bigwedge_{1\leq\ell\leq 2}{\varphi\ly{\ell}_\sigma} = \varphi\ly{1}_\sigma$
are returned.

\end{example}

\begin{theorem}
\label{thm:Conditional-Forward-Pointwise}
Let
$
    \sy\point\ly{L}, \varphi
    \eqdef
    \condsy\dnn\withRef{\point\ly{0}_\RefKW}
    \big(\point\ly{0}\big)
$
and $\theta \vDash \varphi$, then
\begin{cs}
\item\labelInMainText{cond:Conditional-Forward-Pointwise-Correct}
    $
        \eval{\sy\point\ly{L}}_{\theta}
        =
        \dnn^{\theta}\big(\point\ly{0}\big)
    $.
\item\labelInMainText{cond:Conditional-Forward-Pointwise-Linear}
    $\sy\point\ly{L}$ is a linear expression and $\varphi$ is a linear formula.
\item\labelInMainText{cond:Conditional-Forward-Pointwise-LocallyLinear}
    Let the V-polytope
    $
        \polytope\ly{0} \eqdef
            \left\{
                \idx{0}\point\ly{0},
                \idx{1}\point\ly{0},
                \ldots,
                \idx{p-1}\point\ly{0}
            \right\}
    $,
    $
        \idx{j}\sy\point\ly{L}, \idx{j}\varphi \eqdef
            \condsy\dnn
            \withRef{\point\ly{0}_\RefKW}
            \bigl(
                \idx{j}\point\ly{0}
            \bigr)
    $ for $0 \leq j < p$, and
     $\varphi_\polytope \eqdef \bigwedge_j \idx{j}\varphi$.
    If $\,\theta \vDash \varphi_\polytope$,
    then $\dnn^\theta$ is \locallylinear{} for the polytope $\polytope\ly{0}$.
\end{cs} \end{theorem}

\begin{proofsketch}
\pref{cond:Conditional-Forward-Pointwise-Correct}
states the correctness of the
conditional encoding for $\sy\point\ly{L}$.
By~\pref{def:Conditional-Activation-Function}-\ref{cond:Conditional-Activation-Function-C1},
for the first layer, we have
$
    \eval{
        \sy\point\ly{1}
    }_\theta
    =
    \sigma\ly{0}\bigl(
    \eval{
        \sy\point\ly{1}_\PreKW
    }_{\theta}
    \bigr)
    =
    \sigma\ly{0}\bigl(
X\ly{0} \eval{\sy{W}\ly{0}}_{\theta} + \eval{\sy{B}\ly{0}}_{\theta}
\bigr)
    =
$
$\dnn\ly{0}\bigargs{
        X\ly{0}; \eval{\sy{W}\ly{0}}_{\theta}, \eval{\sy{B}\ly{0}}_{\theta}
    }
$.
For later layers $\ell \geq 1$,
$
    \eval{
        \sy\point\ly{\ell+1}
    }_\theta
    =
    \sigma\ly{\ell}\bigl(
    \eval{
        \sy\point\ly{\ell+1}_\PreKW
    }_{\theta}
    \bigr)
    =
    \sigma\ly{\ell}\bigl(
    \eval{
        \sy{X}\ly{\ell}
    }_{\theta}
    {W}\ly{\ell} +
    \eval{
        \sy{B}\ly{\ell}
    }_{\theta}
    \bigr)
    = \dnn\ly{\ell}\bigargs{
        \eval{
            \sy{X}\ly{\ell}
        }_{\theta};
        {W}\ly{\ell},
        \eval{
            \sy{B}\ly{\ell}
        }_{\theta}
        \bigr)
    }
$.
Thus, by induction, $
    \eval{\sy\point\ly{L}}_{\theta}
    =
    \dnn^{\theta}\big(\point\ly{0}\big)
$.
\pref{cond:Conditional-Forward-Pointwise-Linear}
follows from \pref{def:Conditional-Activation-Function}-\ref{cond:Conditional-Activation-Function-C2}
and
is necessary to formulate the repair problem as an LP problem.
\pref{cond:Conditional-Forward-Pointwise-LocallyLinear}
is necessary for provable polytope repair. It lifts and can be proved using
\pref{def:Conditional-Activation-Function}-\ref{cond:Conditional-Activation-Function-C3}.
\onlyfor{arxiv}{\refProof{thm:Conditional-Forward-Pointwise}{proof:Conditional-Forward-Pointwise}
}{}
\end{proofsketch}

Next we present the conditional symbolic forward execution
$\condsy\dnn\bigargs{\polytope\ly{0}}$~(\pref{li:Conditional-Forward-Point}
in~\pref{alg:VPolytopeRepair}) that lifts
$\condsy\dnn\withRef{\point\ly{0}_\RefKW}\bigargs{\point\ly{0}}$ for a concrete
point $\point\ly{0}$ to a concrete V-polytope $\polytope\ly{0}$.

\begin{example}
\label{exa:Conditional-Forward-Polytope}
Consider the DNN $\OverviewDNN$~(\pref{fig:Overview-DNN}) and polytope $
\polytope_1 \eqdef
\bigl\{
    \bpoint{-1.5},
    \bpoint{-0.5}
\bigr\}
$ from~\pref{sec:Overview-Polytope}.
Let
$
    \sy\polytope\ly{2}, \varphi
    \eqdef \condsy\OverviewDNN\bigargs{\polytope_1}
$.
We will show that $\varphi$ is the first two rows
of~\pref{eq:Overview-Polytope-Shift-Activation},
$
\sy\polytope\ly{2} \eqdef
\bigl\{
    \idx{0}\spoint\ly{2},
    \idx{1}\spoint\ly{2}
\bigr\}
$
where
$
\idx{0}\spoint\ly{2} \eqdef \vbscalematrix{1}{0pt}{
    - 0.75 \Ly{\sy{W}}{0}_{0,0}
    \!+\! 0.5  \Ly{\sy{B}}{0}_0
    \!+\!      \Ly{\sy{B}}{1}_0
}
$
and
$
\idx{1}\spoint\ly{2} \eqdef \vbscalematrix{1}{0pt}{
    - 0.25 \Ly{\sy{W}}{0}_{0,0}
    \!+\! 0.5  \Ly{\sy{B}}{0}_0
    \!+\!      \Ly{\sy{B}}{1}_0
}
$.

\pref{li:Conditional-Forward-Polytope-Each} conditionally forwards each vertex
    $
        \idx{j}\sy\point\ly{L}, \idx{j}\varphi_\sigma
        \gets
        \condsy\dnn\withRef{
            \centroid\big( \polytope\ly{0} \big)
        }\bigargs{ \idx{j}\point\ly{0} }
    $
with the same reference point
    $\centroid\big( \polytope\ly{0} \big)$.
$\centroid$ is a custom function that deterministically maps
    $\polytope\ly{0}$ to a concrete point. In this example we assume
    $\centroid\big( \polytope\ly{0} \big) = \bpoint{-1.5}$.
For $\idx{0}\point\ly{0} \eqdef \bpoint{-1.5}$,
    as presented in \pref{exa:Conditional-Forward-Point},
    $
    \idx{0}\sy\point\ly{L} \eqdef \vbscalematrix{1}{0pt}{
        - 0.75 \Ly{\sy{W}}{0}_{0,0}
        \!+\! 0.5  \Ly{\sy{B}}{0}_0
        \!+\!      \Ly{\sy{B}}{1}_0
    }
    $ and $\idx{0}\varphi_\sigma$ is the first row of
    \pref{eq:Overview-Polytope-Shift-Activation}.
For $\idx{1}\point\ly{0} \eqdef \bpoint{-0.5}$
    we can compute
    $
    \idx{1}\sy\point\ly{L} \eqdef \vbscalematrix{1}{0pt}{
        - 0.25 \Ly{\sy{W}}{0}_{0,0}
        \!+\! 0.5  \Ly{\sy{B}}{0}_0
        \!+\!      \Ly{\sy{B}}{1}_0
    }
    $ in the same way,
and $\idx{1}\varphi_\sigma$ is the second row of
    \pref{eq:Overview-Polytope-Shift-Activation}.
On~\pref{li:Conditional-Forward-Polytope-Return}, $\bigl\{
    \idx{0}\spoint\ly{2},
    \idx{1}\spoint\ly{2}
\bigr\}$ and $\idx{0}\varphi_\sigma \wedge \idx{1}\varphi_\sigma$ are returned.
\end{example}

\begin{theorem}
\label{thm:Conditional-Forward-Polytope}
Let the polytope
$
\polytope\ly{0} \eqdef
    \left\{
        \idx{0}\point\ly{0},
        \idx{1}\point\ly{0},
        \ldots,
        \idx{p-1}\point\ly{0}
    \right\}
$.
Let
$
    \sy\polytope\ly{L}, \varphi
    \eqdef
        \condsy\dnn
\big(
                \polytope\ly{0}
            \big)
$
where
$
    \sy\polytope\ly{L} =
        \left\{
            \idx{0}\sy\point\ly{L},
            \idx{1}\sy\point\ly{L},
            \ldots,
            \idx{p-1}\sy\point\ly{L}
        \right\}
$.
Let $\theta \vDash \varphi$.
\begin{cs}
\item\labelInMainText{cond:Conditional-Forward-Polytope-C1}
$
        \dnn^{\theta}
        \bigl( \idx{j}\point\ly{0} \bigr)
        =
        \eval{ \idx{j}\sy\point\ly{L} }_{\theta}
    $ for $0 \leq j < p$.
\item\labelInMainText{cond:Conditional-Forward-Polytope-C2}
$\idx{j}\sy\point\ly{L} \in \sy\polytope\ly{L}$ is a linear expression and $\varphi$ is a linear formula for $0 \leq j < p$.
\item\labelInMainText{cond:Conditional-Forward-Polytope-C3}
$\theta \vDash \varphi$ implies that $\dnn^\theta$ is \locallylinear{} for
    $\polytope\ly{0}$.
\end{cs} \end{theorem}

\begin{proofsketch}
The proof of this theorem follows from \pref{thm:Conditional-Forward-Pointwise}.
\onlyfor{arxiv}{\refProof{thm:Conditional-Forward-Polytope}{proof:Conditional-Forward-Polytope}
}{}
\end{proofsketch}

The following example demonstrates $\foo$~(\pref{li:Shift-And-Assert} in \pref{alg:VPolytopeRepair}).

\begin{example}
Consider the DNN $\OverviewDNNShifted$~(\pref{fig:Overview-Polytope-ShiftedDNN})
and the set of V-polytopes $\polytopeSet_2$ from~\pref{sec:Overview-Polytope}. We
will show
$
    \OverviewDNNPolytopeRepaired
    =
    \foo\bigl(
        \OverviewDNNShifted,
        \OverviewDNNShifted,
        \polytopeSet_2,
        \spec_2,
        k
    \bigr)
$
where $\OverviewDNNPolytopeRepaired$ is shown in~\pref{fig:Overview-Polytope-RepairedDNN} and
$k \eqdef 1$.

\begin{steps}
\item\label{step:ShiftAndRepair-S1}
    Lines
    \ref{li:ShiftAndRepair-ForEachPolytope-Begin}--\ref{li:ShiftAndRepair-ForEachPolytope-End}
    conditionally forward executes each input V-polytope
    $\polytope\ly{k} \in \SliceDNN{\dnn}{0}{k}\bigargs{\polytopeSet\ly{0}}$
    on $\SliceDNN{\condsy\dnn}{k}{L}$.
In this example,
    $\sy\polytope\ly{2}, \varphi_\sigma
        \gets \SliceDNN{\condsy\OverviewDNNShifted}{1}{2}
            \big(
                \polytope\ly{1}
            \big)
    $ for each $\polytope\ly{1} \in \SliceDNN{\OverviewDNNShifted}{0}{1}\bigargs{\polytopeSet_2}$.

\item
    \pref{li:ShiftAndRepair-Spec} adds the repair specification
    $\spec\bigargs{\sy\polytopeSet\ly{L}}$ to $\varphi$.
    In this example, $\spec_2\bigargs{\sy\polytopeSet\ly{L}}$ is added to $\varphi$.

\item
    Lines \ref{li:ShiftAndRepair-Delta}--\ref{li:ShiftAndRepair-Minimize} solves
    the LP problem.
    \pref{li:ShiftAndRepair-Delta} constructs the delta vector
    $\vec\Delta$ that consists of both the functional difference
    $\sy\polytopeSet\ly{L} - \dnn_\OriginalKW\bigargs{\polytopeSet\ly{0}}$ and
    the parameter difference $\sy\theta_{\SliceDNN{\dnn}{k}{L}} -
    \theta_{\SliceDNN{\dnn}{k}{L}}$.
\pref{li:ShiftAndRepair-Minimize} calls an LP solver to find
    new parameters $\theta'_{\SliceDNN{\dnn}{k}{L}}$ that satisfy $\varphi$ while
    minimizing $\vecNorm{\vec\Delta}{p}$.
    In this example, $\dnn\eqdef\OverviewDNNShifted$,
    $\dnn_\OriginalKW\eqdef\OverviewDNNShifted$,
    $\polytopeSet\ly{0}\eqdef\polytopeSet_2$, $k\eqdef 1$ and $L\eqdef 2$.

\item
\pref{li:ShiftAndRepair-Infeasible} returns $\bot$ if $\varphi$ is
    unsatisfiable.
Otherwise \pref{li:ShiftAndRepair-Update} updates
    $\SliceDNN{{\dnn}_\texttt{ret}}{1}{2}$ with the new parameter
    $\theta'_{\SliceDNN{\dnn}{k}{L}}$ where ${\dnn}_\texttt{ret}$
    is a copy of $\dnn\eqdef\OverviewDNNShifted$.
    Finally \pref{li:ShiftAndRepair-Return}
    returns the repaired $\OverviewDNNPolytopeRepaired\eqdef{\dnn}_\texttt{ret}$.

\end{steps}
\end{example}

\begin{theorem}
\label{thm:Shift-And-Assert}
Let
$
    \dnn_\texttt{ret}
    \eqdef
    \foo\bigl(
        \dnn,
        \dnn^\OriginalKW,
        \polytopeSet\ly{0},
        \spec,
        k
    \bigr)
$, $\dnn_\texttt{ret} \neq \bot$
and $\SliceDNN{\dnn}{0}{k}$ is \locallylinear{} for any polytope
$\polytope\ly{0} \in \polytopeSet\ly{0}$.
Then:
\begin{cs}
\item\labelInMainText{cond:Shift-And-Assert-C1}
$\dnn_\texttt{ret}$ is \locallylinear{}
    for any polytope $\polytope\ly{0} \in \polytopeSet\ly{0}$.
\item\labelInMainText{cond:Shift-And-Assert-C2}
$\dnn_\texttt{ret}$ satisfies the polytope specification
    $\bigl(
        \polytopeSet\ly{0}, \spec
    \bigr)$.
\item\labelInMainText{cond:Shift-And-Assert-C3}
$\foo$ runs in polynomial time in the total number of vertices
in $\polytopeSet\ly{0}$ and the size of $\dnn$.
\end{cs}
 \end{theorem}

\begin{proofsketch}
Given $\SliceDNN{\dnn}{0}{k}$ is \locallylinear\ for $\polytopeSet\ly{0}$
    and $\SliceDNN{\dnn_\texttt{ret}}{0}{k}$ is the same as $\SliceDNN{\dnn}{0}{k}$.
    Because $\dnn_\texttt{ret} \neq \bot$, $\theta'_{\SliceDNN{\dnn}{k}{L}}$ satisfies $\varphi$.
    Hence $\SliceDNN{\dnn_\texttt{ret}}{k}{L}$ with the new parameters $\theta'_{\SliceDNN{\dnn}{k}{L}}$
    is \locallylinear{} for any polytope
    $\polytope\ly{k} \in \SliceDNN{\dnn_\texttt{ret}}{0}{k}\bigargs{\polytopeSet\ly{0}}$
    using~\pref{thm:Conditional-Forward-Polytope}-\ref{cond:Conditional-Forward-Polytope-C1}.
    Thus, we have $\dnn_\texttt{ret}$ is \locallylinear\ for any polytope
    $\polytope\ly{0} \in \polytopeSet\ly{0}$, proving \pref{cond:Shift-And-Assert-C1}.

    Consider any polytope $\polytope\ly{0} \in \polytopeSet\ly{0}$
    and its corresponding specification $\psi\in\spec$,
    $\theta'_{\SliceDNN{\dnn}{k}{L}} \vDash \varphi$ implies that
    $\dnn_\texttt{ret}$ satisfies $\psi$ on all vertices of $\polytope\ly{0}$.
Because $\psi$ is a linear formula and $\dnn_\texttt{ret}$ is \locallylinear\ for
    $\polytope\ly{0}$, $\dnn_\texttt{ret}$ satisfies $\psi$ on all points in
    $\convexhull\bigargs{\polytope\ly{0}}$.
Thus, $\dnn_\texttt{ret}$ satisfies the polytope specification
    $\bigl(
        \polytopeSet\ly{0}, \spec
    \bigr)$, proving \pref{cond:Shift-And-Assert-C2}.

\pref{cond:Shift-And-Assert-C3}
    is true because the size of linear formula $\varphi$ is polynomial in the
    total number of vertices in $\polytopeSet\ly{0}$ and the size of $\dnn$.
\onlyfor{arxiv}{\refProof{thm:Shift-And-Assert}{proof:Shift-And-Assert}}{}
\end{proofsketch}

The following example demonstrates $\PolyRepair$ defined in~\pref{alg:VPolytopeRepair}.

\begin{definition}
\label{def:Network-Partition}
    Given a DNN $\dnn$ of $L$ layers, a layer index $k$, and input V-polytopes
    $\polytopeSet\ly{0}$. A network partition $s$, which is a list of network
    index tuples
    $
        [(k_0, l_0), (k_1, l_1), \ldots, (k_{n-1}, l_{n-1})]
    $, is valid for
    $\dnn$, $k$ and $\polytopeSet\ly{0}$ if either $s$ is empty and
    $\SliceDNN{\dnn}{0}{k}$ is \locallylinear\ for $\polytopeSet\ly{0}$,
    or $s$ satisfies:
    \begin{enumerate*}[label=(\alph*)]
    \item for any tuple $(k_i, l_i)$ in $s$, $0 \leq k_i < l_i \leq L$;
    \item for the first tuple $(k_0, l_0)$ in $s$, $\SliceDNN{\dnn}{0}{k_0}$ is
    \locallylinear\ for $\polytopeSet\ly{0}$;
    \item for any two consecutive tuples $(k_{i}, l_{i})$
        and $(k_{i+1}, l_{i+1})$ in $s$,
        $k_{i+1} \leq l_{i}$;
    \item For the last tuple $(k_{n-1}, l_{n-1})$ in $s$,
        $k \leq l_{n-1}$.
    \end{enumerate*}
\end{definition}

\begin{example}
\label{exa:PolyRepair}
For the V-polytope repair in~\pref{sec:Overview-Polytope},
$
    \OverviewDNNPolytopeRepaired
    =
    \PolyRepair\bigl(
        \OverviewDNN,
        \polytopeSet\ly{0}_2,
        \spec_2,
        s,
        k
    \bigr)
$
where $s = [(0,1)]$ and $k = 1$.

\begin{steps}
\item\label{step:PolyRepair-Shift}
    \pref{li:PolyRepair-Copy-Network} makes a copy $\dnn_\texttt{ret}$ of $\dnn$ for repair.
    Lines \ref{li:PolyRepair-ForEachSlice-Begin}--\ref{li:PolyRepair-ForEachSlice-End}
    shift $\dnn_\texttt{ret}$ such that $\SliceDNN{\dnn_\texttt{ret}}{0}{k}$ is
    \locallylinear\ for $\polytopeSet\ly{0}$.
Specifically,
    for each $(k_i, \ell_i)$ pair in the network partition $s$,
    \pref{li:PolyRepair-ShiftAndAssert}
    requires $\SliceDNN{\dnn_\texttt{ret}}{0}{k_i}$ to be
    \locallylinear\ for $\polytopeSet\ly{0}$ and
    shifts $\SliceDNN{\dnn_\texttt{ret}}{k_i}{\ell_i}$
    with an empty specification $\top$
    such that $\SliceDNN{\dnn_\texttt{ret}}{0}{\ell_i}$ is
    \locallylinear\ for $\polytopeSet\ly{0}$.
    The original $\dnn$ is used to minimize the functional difference.
\pref{li:PolyRepair-ShiftLinearRegion-Infeasible} returns
    $\bot$ if $\foo$ fails.
In this example, after the pair $(0, 1) \in s$,
    $\dnn_\texttt{ret}$ is $\OverviewDNNShifted$
    shown in~\pref{fig:Overview-Polytope-ShiftedDNN}.

\item\label{step:PolyRepair-Spec}
\pref{li:PolyRepair-Spec}
    requires $\SliceDNN{\dnn_\texttt{ret}}{0}{k}$ to be
    \locallylinear\ for $\polytopeSet\ly{0}$,
    calls $\foo\bigargs{
        \dnn_\texttt{ret},
        \dnn,
        \polytopeSet\ly{0},
        \spec,
        k
    }$ to repair the shifted DNN $\dnn_\texttt{ret}$
    against the V-polytope specification $(\polytopeSet\ly{0}, \spec)$
    and returns its result.
In this example,
    $\foo\bigargs{
        \OverviewDNNShifted,
        \OverviewDNN,
        \polytopeSet_2\ly{0},
        \spec_2,
        1
    }$
    returns $\OverviewDNNPolytopeRepaired$
    shown in \pref{fig:Overview-Polytope-RepairedDNN}.
\end{steps}
\end{example}

\begin{theorem}
\label{thm:VPolytopeRepair}
    Let
    $
        \dnn_\texttt{ret}
        \eqdef
        \PolyRepair\bigl(
            \dnn,
            \polytopeSet\ly{0},
            \spec,
            s,
            k
        \bigr)
    $,
    $\dnn_\texttt{ret} \neq \bot$,
    $s$ is a valid network partition for $\dnn$ and $k$.
    $\PolyRepair$ satisfies the following properties:
\begin{cs}
\item\labelInMainText{cond:PolyRepair-C1}
    $\dnn_\texttt{ret}$ satisfies the V-polytope specification
        $\bigl(
            \polytopeSet\ly{0}, \spec
        \bigr)$.
\item\labelInMainText{cond:PolyRepair-C2}
    $\PolyRepair$ runs in polynomial time in the number of vertices in $\polytopeSet\ly{0}$, size of $\dnn$ and~$s$.
\end{cs}
\end{theorem}

\begin{proof}
Because \pref{step:PolyRepair-Shift} makes
    $\SliceDNN{\dnn_\texttt{ret}}{0}{k}$ locally linear for $\polytopeSet\ly{0}$
    (using~\pref{thm:Shift-And-Assert}-\ref{cond:Shift-And-Assert-C1} and \pref{def:Network-Partition}),
\pref{step:PolyRepair-Spec} returns a DNN $\dnn_\texttt{ret}$ that satisfies the
    V-polytope repair
    specification if $\dnn_\texttt{ret} \neq \bot$~(using~\pref{thm:Shift-And-Assert}-\ref{cond:Shift-And-Assert-C2}), proving
    \pref{cond:PolyRepair-C1}.
\pref{cond:PolyRepair-C2} follows from
\pref{thm:Shift-And-Assert}-\ref{cond:Shift-And-Assert-C3}
    as $\PolyRepair$ calls $\foo$ for each partition in $s$ and once
    on~\pref{li:PolyRepair-Spec}.
\end{proof}

\section{Qualitative Comparison with Prior Approaches}
\label{sec:Comparison}

In this section, we present a qualitative comparison between \toolname{} (this
work) and prior provable-repair techniques,
PRDNN~\cite{DBLP:conf/pldi/SotoudehT21} and REASSURE~\cite{DBLP:conf/iclr/Fu22}.

\subsubsubsection{Pointwise repair.}
PRDNN and REASSURE are not architecture-preserving for
provable pointwise repair, unlike \toolname{}.
PRDNN introduces the notion of a \emph{Decoupled} DNN, which decouples
the original DNN into an activation network and a value network.
The approach then reduces repairing a single layer in the value network to solving an LP problem.

\newcommand{\area}{\mathcal{A}}

REASSURE reduces repairing a point $X$ to repairing the linear region $\area$
containing that point. For a linear region $\area$, REASSURE adds a patch
network $p_\area$ such that $\dnn(\point)+p_\area(\point)$ for $\point \in
\area$ satisfies the repair specification. To localize changes to $\area$ while
keeping the repaired network $\dnn'$ continuous, REASSURE constructs a support
network $g_\area$ such that $\dnn'(\point)$ equals to
$\dnn(\point)+p_\area(\point)$ for $\point \in \area$ and smoothly goes to the
original output $\dnn(\point)$ for $\point$ outside $\area$.

\subsubsubsection{Polytope repair.}
\toolname{} supports arbitrary V-polytope repair specifications.
PRDNN and REASSURE need to enumerate
linear regions, unlike \toolname{}.
To ensure scalability, PRDNN only supports V-polytopes in 2D subspaces of the input space.
It uses SyReNN~\cite{TACAS2021,DBLP:journals/sttt/SotoudehTT23} to efficiently enumerate the linear regions,
and then performs pointwise repair on the vertices of these linear regions to
guarantee provable V-polytope repair.

REASSURE reduces the repair for a polytope to
the repair for all buggy linear regions in the polytope.
However, REASSURE cannot be used in practice for V-polytope or H-polytope repair
because a polytope has a tremendous number of linear regions, which grows
exponentially in the network depth. Even just enumerating and finding all buggy
linear regions is impractical, and the REASSURE tool does not provide a way to
do so. Moreover, even if one could provide REASSURE with all buggy linear
regions, the repaired network produced by REASSURE has large runtime and memory
overhead (as demonstrated later in this section and ~\pref{sec:Experiment-Overhead}).

\subsubsubsection{Soundness.}
Both \toolname{} and PRDNN are sound; that is, if a repaired DNN is generated
by \toolname{} or PRDNN, then it is guaranteed to satisfy the repair
specification.

REASSURE is unsound for pointwise repair when two points in the same linear
region have conflicting specifications: REASSURE is unable to identify
this conflict, and the tool will return an incorrect DNN instead of reporting
that the repair is infeasible for REASSURE.

REASSURE is unsound for polytope repair in at least two cases. Consider two
input polytopes with different specifications. For any buggy linear that
intersects with both input polytopes, REASSURE will generate an incorrect repair
by adding two overlapping patch networks of the two specifications separately for such
linear region. Moreover, for two adjacent buggy linear regions with different
specifications, the behavior on their common boundary may violate both specifications due to the
support networks.

\subsubsubsection{Completeness for pointwise repair.}
\toolname{}, PRDNN and REASSURE are complete for single-point repair; that is,
they are guaranteed to generate a repaired DNN that satisfies the single-point repair specification.
However, they are incomplete for general pointwise repair.
Given a DNN $\dnn$, consider a pointwise repair specification with three points
in the same linear region of $\dnn$, but with corresponding outputs
constrained to be non-collinear;
e.g., DNN $\OverviewDNN$
in~\pref{fig:Overview-DNN} and the pointwise specification
$
    \OverviewDNN\bigargs{\bpoint{2.5}}_0\leq 0.4 \wedge
    \OverviewDNN\bigargs{\bpoint{3}}_0 \geq 0.5 \wedge
    \OverviewDNN\bigargs{\bpoint{3.5}}_0 \leq 0.5
$.
Although a repair exists, PRDNN fails because the decoupled activation network
keeps buggy points in their original linear region.
REASSURE also fails because it is impossible to repair the same linear region for
three buggy points with conflicting constraints. The REASSURE implementation
fails to detect such a conflict and gives an incorrect repair.
\toolname{} will also fail if the reference points keep the three points in the
same linear region. However, \toolname{} provides the flexibility to try
different reference points such that not all three points are in the same linear
region in the repaired DNN.
For example, with reference points
$
X\ly{0}_\RefKW = \vbscalematrix{1}{2pt}{1.5 & 3 & 3.5}
$,
\toolname{} finds a repair.

\subsubsubsection{Drawdown and generalization.}
\toolname{} and PRDNN both minimize changes in the parameters and functional
difference as a proxy for minimizing drawdown. REASSURE guarantees that changes
are localized to the repaired linear regions. In practice, \toolname{} and PRDNN
have much higher generalization with REASSURE exhibiting 0\% generalization (see
\pref{sec:Experiment-pointwise-mnistc}).

\subsubsubsection{Activation function.}
\toolname{} supports activation functions that have linear pieces.
PRDNN theoretically supports any activation function for pointwise repair and
piecewise-linear functions for polytope repair, but the implementation only
supports piecewise-linear functions.
REASSURE only supports piecewise-linear functions.

\subsubsubsection{Network-size overhead.}
Being architecture preserving, \toolname{} does not have any network size overhead.
Because PRDNN converts the given DNN into a Decoupled DNN, the repaired network size
doubles in a naive implementation. An optimized implementation could only
store the original DNN and the single repaired value layer. The network-size overhead for PRDNN
does not depend on the given repair specification.

In contrast, REASSURE has a significant overhead in network size even when repairing
a single linear region, and the network-size overhead depends on the number of linear regions
being repaired. Consider a
fully-connected ReLU network of $L$ layers.
To repair one linear region, REASSURE adds a patch network $p$,
a support network $g$ and a scalar parameter of the $\norm{p(\point)}_{\infty}$.
A patch network consists of a $n\ly{0}\times n\ly{L}$ fully-connected layer with
$n\ly{0}\times n\ly{L} + n\ly{L}$ parameters.
A support network consists of a $n\ly{0}\times n\ly{hidden}$ fully-connected
layer with $n\ly{0}\times n\ly{hidden} + n\ly{hidden}$ parameters where
$n\ly{hidden} \eqdef \sum_{\ell=1}^{L-1} n\ly{\ell}$ denotes the number of all
hidden neurons.
Therefore, \emph{for each linear region},
the REASSURE repaired network adds $
(n\ly{0}+1)\times (n\ly{L} + \sum_{\ell=1}^{L-1} n\ly{\ell} ) + 1$
new parameters.
For example, the MNIST $3\times 100$ network
from~\pref{sec:Experiment-pointwise-mnistc} has 89,610 parameter,
and REASSURE will add 164,851 \emph{new} parameters \emph{for each buggy linear region}.

\section{Implementation}
\label{sec:Implementation}

\newcommand{\prdnn}{\text{PRDNN}}
\newcommand{\python}{\text{Python}}
\newcommand{\pytorch}{\text{PyTorch}}
\newcommand{\numpy}{\text{NumPy}}
\newcommand{\gurobi}{\text{Gurobi}}
\newcommand{\squeezenet}{\text{SqueezeNet}}

\toolname{} is built using \pytorch{}~\cite{pytorch}, an open source machine learning
framework, and \gurobi{}~\cite{gurobi}, a mathematical optimization solver for linear
programming (LP), quadratic programming (QP) and mixed integer programming (MIP)
problems. The code is available at \url{https://github.com/95616ARG/APRNN}.

To measure parity with \toolname{}, we also re-implemented
PRDNN~\cite{DBLP:conf/pldi/SotoudehT21} in our tool. Our tool provides an easier
interface with which to repair a network using PRDNN, and allows for a fair
comparison between the two algorithms by ensuring that the same repair
specifications and networks are used. 
Compared to the original implementation of
PRDNN~\cite{PRDNN}, the use of PyTorch decreases the time spent on the
Jacobian computation. Further optimizations were achieved by making
our PRDNN implementation GPU compatible.

We also implemented lookup-based override approaches for pointwise and V-polytope
repair of classification DNNs. (a)~For pointwise repair, we implemented
a lookup-table approach that uses a hash-table
of pointwise repair specification to override the output of the DNN.
(b)~For the V-polytope repair,
we implemented a lookup-function approach that uses an LP solver to find inputs in the
V-polytope repair specification and overrides output of the DNN for such inputs.

\section{Experimental Evaluation}
\label{sec:ExperimentalEvalation}

All experiments were run on a machine with
dual 16-core Intel Xeon Silver 4216 CPUs,
384 GB of memory,
SSD and
a NVIDIA RTX A6000 with 48 GB of GPU memory.

\begin{table}[t]
    \centering
    \caption{
Provable pointwise repair of MNIST networks
        using \toolname{} (this work), PRDNN, REASSURE and lookup-table (LT).
        k: the pointwise repair parameter for PRDNN and \toolname{},
        D: drawdown,
        G: generalization,
        T: repair time.
\toolname{} uses empty $s=[]$ for all repairs.
    }
    \label{tab:MnistRepair}
    {\small
    \setlength{\tabcolsep}{3pt}
    \begin{tabular}{@{}cccccccccccccccccc@{}}
        \toprule
        \multicolumn{1}{c}{\multirow{2}{*}{Network$\;$}}
                        & \multicolumn{4}{c}{\toolname{}} &
                        & \multicolumn{4}{c}{PRDNN} &
                        & \multicolumn{3}{c}{REASSURE} &
                        & \multicolumn{3}{c}{LT} \\
                        \cmidrule{2-5}
                        \cmidrule{7-10}
                        \cmidrule{12-14}
                        \cmidrule{16-18}
                        & k & D & G & T &
                        & k & D & G & T &
                            & D & G & T &
                            & D & G & T  \\
        \midrule
        $3\times 100$
        & 1 & 1.28\% & \textbf{31.53\%} & 5s &
            & 0 & 2.13\% & 23.34\% & 187s &
            & \textbf{0\%} & 0\% & 490s &
            & \textbf{0\%} & 0\% & \textbf{<1s}
        \\
$9\times 100$
        & 5 & 2.72\% & \textbf{19.13\%}  & 92s &
            & 5 & 9.55\% &  9.18\% & 11s &
            & \textbf{0\%} & 0\% & 1503s &
            & \textbf{0\%} & 0\% & \textbf{<1s}
        \\
$9\times 200$
            & 6 & 1.38\% & \textbf{24.69\%} & 372s &
            & 1 & 3.92\% &  6.55\% & 15s  &
            & \textbf{0\%} & 0\% & 15356s &
            & \textbf{0\%} & 0\% & \textbf{<1s}
        \\
        \bottomrule
    \end{tabular}
    }
\end{table} 
\subsection{Pointwise MNIST Image Corruption Repair}
\label{sec:Experiment-pointwise-mnistc}

\subsubsubsection{Buggy networks.}
We repair MNIST fully-connected DNNs with ReLU activation layers
from~\cite{ERAN} with a varying number of layers and
parameters for classifying handwritten digits. We refer to the MNIST networks by
the number of layers and the width of hidden layers.
For example, the ``$3\times 100$'' network has
3 layers with 100 neurons per hidden layer
($784\times 100\times 100\times 10$).

\subsubsubsection{Pointwise repair specification.}
The MNIST-C dataset~\cite{DBLP:journals/corr/abs-1906-02337} consists of
images from the official MNIST test set that have been corrupted.
The pointwise
repair specification consists of the first $100$ fog-corrupted images from
MNIST-C. Buggy networks have 5\%-15\% accuracy on it.

\subsubsubsection{Generalization set.}
9,000 images from MNIST-C's foggy test set, disjoint from the repair set.
The accuracy of the buggy DNNs on this
generalization set is under 20\%.

\subsubsubsection{Drawdown set.}
The drawdown set is the official MNIST test set containing 10,000 images.

\subsubsubsection{Results.}
All tools
were able to repair all networks so they satisfied the pointwise repair specification.
\pref{tab:MnistRepair} summarizes the results.
Both REASSURE and lookup-table have zero drawdown but zero generalization,
while both \toolname{} and PRDNN have some drawdown but good generalization.
\toolname{} has good (low) drawdown and the best (highest) generalization
on all networks.
PRDNN also has good generalization, but the drawdown is worse (higher) than \toolname{}.

Regarding repair time, the lookup-table finished instantly because it
just caches the repair specification for lookup during inference.
Both PRDNN and \toolname{} took a short amount of time, while
PRDNN was faster than \toolname{} for two of the networks.
By contrast,
REASSURE took the longest time for all networks, and the time increases
significantly as the network size increases. For the $9\times 200$ network,
REASSURE took 4 hours and 16 minutes.
Moreover, the REASSURE repaired networks have significant runtime overhead,
which is evaluated in \pref{sec:Experiment-Overhead}.

This experiment demonstrates that \toolname{} is overall the best pointwise
repair approach while preserving the original DNN architecture, because
\toolname{} can efficiently repair a buggy network while maintaining good drawdown
and achieving the best generalization.

\subsection{Pointwise ImageNet Natural Adversarial Examples and Corruption Repair}
\label{sec:Experiment-pointwise-imagenet}

\subsubsubsection{Buggy networks.}
We repair modern ImageNet CNNs ResNet152~\cite{DBLP:conf/cvpr/HeZRS16} (60.2
million parameters) and VGG19~\cite{DBLP:journals/corr/SimonyanZ14a} (143.7 million parameters). ResNet152 has
78.312\% top-1 accuracy and 94.046\% top-5 accuracy. VGG19 has 72.376\% top-1
accuracy and 90.876\% top-5 accuracy.

\subsubsubsection{Pointwise repair specification.}
We consider two pointwise repair specifications from the following two datasets:
1) the Natural Adversarial Examples~(NAE)
dataset~\cite{DBLP:journals/corr/abs-1907-07174} that consists of 7,500 images
which are commonly misclassified by modern ImageNet networks;
2)~the ImageNet-C dataset~\cite{DBLP:conf/iclr/HendrycksD19} that consists of
algorithmically generated corruptions applied to the ImageNet validation set
with severity 1 to 5.
For each network,
1) the pointwise repair specifications for the NAE dataset
consist of the first 50 NAE images that were misclassified by that network;
2) the pointwise repair specifications for the ImageNet-C dataset consists of
the first 50 fog-corrupted images of severity 3 that were misclassified by
that network.

\subsubsubsection{Generalization set.}
There is no generalization set for the NAE dataset because the images
do not have common features to generalize.
For the ImageNet-C dataset, it is reasonable to expect that the repair of one
corrupted image generalizes to the same image with other severities of the same
corruption. Thus, we take the same subset of fog-corrupted images as the
corresponding repair specification but with different severities 1, 2, 4 and 5 as the
generalization set.

\subsubsubsection{Drawdown set.} We use the ILSVRC2012 ImageNet validation set of
50,000 images~\cite{imagenet} as the drawdown set. It is the standard dataset for evaluating the accuracy of ImageNet~networks.

\subsubsubsection{Repair layer.}
Both \toolname{} and PRDNN repair a convolutional layer in the last bottleneck
block of ResNet152 and a fully-connected layer in the last classifier block of VGG19.

\subsubsubsection{Results.}
For both NAE and ImageNet-C pointwise repair specifications, \toolname{}
successfully repairs both ResNet152 and VGG19, while PRDNN runs out of memory
during repair.
For NAE, \toolname{} took 2,935 seconds to repair ResNet152 and 3,671 seconds to
repair VGG19. The repaired ResNet152 has 1.33\% top-1 and 0.66\% top-5 drawdown,
and the repaired VGG19 has 0.72\% top-1 and 0.33\% top-5 drawdown.
For ImageNet-C, \toolname{} took 2,666 seconds to repair ResNet152 and 1,918
seconds to repair VGG19.
The repaired ResNet152 has 1.32\% top-1 and 0.56\% top-5 drawdown, as well as
19.00\% top-1 and 11.50\% top-5 generalization;
the repaired VGG19 has 0.02\% top-1 and 0.00\% top-5 drawdown, as well as
23.00\% top-1 and 20.00\% top-5 generalization.
This experiment highlights \toolname{}'s ability to scale to very large networks
with low drawdown and high generalization.

\subsection{V-Polytope MNIST Image Corruption And Rotation Repair}
\label{sec:Experiment-MNIST-Poly}

\begin{figure}[t]
    \centering
    \begin{subfigure}{\textwidth}
        \centering
        \includegraphics[width=\textwidth]{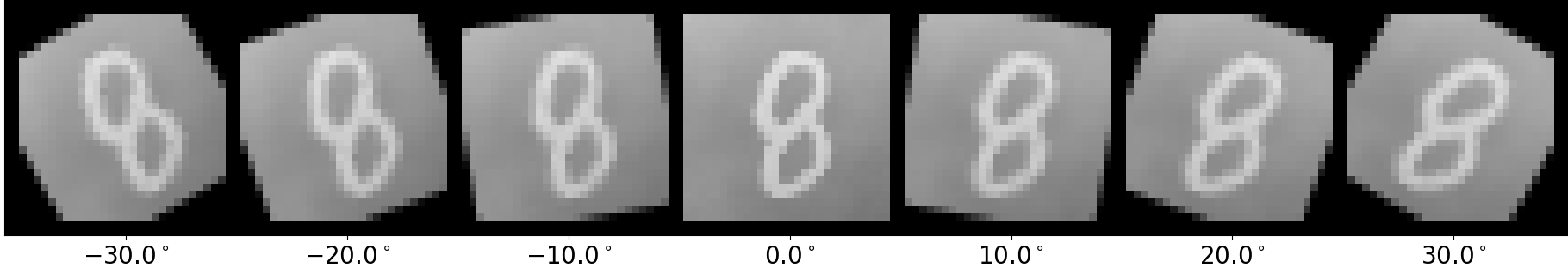}
\caption{Images (vertices) of a V-polytope $\polytope$ from the repair set.}
        \label{fig:Eval-MNIST-VPoly-RepairSet}
    \end{subfigure}
    \\
    \begin{subfigure}[t]{0.29\textwidth}
        \centering
        \includegraphics[width=\textwidth]{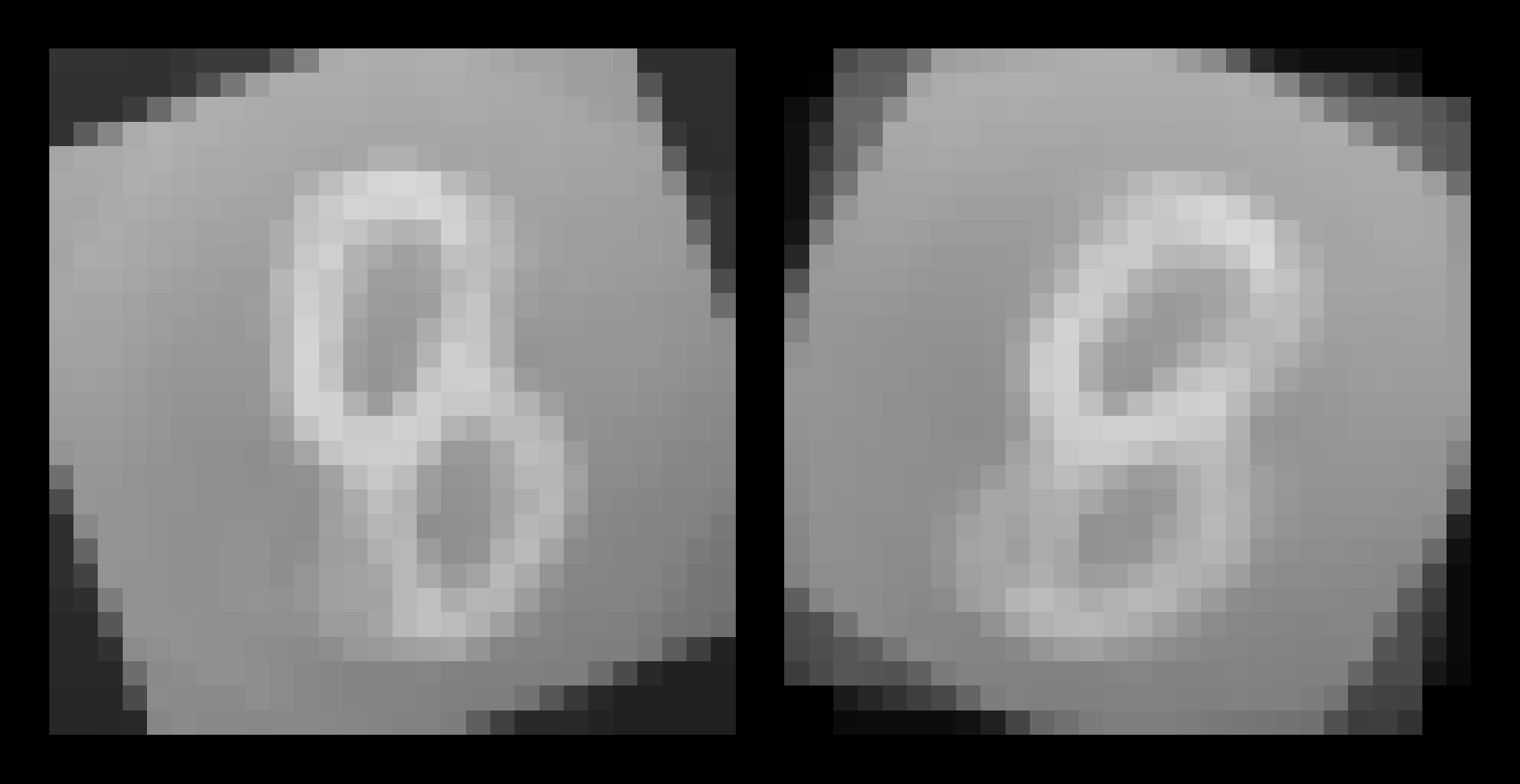}
        \caption{Images in $\convexhull(\polytope)$.}
        \label{fig:Eval-MNIST-VPoly-RepairSet-Inner}
    \end{subfigure}
    \hfill
    \begin{subfigure}[t]{0.29\textwidth}
        \centering
        \includegraphics[width=\textwidth]{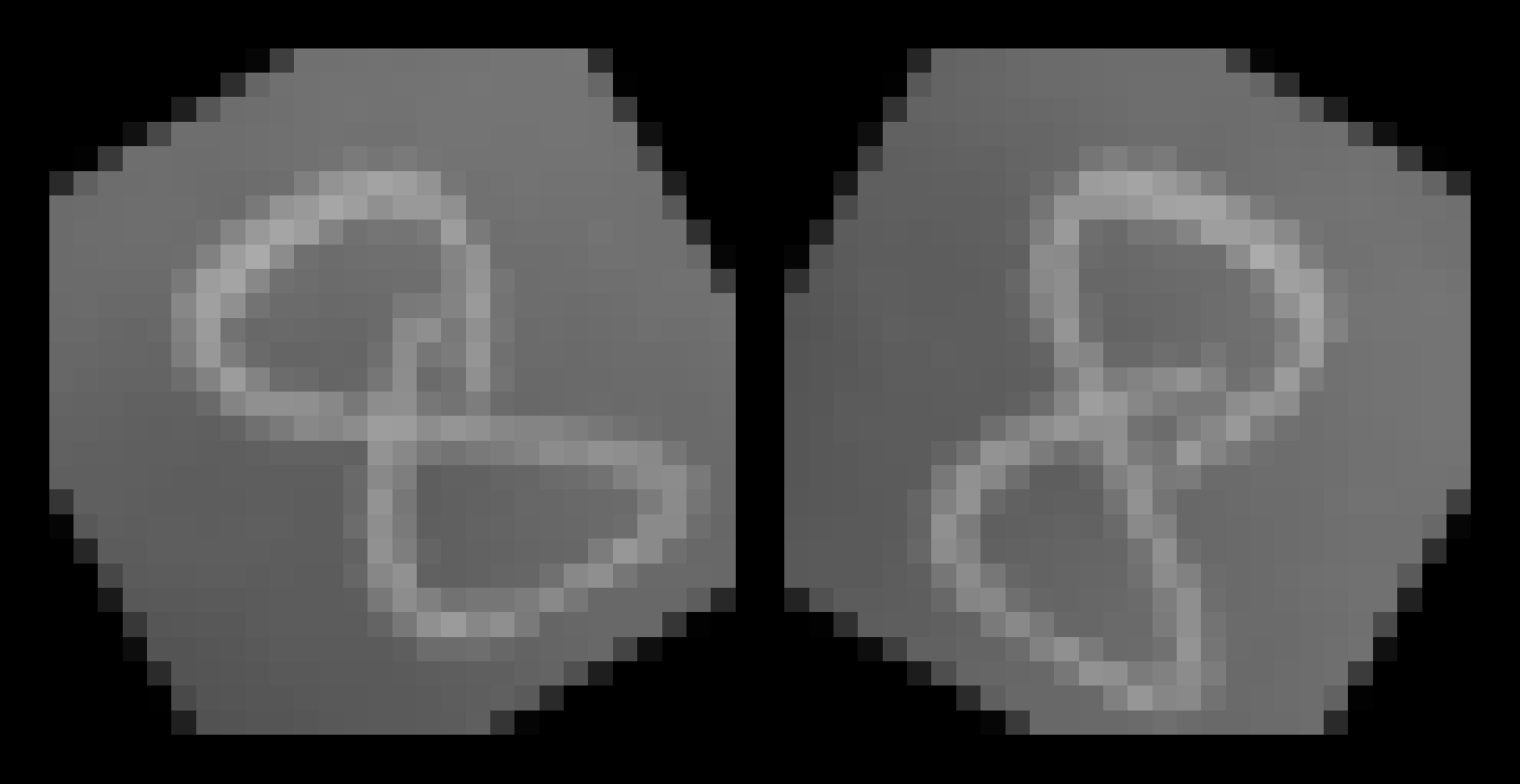}
        \caption{Images from~$S_1$.}
        \label{fig:Eval-MNIST-VPoly-GenSet-I}
    \end{subfigure}
    \hfill
    \begin{subfigure}[t]{0.29\textwidth}
        \centering
        \includegraphics[width=\textwidth]{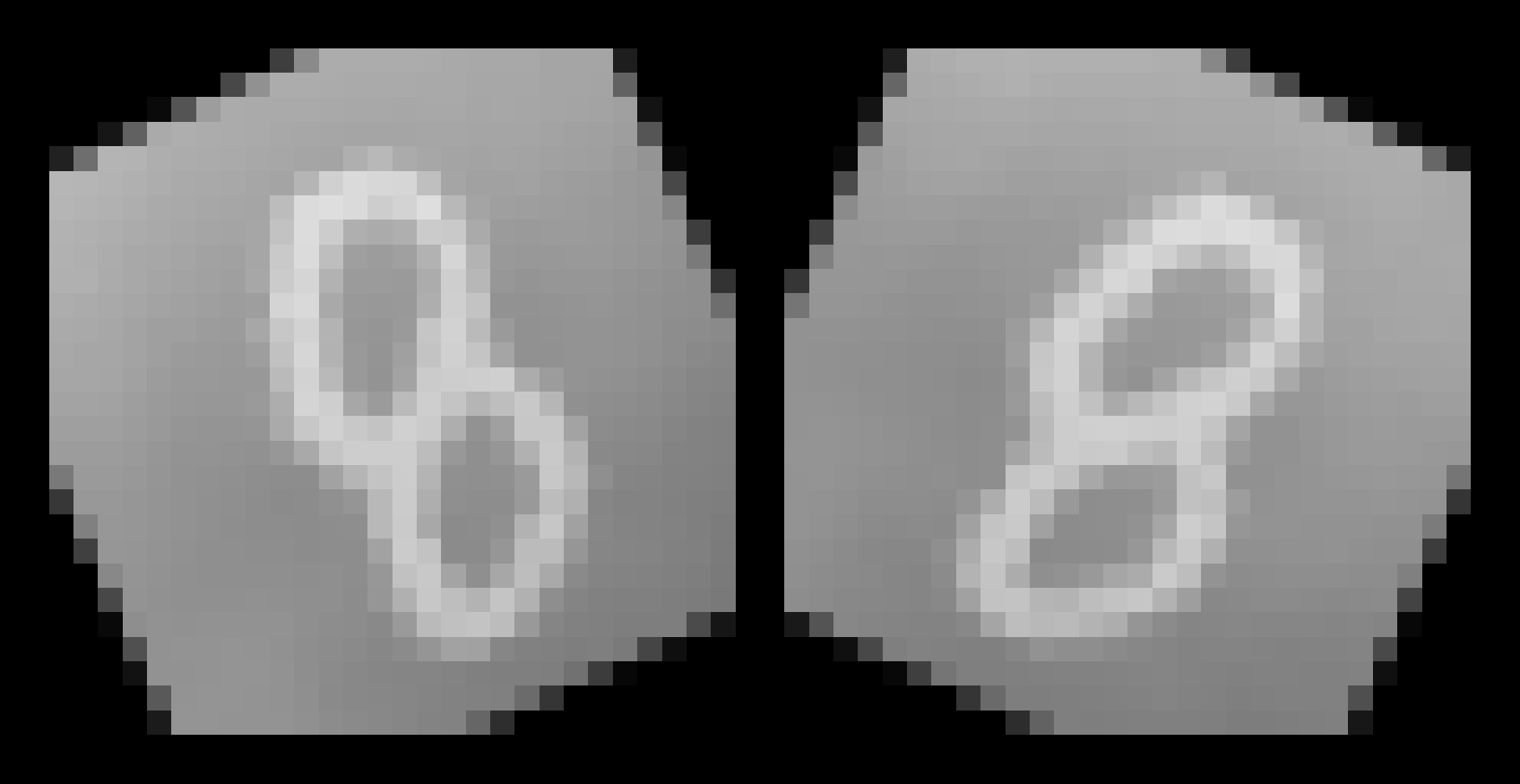}
        \caption{Images from~$S_2$.}
        \label{fig:Eval-MNIST-VPoly-GenSet-II}
    \end{subfigure}
    \pldivspace{-1.5ex}
    \caption{Images illustrating repair specification and generalization set
    for provable V-polytope MNIST repair. }
\end{figure}
 
\subsubsubsection{Buggy network.} In this experiment, we repair the MNIST
$9\times 100$ DNN from~\cite{ERAN}.

\subsubsubsection{V-polytope repair specification for \toolname{} and lookup-function.}
The V-polytope specification we use consists of $n$ V-polytopes defined by $m$
images along with the constraint that all images in the convex hull of this
polytope have the same (correct) classification. Specifically, we take the first
five foggy images with label ``8'' from the MNIST-C dataset that are
misclassified by the network. We rotate each image $-30^\circ$, $-20^\circ$,
$-10^\circ$, $0^\circ$, $10^\circ$, $20^\circ$, $30^\circ$
in pixel space.
The V-polytope
repair specification consists of the five 784-dimensional V-polytopes of the
convex hull of these seven rotated images with constraint that all images
within each V-polytope are classified as ``8''.
\pref{fig:Eval-MNIST-VPoly-RepairSet} shows images (vertices) of a V-polytope
$P$ in our repair specification.
\pref{fig:Eval-MNIST-VPoly-RepairSet-Inner} shows two images in $\convexhull( P )$.

\subsubsubsection{Approximate 2D V-polytope repair specification for PRDNN and REASSURE.}
As discussed in~\pref{sec:Comparison}, both PRDNN and REASSURE do not scale to
higher-dimensional V-polytopes.
Thus, we sample 2D linear regions from the true V-polytope repair specification
to form an approximate V-polytope repair specification for PRDNN and REASSURE.
We use SyReNN to sample 3,187 2D linear regions (of 12,394 unique
vertices) from the 2D triangles among vertices of each V-polytope.

\subsubsubsection{Approximate pointwise repair specification for REASSURE.}
Because REASSURE does not scale to repair a large number of linear regions (as
discussed in~\pref{sec:Comparison} and shown in
\pref{sec:Experiment-pointwise-mnistc}), we sample a smaller
pointwise repair specification for it. This repair specification consists of 500
unique buggy linear regions sampled from the V-polytope repair specification,
represented by 500 points in those linear regions. As a comparison, we also
evaluate APRNN pointwise repair on this specification.

\subsubsubsection{Generalization sets.}
We consider two generalization sets:
(a) the generalization set~$S_1$ consists of foggy misclassified images of label
``8'' (disjoint from those included in the repair set) rotated using the same
seven rotations used to construct the repair specification;
(b)~the generalization set~$S_2$ consists of the same five foggy misclassified
images of label ``8'' that were chosen to form the repair specification, but
rotated by angles ranging from $-30^\circ$ to $30^\circ$ with a step of
$1^\circ$.
The buggy network has 36.53\% accuracy on $S_1$ and 35.77\% accuracy on $S_2$.
\pref{fig:Eval-MNIST-VPoly-GenSet-I} shows two images from~$S_1$ and
\pref{fig:Eval-MNIST-VPoly-GenSet-II} shows two images from~$S_2$.

\subsubsubsection{Drawdown set.}
The drawdown set is the official MNIST test set containing 10,000 images.

\subsubsubsection{Results.}
For the (true) V-polytope repair specification,
only \toolname{} and the lookup-function can provably repair it.
\toolname{} took 59 seconds with parameters $s = [(0,2),(2,3),\ldots,(8,9)]$ and
$k = 8$, while the lookup-function finishes instantly because it just caches the
specification for lookup during inference.
However, \toolname{} has better drawdown and significantly better generalization
compared to the lookup-function. \toolname{} has -0.07\% drawdown, which
indicates improvement in drawdown set accuracy, as well as 60.98\% generalization on $S_1$
and 63.53\% generalization on $S_2$.
By contrast, the lookup-function has no drawdown and no generalization.
Moreover, the lookup-function introduces significant runtime overhead to the
network, which is evaluated in~\pref{sec:Experiment-Overhead}.

For the approximate 2D V-polytope repair specification,
PRDNN took 3,834 seconds with parameter $k=5$, while REASSURE \emph{times out in
one day}. PRDNN has -0.07\% drawdown, 53.97\% generalization on $S_1$ and 61.5\%
generalization on $S_2$. Although PRDNN provably repairs this approximate 2D
V-polytope repair specification, it does not guarantee provable repair for the
true V-polytope repair specification---there are still violations of the true
specification.

For the approximate pointwise repair specification,
\toolname{} took 88 seconds with parameters ${s=[]}$ and $k=7$, while REASSURE took 9,958 seconds.
\toolname{} has no drawdown, 48.85\% generalization on $S_1$ and
52.31\% generalization on $S_2$.
By contrast, REASSURE has no drawdown but also no generalization.
Although both \toolname{} and REASSURE's repairs for this approximate
specification do not guarantee provable repair for the true V-polytope repair
specification, \toolname{} achieves much higher generalization in a
significantly shorter time.
Moreover, the REASSURE repaired network has a significant runtime overhead,
which is evaluated in~\pref{sec:Experiment-Overhead}. The REASSURE repaired
$9\times 100$ network has $2,117\times$ more parameters (150,210 to 318,075,710);
it takes one minute and 8~GB of GPU memory to evaluate one image using NVIDIA
RTX A6000.

This experiment highlights \toolname{}'s superior efficiency in provable
V-polytope repair with low drawdown and high generalization.

\subsection{V-Polytope ACAS Xu Local Safety and Robustness Repair}
\label{sec:EvaluationACASPoly}

\subsubsubsection{Buggy network.}
We repair ACAS Xu network
$\dnn_{2,9}$~\cite{DBLP:journals/corr/abs-1810-04240}, which processes a
five-dimensional input representing the state of an unmanned aircraft and an
intruder, and issues navigation advisories to avoid the intruder. This network
has seven layers and $13,350$ parameters.

\newcommand{\acasBoxSize}{0.05}
\newcommand{\acasBoxSampleSpacing}{0.005}
\newcommand{\acasBoxSampleSpacingPRDNN}{0.016}

\newcommand{\acasInputBoxLBI}{0}
\newcommand{\acasInputBoxUBI}{60760}
\newcommand{\acasInputBoxLBII}{-3.141593}
\newcommand{\acasInputBoxUBII}{3.141593}
\newcommand{\acasInputBoxLBIII}{-3.141593}
\newcommand{\acasInputBoxUBIII}{3.141593}
\newcommand{\acasInputBoxLBIV}{100}
\newcommand{\acasInputBoxUBIV}{1200}
\newcommand{\acasInputBoxLBV}{0}
\newcommand{\acasInputBoxUBV}{1200}
\newcommand{\myargmax}{\texttt{argmax}}
\newcommand{\myargmin}{\texttt{argmin}}

\subsubsubsection{V-polytope repair specification.}
There are five safety properties that the ACAS~Xu network $\dnn_{2,9}$ should
satisfy; denoted as $\phi_1$, $\phi_2$, $\phi_3$, $\phi_4$, and $\phi_8$
in~\cite{DBLP:conf/cav/KatzBDJK17}. Let $\phi \eqdef \phi_1 \wedge \phi_2 \wedge
\phi_3 \wedge \phi_4 \wedge \phi_8$. There are known violations of safety
property $\phi$ in network $\dnn_{2,9}$'s
behavior~\cite{DBLP:conf/cav/KatzBDJK17}.
We evenly partition the 5D input
polytopes into boxes with a spacing of $\acasBoxSize$, and take 24 boxes that
contain points violating safety property $\phi$ to form the V-polytope repair
specification.
For each 5D box~$\polytope$, the repair specification states that
(a)~for all points $X \in \convexhull(\polytope)$, the network $\dnn_{2,9}$ should
satisfy $\phi$, \emph{and} (b)~for all $X$ and $X'$ in $\convexhull(\polytope)$,
$\argmin(\dnn_{2,9}(X)) = \argmin(\dnn_{2,9}(X'))$. Part~(b) of the repair
specification states that the repaired network should be robust in the polytope
$\polytope$. Note that ACAS Xu uses $\argmin$
instead of $\argmax$ to compute the output classification.

\subsubsubsection{Approximate pointwise repair specification for PRDNN.}
Because PRDNN is unable to handle 5D input V-polytopes, we evenly sample each 5D
box in the V-polytope repair specification with a spacing of
$\acasBoxSampleSpacingPRDNN$ to form an approximate
pointwise repair specification of 1,368 points.
For each of these sampled points, the network has to satisfy the safety property $\phi$.
Further, to approximate robustness,
all points sampled from the same 5D box has to have the same output classification.

\subsubsubsection{Generalization set.}
The generalization set consists of 169,625 points that violate safety property $\phi$,
disjoint from the polytopes in the repair specification.
We define \emph{property generalization} as the percentage of
points that satisfy $\phi$ in the repaired network; higher property generalization is better.

\subsubsubsection{Drawdown set.}
The drawdown set consists of 283,772,487 points for which
the original network $\dnn_{2,9}$ satisfies the safety property~$\phi$.
We define \emph{property drawdown} as the percentage of
points in the drawdown set that do not satisfy $\phi$ in the repaired network;
lower property drawdown is better.

\subsubsubsection{Results.}
\toolname{} took 56 seconds to repair
the network $\dnn_{2,9}$ to satisfy the
V-polytope repair specification
with parameters $s=[(0,1),(1,2),\ldots,(5,6)]$ and $k=6$.
By contrast, PRDNN took 848 seconds to repair the network $\dnn_{2,9}$ to
satisfy the approximate pointwise repair specification (with parameter $k=6$),
while the resulting network still violates the V-polytope repair specification.
The property drawdown for \toolname{} is 0.60\%, while that
for PRDNN is 0.45\%.
The property generalization for \toolname{} is 100\%,
while that for PRDNN is 92.58\%.
This experiment shows \toolname{}'s support~for higher-dimensional V-polytope repair specifications and its superior efficiency compared to PRDNN.

\subsection{V-Polytope ACAS Xu Global Safety Repair}
\label{sec:EvaluationACASPolyComplete}

\newcommand{\acasCompleteBoxSize}{0.13}
\newcommand{\acasCompleteBoxSampleSpacing}{0.01}

\subsubsubsection{Buggy network.} We use the same ACAS~Xu network $\dnn_{2,9}$ used
in \pref{sec:EvaluationACASPoly}.

\subsubsubsection{V-polytope repair specification.}
The V-polytope repair specification states that the safety property $\phi$ (from
\pref{sec:EvaluationACASPoly}) should be satisfied by all points in the (valid)
input space for the network $\dnn_{2,9}$.
Specifically, we evenly partition the input polytopes for the five safety properties
that are applicable to $\dnn_{2,9}$ into disjoint 5D boxes of size~$\acasCompleteBoxSize$.
Each of the 5D~box is further partitioned into 120 disjoint 5D~simplices among
its $32$ box vertices.
The final set of 27,600 5D simplices covering the \emph{entire valid input space}
is used to define the V-polytope repair specification.

\subsubsubsection{Results.}
\toolname{} took 102~seconds to repair the network $\dnn_{2,9}$
to satisfy the V-polytope repair specification
with parameter $s=[(0,3),(3,4),(4,5),(5,6)]$ and $k=6$.
Because this repair covers
the \emph{entire} input space for each safety property, the resulting
repaired DNN is guaranteed to satisfy \emph{all} five safety properties.
In other words, the repaired ACAS Xu $\dnn_{2,9}$ is provably safe.
This experiment highlights \toolname{}'s scalability to large numbers of
V-polytopes in the repair specification and its ability to repair the
entirety of an input space.

\subsection{Pointwise MNIST Image Corruption Repair for Non-PWL DNNs}
\label{sec:Experiment-Non-PWL}

In this experiment we study \toolname{}'s support for provably repairing Non-PWL DNNs.

\subsubsubsection{Buggy networks.}
We repair MNIST $3\!\times\! 100$ DNNs with only Hardswish or
GELU activation functions. Hardswish is used in MobileNetV3~\cite{DBLP:conf/iccv/HowardPALSCWCTC19}, and
GELU is used in most transformer-based models like GPT-3~\cite{DBLP:conf/nips/BrownMRSKDNSSAA20} and BERT~\cite{DBLP:conf/naacl/DevlinCLT19}.

\subsubsubsection{Pointwise repair specification.}
The MNIST-C dataset~\cite{DBLP:journals/corr/abs-1906-02337} consists of MNIST
images from the official MNIST test set that have been corrupted. The pointwise
repair specifications consists of the first $1, 10, 50$ and 100 fog-corrupted
images from the MNIST-C dataset that were \emph{misclassified} by that network
along with their correct classification.

\subsubsubsection{Generalization set.}
9,000 images from MNIST-C's foggy test set, disjoint from the repair set.

\subsubsubsection{Drawdown set.}
The drawdown set is the official MNIST test set containing 10,000 images.

\subsubsubsection{Results.}
\toolname{} was able to repair all networks such that they satisfied the pointwise
repair specification. \pref{tab:MnistPointwiseRepairNonPWL} summarizes the
results. As the number of repair points increases, the generalization
improves (increases), but the drawdown worsens (increases). Although
\toolname{} is able to provably repair DNNs with non-PWL activation functions,
maintaining a good (low) drawdown down remains a challenge. We leave this as
interesting future work.
 
\begin{figure}[t]
    \centering
    \hfill
    \begin{minipage}{0.45\textwidth}
        \centering
        \hspace{-3.5ex}
        \begin{tikzpicture}
            \begin{axis}
            [
                scale=0.6,
                xlabel=Dimensionality $d$,
                ylabel=Repair Time (s),
ymode=log,
                width=1.5\textwidth,
                height=1.2\textwidth,
                y label style={font=\small, at={(axis description cs:0.08,.5)}},
                x label style={font=\small, at={(axis description cs:.5,0.05)}},
                y tick label style={font=\small},
                x tick label style={font=\small},
                symbolic x coords={5,6,7,8,9,10,11,12,13,14,15,16},
                xtick={5,6,7,8,9,10,11,12,13,14,15,16},
legend style={font=\small, nodes={scale=0.8, transform shape}},
legend pos=south east, ]
\addplot+[mark=*, mark size=1.5pt]
                table [
                    x=n,
                    y=n_pixel,
                ]
                {data/ndim_local_robust.csv};
            \addlegendentry{$d$-pixel};
\addplot+[mark=*, mark size=1.5pt]
            table [
                x=n,
                y=n_partition,
            ]
            {data/ndim_local_robust.csv};
            \addlegendentry{$d$-partition};
            \end{axis}
        \end{tikzpicture}
        \vspace{-2ex}
        \caption{\toolname{} repair time for MNIST $d$-dimen\-sional $L^\infty$
        local-robustness repair.}
        \label{fig:Local-Robustness}
    \end{minipage}
\hfill
    \begin{minipage}{0.49\textwidth}
        \centering
\captionof{table}{
Provable pointwise repair of MNIST networks with non-PWL activation functions
            using \toolname{}.
            k:~repair parameter,
            D:~drawdown,
            G:~generalization,
            T:~time.
The repair parameter $s$ is empty for all runs.
        }
        \label{tab:MnistPointwiseRepairNonPWL}
{\footnotesize
        \setlength{\tabcolsep}{2pt}
\begin{tabular}{ccccccccccc}
        \toprule
        \multicolumn{2}{c}{\multirow{2}{*}{$\;$Points$\;$}} &
        \multicolumn{4}{c}{Hardswish} & &
        \multicolumn{4}{c}{GELU} \\
        \cmidrule{3-6}
        \cmidrule{8-11}
        && k & D & G & T
        && k & D & G & T \\
        \midrule
        \multicolumn{2}{c}{1  } & 1 & 0.32\%  & 5.54\%   & 1s && 1 & 0.05\%  & 1.38\%  & 1s \\
        \multicolumn{2}{c}{10 } & 1 & 9.15\%  & 13.08\%  & 1s && 1 & 13.24\% & 6.49\%  & 1s \\
        \multicolumn{2}{c}{50 } & 1 & 25.61\% & 17.53\%  & 2s && 1 & 32.60\% & 8.66\%  & 3s \\
        \multicolumn{2}{c}{100} & 1 & 25.28\% & 25.44\%  & 5s && 1 & 42.21\% & 13.80\% & 8s \\
        \bottomrule
        \end{tabular}
}
    \end{minipage}
    \hfill
\vspace{-2ex}
\end{figure} 
\subsection{\texorpdfstring{$d$}{d}-Dimensional \texorpdfstring{$L^\infty$}{L{\textasciicircum}infty} Local-Robustness Repair for MNIST}
\label{sec:Experiment-scale}

In this experiment, we study the scalability of \toolname{}'s provable
V-polytope repair. For MNIST DNNs, an $L^\infty$ local-robustness specification
is expressed with a $784$-dimensional cube of $2^{784}$ vertices in its
V-representation. Such a V-polytope repair specification is challenging because
\toolname{}'s runs in polynomial time in the number of vertices. We
define two $d$-dimensional local-robustness specifications
for MNIST where $d \leq 784$, and evaluate \toolname{}'s scalability as
$d$~increases.

\subsubsubsection{Buggy network.} We repair the MNIST $3\times
100$ DNN from~\cite{ERAN}.

\subsubsubsection{V-polytope repair specification.}
For each misclassified fog-corrupted image $\point$ of label $l$ among five
random choices from the MNIST-C
dataset~\cite{DBLP:journals/corr/abs-1906-02337}, we consider the following two
classes of $d$-dimensional local-robustness specifications where $5 \leq d \leq
16$ and $\varepsilon \eqdef 0.1$.
(a)~For a subset of $d$ pixels ($d$-pixel), the DNN is locally robust to any
$L^\infty$-norm-bounded $\varepsilon$-perturbation of $\point$ on only those $d$
pixels.
Specifically, we consider the first $d$ non-zero pixels and four random choices
of $d$ pixels.
(b)~For $d$ equal-or-near-equal-sized partitions of all pixels ($d$-partition),
the DNN is locally robust to any $L^\infty$-norm-bounded
$\varepsilon$-perturbation of $\point$ on those $d$ partitions, where pixels in
the same partition are perturbed with the same delta.
Specifically we consider three partitions that consist of $4\times 4$, $7\times
7$ or $1\times 28$ blocks.

\subsubsubsection{Drawdown set.}
The drawdown set is the official MNIST test set containing 10,000 images.

\subsubsubsection{Results.}
\pref{fig:Local-Robustness} shows the average repair time for $d$-pixel and
$d$-partition $L^\infty$ local-robustness repairs
with parameters $s=[(0,1)]$ and $k=1$ as well as a time limit of 20,000 seconds.
The choice of images and the $d$
pixels/partitions makes little difference in repair time and drawdown.
For the $d$-pixel $L^\infty$ local-robustness repair, \toolname{} scales to
$16$~dimensions (65,536 vertices) in 13,193 seconds. The worst drawdown is under 0.5\%
but most are under 0.2\%.
For the $d$-partition $L^\infty$ local-robustness repair, \toolname{} scales to
$14$~dimensions (16,384 vertices) in 18,552 seconds. The worst drawdown is under
1\% but most are under~0.2\%.

\subsection{Runtime Overhead Analysis}
\label{sec:Experiment-Overhead}

In this experiment we evaluate the runtime overhead of
non-architecture-preserving provable repair of DNNs by comparing \toolname{}
with PRDNN, REASSURE and the lookup-based approach on pointwise and polytope repair.
We use the MNIST $3\times 100$ DNN from~\cite{ERAN}.

\begin{wrapfigure}[24]{r}{0.3\textwidth}
\begin{subfigure}[t]{0.3\textwidth}
\begin{tikzpicture}
\begin{axis}
[scale=0.45, xlabel=$m$ Repair Points,ylabel=Inference Time (s),ymin=0,ymax=220,
y label style={at={(axis description cs:.16,.5)}, font=\small},
    x label style={at={(axis description cs:.5,.08)}, font=\small},
    x tick label style={font=\tiny},
    y tick label style={font=\tiny},
    symbolic x coords={0,10,20,30,40,50,60,70,80,90,100},
    xtick={0,10,20,30,40,50,60,70,80,90,100},
legend style={font=\tiny},
    legend pos=north west, ]
\addplot+[mark=*, mark size=1.5pt, color=blue] table [x=Points, y=Time, col sep=comma]
    {data/overhead_pointwise_reassure.csv};
\addlegendentry{REASSURE};

\addplot+[mark=*, mark size=1.5pt, color=red] table [x=Points, y=Time, col sep=comma]
    {data/overhead_pointwise_aprnn.csv};
\addlegendentry{\toolname{}};
\end{axis}
\end{tikzpicture}
\vspace{-1ex}
\caption{For pointwise repair.}
\label{fig:overhead-pointwise}
\end{subfigure}
\\
\vspace{3ex}
\begin{subfigure}[t]{0.3\textwidth}
\begin{tikzpicture}
\begin{axis}
[
    scale=0.45,
    xlabel=$m$ Repair Polytopes,
    ylabel=Inference Time (s),
    ymin=0,
y label style={at={(axis description cs:.16,.5)}, font=\small},
    x label style={at={(axis description cs:.5,.08)}, font=\small},
    x tick label style={font=\tiny},
    y tick label style={font=\tiny},
    symbolic x coords={1,2,3,4,5,6,7,8,9,10},
    xtick={1,2,3,4,5,6,7,8,9,10},
legend style={font=\tiny},
    legend pos=north west, ]
\addplot+[mark=*, mark size=1.5pt, color=blue] table [x=Polytopes, y=Time]
    {data/overhead_polytope_lookup.csv};
\addlegendentry{Lookup};
\addplot+[mark=*, mark size=1.5pt, color=red] table [x=Polytopes, y=APRNN_Time]
    {data/overhead_polytope_lookup.csv};
\addlegendentry{\toolname{}};
\end{axis}
\end{tikzpicture}
\vspace{-1ex}
\caption{For polytope repair.}
\label{fig:overhead-polytope}
\end{subfigure}
\caption{Runtime overhead.}
\label{fig:overhead}
\end{wrapfigure}

\subsubsubsection{Results.}
We measure the inference time for 10,000 images from the MNIST test set in a
single batch using GPU.
The inference time for the original network is 0.2 seconds.
\toolname{} is architecture-preserving, hence has no runtime overhead; the
inference time of the \toolname{} repaired network is also 0.2 seconds.
PRDNN's runtime overhead only depends on the network size, the inference time
of the PRDNN repaired network is 0.35 seconds.

REASSURE has a \emph{significant} runtime overhead, which
is polynomial in the network size and the number of buggy linear regions.
For pointwise repair,
\pref{fig:overhead-pointwise} shows the runtime overhead of REASSURE repaired
networks for $m$ points
(which reduces to $m$ linear regions).
For polytope repair, REASSURE needs to enumerate and repair all buggy linear
regions in each polytope. As described in~\pref{sec:Comparison}, there can be a
very large number of such linear regions and REASSURE does not provide a way to
enumerate them. Thus, if the REASSURE is able to repair the network, then the
runtime overhead will be significantly higher than the pointwise repair case.

For the lookup-based override approach, the runtime overhead for pointwise
repair is constant due to the use of a hash-table; the inference time is 0.26
seconds. However, the runtime overhead for polytope repair is \emph{significant}
due to the use of LP solvers to check whether an input point is in a V-polytope.
It is polynomial in the total number of vertices of buggy V-polytopes.
\pref{fig:overhead-polytope} shows the runtime overhead of lookup-based override
approach repaired networks for $m$ 5D boxes (32 vertices per box).

\section{Related Work}
\label{sec:RelatedWork}

\subsubsubsection{Provable DNN repair.}
Our work falls into the class of provable DNN repair methods. The most closely related
works are PRDNN~\cite{DBLP:conf/pldi/SotoudehT21}, REASSURE~\cite{DBLP:conf/iclr/Fu22} (discussed
in \pref{sec:Comparison}), and
Minimal Modifications of DNNs (MMDNN)~\cite{DBLP:conf/lpar/GoldbergerKAK20}.
MMDNN performs architecture-preserving provable pointwise repair of a DNN by
reducing the problem to the NP-complete DNN verification problem. It is not able
to scale to large DNNs, only supports the ReLU activation function, and does not
support polytope repair or multi-layer repairs.

\subsubsubsection{Heuristic DNN repair.}
Heuristic DNN repair methods do
not provide guarantees about repair efficacy.
MEND~\cite{mitchell2022fast}
preserves the network architecture by training auxiliary editor networks
to determine edits to the original DNN's parameters.
MEND supports multi-layer edits,
but does not handle V-polytope repair specifications.

SERAC~\cite{DBLP:conf/icml/MitchellLBMF22}
does not preserve the network architecture;
instead of altering the original
buggy DNN's parameters, SERAC trains editor networks and modulates the final
output based on the cached repair points and editor networks.
SERAC supports the repair of DNNs with any activation functions,
but does not handle V-polytope repair specifications.

NNRepair~\cite{DBLP:conf/cav/UsmanGSNP21} modifies network parameters such that
a given repair input follows a certain activation pattern. It produces a
collection of experts (one per class label). NNRepair proposes three variants on
how to combine these experts, and at least one of these is not architecture
preserving. It does not support V-polytope repair specifications.

\subsubsubsection{DNN training for repair.}
Training-based DNN repair methods support any activation functions and
multi-layer edits, but do not provide any correctness guarantees.
Sinitsin et al.~\cite{DBLP:conf/iclr/SinitsinPPPB20} propose
editable neural networks, a framework that aims to make DNNs more amenable to an
editing function that enforces changes in their behavior.
Lin et al.~\cite{DBLP:conf/fmcad/LinZSJ20} propose a method to use safety
properties as loss functions during training.
There are also certified training approaches for local~\cite{DBLP:conf/icml/MirmanGV18} and global robustness properties~\cite{DBLP:conf/icml/LeinoWF21}.

\subsubsubsection{DNN verification.}
DNN verification methods~\cite{DBLP:conf/cav/KatzBDJK17,DBLP:conf/nips/WangPWYJ18,DBLP:conf/fmcad/LahavK21,10.1145/3498704,DBLP:conf/iclr/XuZ0WJLH21,DBLP:conf/iclr/PalmaBBTK21,DBLP:conf/iclr/FerrariMJV22,DBLP:conf/nips/SinghGMPV18,DBLP:journals/pacmpl/SinghGPV19}
can be used in conjunction with DNN repair to identify incorrect DNN behavior
and verify the correctness of the DNN post-repair. Because \toolname{} is architecture-preserving,
it can be easily integrated with these existing DNN verification methods.

\section{Conclusion}
\label{sec:Conclusion}

We presented a new approach for architecture-preserving provable V-polytope
repair of DNNs, which runs in polynomial time, supports a wide
variety of activation functions, and has the flexibility to modify weights in multiple
layers. To the best of our knowledge, it is the first approach for provable DNN repair
that supports all of these features. We implemented our approach in a tool called
\toolname{}.
Using MNIST, ImageNet, and ACAS Xu DNNs, our experiments showed that \toolname{}
has better efficiency, scalability, and generalization
compared to PRDNN and REASSURE, prior provable-repair techniques.
We used V-polytope repair specifications to repair MNIST networks for rotations
and ACAS Xu networks for robustness.
We also
showed how \toolname{} can be used to repair polytopes covering the entire input
space of an ACAS Xu network; in effect, ensuring that the repaired network
satisfies global safe properties.
However, it is not feasible to
represent an $L^\infty$ local robustness specification on MNIST
DNNs using V-polytopes due to the exponential number of vertices.
Developing a provable repair approach that supports
H-representation of polytopes is a natural and challenging direction for future research.

\begin{acks}
\onlyfor{arxiv}{
    We would like to thank the PLDI~2023
    anonymous reviewers as well as our shepherd
    Gagandeep Singh for their feedback and suggestions, which have greatly
    improved the quality of the paper.
}{
    We would like to thank the
    anonymous reviewers as well as our shepherd
    Gagandeep Singh for their feedback and suggestions, which have greatly
    improved the quality of the paper.
}
    This work is supported in part by
    NSF grant CCF-2048123 and DOE Award DE-SC0022285.
\end{acks}

\section*{Data-Availability Statement}
The artifact associated with the paper can be found at~\cite{APRNNArtifact}.
The latest version of the APRNN tool is available at~\url{https://github.com/95616ARG/APRNN}.

\bibliographystyle{ACM-Reference-Format}
\bibliography{main}

\onlyfor{arxiv}{
\clearpage
\appendix
\section{Proofs of Theorems}
\label{app:Appendix-Proof}

\begin{reptheorem}{thm:Conditional-ReLU}
\defProof{proof:Conditional-ReLU}
$\condsy\relu$
is a conditional symbolic activation function for \relu. \end{reptheorem}

\begin{proof}
Because the ReLU activation function is defined component-wise~(\pref{def:relu}), we prove all properties in a component-wise manner.
Let
$
    \sy\pointAlt, \varphi \eqdef \condsy\relu\withRef{\point_\RefKW}\bigl(
        \sy\point
    \bigr)
$
and $\theta \vDash \varphi$.
Consider any $i$th component for $0 \leq i < n$.
Here we use $\sy{X}^\RefKW_i$ to denote the $i$th component of $\sy{X}_\RefKW$.

\begin{cs}[align=left,leftmargin=*,wide=0pt]

\item[\ref{cond:Conditional-Activation-Function-C1}]
\textbf{Case~(i)}:
If $\sy{X}^\RefKW_i \geq 0$,
then $\sy{Y}_i \eqdef \sy{X}_i$
and $\sy{X}_i \geq 0$ is conjoined into $\varphi$~(\pref{li:SymbolicReLU-On}).
Hence, $\theta \vDash \varphi$ implies $\theta \vDash \sy{X}_i \geq 0$.
\textbf{Case~(ii)}: If $\sy{X}^\RefKW_i < 0$,
then $\sy{Y}_i \eqdef 0$
and $\sy{X}_i < 0$ is conjoined into $\varphi$~(\pref{li:SymbolicReLU-Off}).
Hence, $\theta \vDash \varphi$ implies $\theta \vDash \sy{X}_i < 0$.
In both cases, $
    \eval{\sy{Y}_i}_{\theta} = \scalarReLU\bigl(
        \eval{\sy{X}_i}_{\theta}
    \bigr)
$ using~\pref{def:relu}.

\item[\ref{cond:Conditional-Activation-Function-C2}]
Suppose $\sy{X}$ is a linear expression.
\textbf{Case~(i)}: If $\sy{X}^\RefKW_i \geq 0$,
then $\sy{Y}_i \eqdef \sy{X}_i$ is a linear expression
and a linear formula $\sy{X}_i \geq 0$ is conjoined into $\varphi$~(\pref{li:SymbolicReLU-On}).
\textbf{Case~(ii)}: If $\sy{X}^\RefKW_i < 0$, $\sy{Y}_i \eqdef 0$ is a linear expression
and a linear formula $\sy{X}_i < 0$ is conjoined into $\varphi$~(\pref{li:SymbolicReLU-Off}).
Hence, $\sy{Y}$ is a linear expression and $\varphi$ is a linear formula on~\pref{li:SymbolicReLU-Return}.

\item[\ref{cond:Conditional-Activation-Function-C3}]
Let the symbolic polytope
$
    \sy\polytope \eqdef
        \left\{
            \idx{0}\sy\point,
            \idx{1}\sy\point,
            \ldots,
            \idx{p-1}\sy\point,
        \right\}
$
and
$
    \idx{j}\sy\pointAlt, \idx{j}\varphi \eqdef
        \condsy\sigma\withRef{\point_\RefKW}
        \bigl( \idx{j}\sy\point \bigr)
$ for $0 \leq j < p$.
Let $\theta \vDash \bigwedge_j \idx{j}\varphi$.
To show that ReLU is \locallylinear{} for $\eval{ \sy\polytope }_\theta$,
we construct a linear function $f$ such that
$
    f\bigl( \point \bigr) =
    \relu\bigl( \point \bigr)
$ for any $\point \in \convexhull\bigl(\eval{ \sy\polytope }_\theta\bigr)$~(\pref{def:locallylinear}).
Again, consider any $i$th component for $0 \leq i < n$.
\textbf{Case~(i)}:
If $\sy{X}^\RefKW_i \geq 0$,
on~\pref{li:SymbolicReLU-On}, $\idx{j}\sy{X}_i \geq 0$ is conjoined into $\idx{j}\varphi$.
Hence, $\theta \vDash \bigwedge_j \idx{j}\varphi$ implies $\idx{j}\sy{X}_i \geq 0$
for all $j$.
Because $\point \in \convexhull\bigl(\eval{ \sy\polytope }_\theta\bigr)$,
$
    \point
    \geq \min\limits_{j}{ \eval{ \idx{j}\sy{X}_i }_\theta }
    \geq 0
$.
Hence, $\point \geq 0$
and there exists a linear function
$
    f\bigl( X \bigr)_i = X_i
$
such that
$
    f\bigl( X \bigr)_i = \relu\bigl( X \bigr)_i
$.
\textbf{Case~(ii)}:
If $\sy{X}^\RefKW_i < 0$,
on~\pref{li:SymbolicReLU-Off}, $\idx{j}\sy{X}_i < 0$ is conjoined into $\idx{j}\varphi$.
Hence, $\theta \vDash \bigwedge_j \idx{j}\varphi$ implies $\idx{j}\sy{X}_i < 0$
for all $j$.
Because $\point \in \convexhull\bigl(\eval{ \sy\polytope }_\theta\bigr)$,
$
\point
    \leq \max\limits_{j}{ \eval{ \idx{j}\sy{X}_i }_\theta }
    < 0
$.
Hence, $\point < 0$
and there exists a linear function
$
    f\bigl( X \bigr)_i = 0
$
such that
$
    f\bigl( X \bigr)_i = \relu\bigl( X \bigr)_i
$.
\qedhere
\end{cs}
\end{proof}
 
\begin{reptheorem}{thm:Conditional-Hardswish}
\defProof{proof:Conditional-Hardswish}
$\condsy\hardswish$
is a conditional symbolic activation function for \hardswish. \end{reptheorem}

\begin{proof}
    Because the Hardswish activation function is defined component-wise~(\pref{def:hardswish}), we prove all properties in a component-wise manner.
Let
    $
        \sy\pointAlt, \varphi \eqdef \condsy\hardswish\withRef{\point_\RefKW}\bigl(
            \sy\point
        \bigr)
    $
    and $\theta \vDash \varphi$.
    Consider any $i$th component for $0 \leq i < n$.
    Here we use $\sy{X}^\RefKW_i$ to denote the $i$th component of $\sy{X}_\RefKW$.

\begin{cs}[align=left,leftmargin=*,wide=0pt]

\item[\ref{cond:Conditional-Activation-Function-C1}]
    \textbf{Case~(i)}:
    If $\sy{X}^\RefKW_i \geq 0$,
    then $\sy{Y}_i \eqdef \sy{X}_i$
    and $\sy{X}_i \geq 3$
    is conjoined into $\varphi$~(\pref{li:SymbolicHardSwish-On}).
    Hence, $\theta \vDash \varphi$
    implies $\theta \vDash \sy{X}_i \geq 3$.
\textbf{Case~(ii)}: If $\sy{X}^\RefKW_i < 0$,
    then $\sy{Y}_i \eqdef 0$
    and $\sy{X}_i \leq -3$ is conjoined into $\varphi$~(\pref{li:SymbolicHardSwish-Off}).
    Hence, $\theta \vDash \varphi$ implies $\theta \vDash \sy{X}_i \leq -3$.
    In both cases, $
        \eval{\sy{Y}_i}_{\theta} = \scalarHardswish\bigl(
            \eval{\sy{X}_i}_{\theta}
        \bigr)
    $ using~\pref{def:hardswish}.

\item[\ref{cond:Conditional-Activation-Function-C2}]
    Suppose $\sy{X}$ is a linear expression.
    \textbf{Case~(i)}: If $\sy{X}^\RefKW_i \geq 0$,
    then $\sy{Y}_i \eqdef \sy{X}_i$ is a linear expression
    and a linear formula
    $\sy{X}_i \geq 3$ is conjoined into $\varphi$~(\pref{li:SymbolicHardSwish-On}).
    \textbf{Case~(ii)}: If $\sy{X}^\RefKW_i < 0$,
    $\sy{Y}_i \eqdef 0$ is a linear expression
    and a linear formula
    $\sy{X}_i \leq -3$ is conjoined into $\varphi$~(\pref{li:SymbolicHardSwish-Off}).
    Hence, $\sy{Y}$ is a linear expression and $\varphi$ is a linear formula on~\pref{li:SymbolicHardSwish-Return}.

\item[\ref{cond:Conditional-Activation-Function-C3}]
    Let the symbolic polytope
    $
        \sy\polytope \eqdef
            \left\{
                \idx{0}\sy\point,
                \idx{1}\sy\point,
                \ldots,
                \idx{p-1}\sy\point,
            \right\}
    $
    and
    $
        \idx{j}\sy\pointAlt, \idx{j}\varphi \eqdef
            \condsy\sigma\withRef{\point_\RefKW}
            \bigl( \idx{j}\sy\point \bigr)
    $ for $0 \leq j < p$.
    Let $\theta \vDash \bigwedge_j \idx{j}\varphi$.
    To show that Hardswish is \locallylinear{} for $\eval{ \sy\polytope }_\theta$,
    we construct a linear function $f$ such that
    $
        f\bigl( \point \bigr) =
        \hardswish\bigl( \point \bigr)
    $ for any $\point \in \convexhull\bigl(\eval{ \sy\polytope }_\theta\bigr)$~(\pref{def:locallylinear}).
Again, consider any $i$th component for $0 \leq i < n$.
\textbf{Case~(i)}:
    If $\sy{X}^\RefKW_i \geq 0$,
    on~\pref{li:SymbolicHardSwish-On},
    $\idx{j}\sy{X}_i \geq 3$ is conjoined into $\idx{j}\varphi$.
    Hence, $\theta \vDash \bigwedge_j \idx{j}\varphi$
    implies $\idx{j}\sy{X}_i \geq 3$
    for all $j$.
    Because $\point \in \convexhull\bigl(\eval{ \sy\polytope }_\theta\bigr)$,
    $
        \point
        \geq \min\limits_{j}{ \eval{ \idx{j}\sy{X}_i }_\theta }
        \geq 3
    $.
    Hence, $\point \geq 3$
    and there exists a linear function
    $
        f\bigl( X \bigr)_i = X_i
    $
    such that
    $
        f\bigl( X \bigr)_i = \hardswish\bigl( X \bigr)_i
    $.

\textbf{Case~(ii)}:
    If $\sy{X}^\RefKW_i < 0$,
    on~\pref{li:SymbolicHardSwish-Off},
    $\idx{j}\sy{X}_i \leq -3$ is conjoined into $\idx{j}\varphi$.
    Hence, $\theta \vDash \bigwedge_j \idx{j}\varphi$
    implies $\idx{j}\sy{X}_i \leq -3$
    for all $j$.
    Because $\point \in \convexhull\bigl(\eval{ \sy\polytope }_\theta\bigr)$,
    $
\point
        \leq \max\limits_{j}{ \eval{ \idx{j}\sy{X}_i }_\theta }
        \leq -3
    $.
    Hence, $\point \leq -3$
    and there exists a linear function
    $
        f\bigl( X \bigr)_i = 0
    $
    such that
    $
        f\bigl( X \bigr)_i = \hardswish\bigl( X \bigr)_i
    $.
\qedhere
\end{cs}
\end{proof} 
\begin{reptheorem}{thm:Conditional-Forward-Pointwise}
\defProof{proof:Conditional-Forward-Pointwise}
Let
$
    \sy\point\ly{L}, \varphi
    \eqdef
    \condsy\dnn\withRef{\point\ly{0}_\RefKW}
    \big(\point\ly{0}\big)
$
and $\theta \vDash \varphi$, then
\begin{cs}
\item\labelInMainText{cond:Conditional-Forward-Pointwise-Correct}
    $
        \eval{\sy\point\ly{L}}_{\theta}
        =
        \dnn^{\theta}\big(\point\ly{0}\big)
    $.
\item\labelInMainText{cond:Conditional-Forward-Pointwise-Linear}
    $\sy\point\ly{L}$ is a linear expression and $\varphi$ is a linear formula.
\item\labelInMainText{cond:Conditional-Forward-Pointwise-LocallyLinear}
    Let the V-polytope
    $
        \polytope\ly{0} \eqdef
            \left\{
                \idx{0}\point\ly{0},
                \idx{1}\point\ly{0},
                \ldots,
                \idx{p-1}\point\ly{0}
            \right\}
    $,
    $
        \idx{j}\sy\point\ly{L}, \idx{j}\varphi \eqdef
            \condsy\dnn
            \withRef{\point\ly{0}_\RefKW}
            \bigl(
                \idx{j}\point\ly{0}
            \bigr)
    $ for $0 \leq j < p$, and
     $\varphi_\polytope \eqdef \bigwedge_j \idx{j}\varphi$.
    If $\,\theta \vDash \varphi_\polytope$,
    then $\dnn^\theta$ is \locallylinear{} for the polytope $\polytope\ly{0}$.
\end{cs} \end{reptheorem}

\begin{proof}

We prove each condition of \pref{thm:Conditional-Forward-Pointwise}.

\begin{cs}[align=left,leftmargin=*,wide=0pt]
\item[\ref{cond:Conditional-Forward-Pointwise-Correct}]
We prove this condition by structural induction.
\textbf{Base case}: For the first layer $\ly{\ell} = 0$, we will show
$
    \eval{\sy\point\ly{1}}_{\theta}
    =
    \dnn\ly{0}\bigargs{
        X\ly{0}; \eval{\sy{W}\ly{0}}_{\theta}, \eval{\sy{B}\ly{0}}_{\theta}
    }
= \sigma\ly{0}\bigl(
X\ly{0} \eval{\sy{W}\ly{0}}_{\theta} + \eval{\sy{B}\ly{0}}_{\theta}
\bigr)
$.
On~\pref{li:Conditional-Forward-Point-Symbolic-Pre-First},
$
    \sy\point\ly{1}_\PreKW \eqdef X\ly{0} \sy{W}\ly{0} + \sy{B}\ly{0}
$.
On~\pref{li:Conditional-Forward-Point-Symbolic-Post},
$
    \sy\point\ly{1}, \varphi_\sigma\ly{1}
    \eqdef \condsy{{\sigma}}\ly{0}\withRef{
        X_{\RefKW}\ly{1}
    }\bigl( \sy{X}\ly{1}_\PreKW \bigr)
$.
Because
$\varphi\eqdef \bigwedge_{1\leq\ell\leq L}{\varphi_\sigma\ly{\ell}}$,
$\theta \vDash \varphi$ implies $\theta \vDash \varphi_\sigma\ly{1}$.
By~\pref{def:Conditional-Activation-Function}-\ref{cond:Conditional-Activation-Function-C1}, we have
$
    \eval{
        \sy\point\ly{1}
    }_\theta
    =
    \sigma\ly{0}\bigl(
    \eval{
        \sy\point\ly{1}_\PreKW
    }_{\theta}
    \bigr)
    = \sigma\ly{0}\bigl(
X\ly{0} \eval{\sy{W}\ly{0}}_{\theta} + \eval{\sy{B}\ly{0}}_{\theta}
\bigr)
$.
\textbf{Induction step}: Consider $\ell \geq 1$.
Assume that
$
    \eval{\sy\point\ly{\ell}}_{\theta}
    =
    \SliceDNN{\dnn}{0}{\ell}\big(
        \point\ly{0};
        \eval{\SliceDNN{\stheta}{0}{\ell}}_\theta
    \big)
$.
We will show
$
    \eval{\sy\point\ly{\ell+1}}_{\theta}
=
    \dnn\ly{\ell+1}\bigargs{
        \eval{\spoint\ly{\ell}}_{\theta};
        W\ly{\ell},
        \eval{\sy{B}\ly{\ell}}_{\theta}
    }
    = \sigma\ly{\ell}\bigl(
        \eval{\sy{X}\ly{\ell}}_\theta
        {W}\ly{\ell} +
        \eval{\sy{B}\ly{\ell}}_{\theta}
    \bigr)
$.
On~\pref{li:Conditional-Forward-Point-Symbolic-Pre-Other},
$
    \sy\point\ly{\ell+1}_\PreKW \eqdef \sy{X}\ly{\ell} {W}\ly{\ell} + \sy{B}\ly{\ell}
$.
On~\pref{li:Conditional-Forward-Point-Symbolic-Post},
$
    \sy\point\ly{\ell+1}, \varphi_\sigma\ly{\ell+1}
    \eqdef \condsy{{\sigma}}\ly{\ell}\withRef{
        X_{\RefKW}\ly{\ell+1}
    }\bigl( \sy{X}\ly{\ell+1}_\PreKW \bigr)
$.
Because $\theta \vDash \varphi$ implies $\theta \vDash \varphi_\sigma\ly{\ell+1}$,
by~\pref{def:Conditional-Activation-Function}-\ref{cond:Conditional-Activation-Function-C1}, we have
$
    \eval{
        \sy\point\ly{\ell+1}
    }_\theta
    =
    \sigma\ly{\ell}\bigl(
    \eval{
        \sy\point\ly{1+1}_\PreKW
    }_{\theta}
    \bigr)
    = \sigma\ly{\ell}\bigl(
        \eval{\sy{X}\ly{\ell}}_\theta
        {W}\ly{\ell} +
        \eval{\sy{B}\ly{\ell}}_{\theta}
    \bigr)
$.
Hence, by induction, we prove
$
    \eval{\sy\point\ly{\ell}}_{\theta}
    =
    \dnn^{\theta}\big(\point\ly{0}\big)
$.

\item[\ref{cond:Conditional-Forward-Pointwise-Linear}]
As seen in the proof for \pref{cond:Conditional-Forward-Pointwise-Correct},
for any layer $\ell$,
$
    \sy{X}\ly{\ell+1}_\PreKW
$ is a linear expression.
Because
$
    \sy\point\ly{\ell+1}, \varphi_\sigma\ly{\ell+1}
    \eqdef \condsy{{\sigma}}\ly{\ell}\withRef{
        X_{\RefKW}\ly{\ell+1}
    }\bigl( \sy{X}\ly{\ell+1}_\PreKW \bigr)
$~(\pref{li:Conditional-Forward-Point-Symbolic-Post}),
by~\pref{def:Conditional-Activation-Function}-\ref{cond:Conditional-Activation-Function-C2},
$\sy\point\ly{\ell+1}$ is a
linear expression and $\varphi_\sigma\ly{\ell+1}$ is a linear formula.
Hence, $\sy\point\ly{L}$ is a linear expression and
$\varphi\eqdef \bigwedge_{1\leq\ell\leq L}{\varphi_\sigma\ly{\ell}}$
is a linear formula.

\item[\ref{cond:Conditional-Forward-Pointwise-LocallyLinear}]
Let the polytope
$
    \polytope\ly{0} \eqdef
        \left\{
            \idx{0}\point\ly{0},
            \idx{1}\point\ly{0},
            \ldots,
            \idx{p-1}\point\ly{0}
        \right\}
$
and
$
    \idx{j}\sy\point\ly{L}, \idx{j}\varphi \eqdef
        \condsy\dnn
        \withRef{\point\ly{0}_\RefKW}
        \bigl(
            \idx{j}\point\ly{0}
        \bigr)
$ for $0 \leq j < p$.
Let $\theta \vDash \bigwedge_j \idx{j}\varphi$.
Let $\point\ly{0} \in \convexhull\bigl( \polytope\ly{0} \bigr)$.
To show that $\dnn^\theta$ is \locallylinear{} for
$
    \polytope\ly{0}
$,
we construct a linear function $f$ such that
$
    f\bigl( \point\ly{0} \bigr) =
    \dnn^\theta\bigl( \point\ly{0} \bigr)
$
~(\pref{def:locallylinear}).
We prove this by structural induction.
\textbf{Base case}:
For the first layer $\ly{\ell} = 0$.
As seen in the proof for \pref{cond:Conditional-Forward-Pointwise-Correct},
$\theta \vDash\!\!\idx{j}\varphi$
implies
$
    \eval{
        \idx{j}\sy\point\ly{1}
    }_\theta
    =
    \sigma\ly{0}\bigl(
        \idx{j}X\ly{0}
        \eval{\sy{W}\ly{0}}_{\theta} +
        \eval{\sy{B}\ly{0}}_{\theta}
    \bigr)
$.
Let
$
    P' \eqdef
    \bigcup_j \left\{
        \idx{j}X\ly{0}
        \eval{\sy{W}\ly{0}}_{\theta} +
        \eval{\sy{B}\ly{0}}_{\theta}
    \right\}
$.
By~\pref{def:Conditional-Activation-Function}-\ref{cond:Conditional-Activation-Function-C3},
we know there exists
a linear function $g$ such that
$
    g( X' ) = \sigma\ly{0}( X' )
$
for any
$
    X' \in \convexhull( P' )
$.
Hence, there exist a linear function
$
    f( X\ly{0} ) = g\bigl(
        X\ly{0}
        \eval{ \sy{W}\ly{0} }_\theta +
        \eval{ \sy{B}\ly{0} }_\theta
    \bigr)
$
such that
$
    f( X\ly{0} ) =
    \dnn\ly{0}\bigargs{
        X\ly{0}; \eval{\sy{W}\ly{0}}_{\theta}, \eval{\sy{B}\ly{0}}_{\theta}
    }
$
for any $\point\ly{0} \in \convexhull\bigl( \polytope\ly{0} \bigr)$.
\textbf{Induction step}:
Consider $\ell \geq 1$.
Assume that $
\eval{ \idx{j}\sy\point\ly{\ell} }_{\theta}
=
\SliceDNN{\dnn}{0}{\ell}\big(
    \idx{j}\point\ly{0};
    \eval{\SliceDNN{\stheta}{0}{\ell}}_\theta
\big)
$.
As seen in the proof for \pref{cond:Conditional-Forward-Pointwise-Correct},
$\theta \vDash\!\!\idx{j}\varphi$
implies
$
    \eval{
        \idx{j}\sy\point\ly{\ell+1}
    }_\theta
    =
    \sigma\ly{\ell}\bigl(
        \eval{\idx{j}\sy{X}\ly{\ell}}_{\theta}
        {W}\ly{\ell} +
        \eval{\sy{B}\ly{\ell}}_{\theta}
    \bigr)
$.
Assume a linear function $h$ where
$
    h( X\ly{0} ) =
    \SliceDNN{\dnn}{0}{\ell}\big(
        \point\ly{0};
        \eval{\SliceDNN{\stheta}{0}{\ell}}_\theta
    \big)
$
for any $\point\ly{0} \in \convexhull\bigl( \polytope\ly{0} \bigr)$.
Let
$
    P' \eqdef
    \bigcup_j \left\{
        \eval{ \idx{j}\sy{X}\ly{\ell} }_\theta
        {W}\ly{\ell} +
        \eval{ \sy{B}\ly{\ell} }_\theta
    \right\}
$.
By~\pref{def:Conditional-Activation-Function}-\ref{cond:Conditional-Activation-Function-C3}, we know there exists
a linear function $g$ such that
$
    g( X' ) = \sigma\ly{\ell}( X' )
$
for any
$
    X' \in \convexhull( P' )
$.
Hence, there exist a linear function
$
    f( X\ly{0} ) = g\bigl(
        h\bigl( X\ly{0} \bigr)
        \eval{ \sy{W}\ly{0} }_\theta +
        \eval{ \sy{B}\ly{0} }_\theta
    \bigr)
$
such that
$
    f( X\ly{0} ) =
    \SliceDNN{\dnn}{0}{\ell+1}\big(
        \point\ly{0};
        \eval{\SliceDNN{\stheta}{0}{\ell+1}}_\theta
    \big)
$
for any $\point\ly{0} \in \convexhull\bigl( \polytope\ly{0} \bigr)$.
Thus, by induction, we have a linear function $f$ such that
$
    f\bigl( \point\ly{0} \bigr) =
    \dnn^\theta\bigl( \point\ly{0} \bigr)
$
for any $\point\ly{0} \in \convexhull\bigl( \polytope\ly{0} \bigr)$.
Hence, $\dnn^\theta$ is \locallylinear{} for $\polytope\ly{0}$.
\qedhere
\end{cs}
\end{proof}
 
\begin{reptheorem}{thm:Conditional-Forward-Polytope}
\defProof{proof:Conditional-Forward-Polytope}
Let the polytope
$
\polytope\ly{0} \eqdef
    \left\{
        \idx{0}\point\ly{0},
        \idx{1}\point\ly{0},
        \ldots,
        \idx{p-1}\point\ly{0}
    \right\}
$.
Let
$
    \sy\polytope\ly{L}, \varphi
    \eqdef
        \condsy\dnn
\big(
                \polytope\ly{0}
            \big)
$
where
$
    \sy\polytope\ly{L} =
        \left\{
            \idx{0}\sy\point\ly{L},
            \idx{1}\sy\point\ly{L},
            \ldots,
            \idx{p-1}\sy\point\ly{L}
        \right\}
$.
Let $\theta \vDash \varphi$.
\begin{cs}
\item\labelInMainText{cond:Conditional-Forward-Polytope-C1}
$
        \dnn^{\theta}
        \bigl( \idx{j}\point\ly{0} \bigr)
        =
        \eval{ \idx{j}\sy\point\ly{L} }_{\theta}
    $ for $0 \leq j < p$.
\item\labelInMainText{cond:Conditional-Forward-Polytope-C2}
$\idx{j}\sy\point\ly{L} \in \sy\polytope\ly{L}$ is a linear expression and $\varphi$ is a linear formula for $0 \leq j < p$.
\item\labelInMainText{cond:Conditional-Forward-Polytope-C3}
$\theta \vDash \varphi$ implies that $\dnn^\theta$ is \locallylinear{} for
    $\polytope\ly{0}$.
\end{cs} \end{reptheorem}

\begin{proof}
Consider any concrete input vertex
$\idx{j}\point\ly{0} \in \polytope\ly{0}$.
\pref{li:Conditional-Forward-Polytope-Each}
calculates its corresponding symbolic output vertex with constraints
$
    \idx{j}\sy\point\ly{L}, \idx{j}\varphi_\sigma \eqdef
            \condsy\dnn\withRef{
\centroid\big( \polytope\ly{0} \big)
            }\big(
                \idx{j}\point\ly{0}
            \big)
$.

\begin{cs}[align=left,leftmargin=*,wide=0pt]
\item[\ref{cond:Conditional-Forward-Polytope-C1}]
Because $\theta \vDash \varphi$ implies $\theta \vDash \idx{j}\varphi_\sigma$.
$
    \dnn^{\theta}
    \bigl( \idx{j}\point\ly{0} \bigr)
    =
    \eval{ \idx{j}\sy\point\ly{L} }_{\theta}
$
using~\pref{thm:Conditional-Forward-Pointwise}-\ref{cond:Conditional-Forward-Pointwise-Correct}.

\item[\ref{cond:Conditional-Forward-Polytope-C2}]
By~\pref{thm:Conditional-Forward-Pointwise}-\ref{cond:Conditional-Forward-Pointwise-Linear},
$\idx{j}\sy\point\ly{L}$ is a linear expression
and $\idx{j}\varphi_\sigma$ is a linear formula.
Because $\varphi$ is the conjunction of all $\idx{j}\varphi_\sigma$,
$\varphi$ is a linear formula.

\item[\ref{cond:Conditional-Forward-Polytope-C3}]
On~\pref{li:Conditional-Forward-Polytope-Each},
$\idx{j}\point\ly{0}$ for $0 \leq j < p$
were conditionally forwarded on $\dnn$ with
the same reference point $\centroid\big( \polytope\ly{0} \big)$.
Hence, $\theta \vDash \varphi$ implies
$\dnn^\theta$ is \locallylinear{} for $\polytope\ly{0}$
using~\pref{thm:Conditional-Forward-Pointwise}-\ref{cond:Conditional-Forward-Pointwise-LocallyLinear}.
\qedhere
\end{cs}
\end{proof} 
\begin{reptheorem}{thm:Shift-And-Assert}
\defProof{proof:Shift-And-Assert}
Let
$
    \dnn_\texttt{ret}
    \eqdef
    \foo\bigl(
        \dnn,
        \dnn^\OriginalKW,
        \polytopeSet\ly{0},
        \spec,
        k
    \bigr)
$, $\dnn_\texttt{ret} \neq \bot$
and $\SliceDNN{\dnn}{0}{k}$ is \locallylinear{} for any polytope
$\polytope\ly{0} \in \polytopeSet\ly{0}$.
Then:
\begin{cs}
\item\labelInMainText{cond:Shift-And-Assert-C1}
$\dnn_\texttt{ret}$ is \locallylinear{}
    for any polytope $\polytope\ly{0} \in \polytopeSet\ly{0}$.
\item\labelInMainText{cond:Shift-And-Assert-C2}
$\dnn_\texttt{ret}$ satisfies the polytope specification
    $\bigl(
        \polytopeSet\ly{0}, \spec
    \bigr)$.
\item\labelInMainText{cond:Shift-And-Assert-C3}
$\foo$ runs in polynomial time in the total number of vertices
in $\polytopeSet\ly{0}$ and the size of $\dnn$.
\end{cs}
 \end{reptheorem}

\begin{proof}

We prove each condition of \pref{thm:Shift-And-Assert}.

\begin{cs}[align=left,leftmargin=*,wide=0pt]

\item[\ref{cond:Shift-And-Assert-C1}]
We will show that after the update
on~\pref{li:ShiftAndRepair-Update},
$\SliceDNN{\dnn_\texttt{ret}}{k}{L}$ is \locallylinear{} for $\polytope\ly{k}$.
Hence,
$\SliceDNN{\dnn_\texttt{ret}}{0}{L}$ is \locallylinear{} for $\polytope\ly{0}$.
\pref{li:ShiftAndRepair-ForEachPolytope-Call}
calculates symbolic output polytopes and constraints
$\sy\polytope\ly{L}, \varphi_\sigma
    \gets \SliceDNN{\condsy\dnn}{k}{L}
        \big(
            \polytope\ly{k}
        \big)
$
by conditionally forward executing
$\polytope\ly{k} \in \SliceDNN{\dnn}{0}{k}\bigargs{\polytopeSet\ly{0}}$
on $\SliceDNN{\condsy\dnn}{k}{L}$.
Because
\pref{li:ShiftAndRepair-ForEachPolytope-Add}
adds each $\varphi_\sigma$ to $\varphi$
and
\pref{li:ShiftAndRepair-Minimize}
solves $\varphi$,
if $\dnn_\texttt{ret}$ is not $\bot$,
the feasible parameters $\theta'_{\SliceDNN{\dnn}{k}{L}}$
satisfies $\varphi$.
Hence, $\theta'_{\SliceDNN{\dnn}{k}{L}} \vDash \varphi_\sigma$
and $\SliceDNN{\dnn_\texttt{ret}}{k}{L}$ is \locallylinear{}
for $\polytope\ly{k}$ using~\pref{thm:Conditional-Forward-Polytope}-\ref{cond:Conditional-Forward-Polytope-C3}.
Because $\SliceDNN{\dnn_\texttt{ret}}{0}{k}$ is not modified,
On~\pref{li:ShiftAndRepair-Return},
$\SliceDNN{\dnn_\texttt{ret}}{0}{k}$ is still \locallylinear{} for $\polytope\ly{0}$.
Hence, on~\pref{li:ShiftAndRepair-Return},
$\dnn'\eqdef\dnn_\texttt{ret}$ is \locallylinear{} for $\polytope\ly{0}$.

\item[\ref{cond:Shift-And-Assert-C2}]
Consider any input polytope $\polytope\ly{0} \in \polytopeSet\ly{0}$
and its corresponding specification $\psi \in \spec$.
Because ~\pref{li:ShiftAndRepair-Spec} adds $
\psi\bigl(
    \idx{j}\spoint\ly{L}
\bigr)
$
for any vertex
$
    \idx{j}\spoint\ly{L}
    \in
    \sy\polytope\ly{L}
$
to $\varphi$, we have
$
\theta'_{\SliceDNN{\dnn}{k}{L}} \vDash
    \bigwedge_{j}{
        \psi\bigl(
            \idx{j}\spoint\ly{L}
        \bigr)
    }
$.
Because
$\dnn' \neq \bot$ implies
$\theta'_{\SliceDNN{\dnn}{k}{L}} \neq \bot$
we have
$
    \dnn'
    \bigl( \idx{j}\point\ly{0} \bigr)
    =
    \eval{ \idx{j}\sy\point\ly{L} }_{\theta'_{\SliceDNN{\dnn}{k}{L}}}
$
using~\pref{thm:Conditional-Forward-Polytope}-\ref{cond:Conditional-Forward-Polytope-C1}.
Hence, we have
$
\theta \vDash
    \bigwedge_{j}{
    \psi\bigl(
        \dnn'\bigl( \idx{j}\point\ly{0} \bigr)
    \bigr)
    }
$.
As seen in the proof for \pref{cond:Shift-And-Assert-C1},
$\dnn'$ is \locallylinear{} for $\polytope\ly{0}$.
Because
$
    \psi\bigl(
        \dnn'\bigl( \idx{j}\point\ly{0} \bigr)
    \bigr)
$
is a linear formula using~\pref{def:Preliminaries-PolytopeSpecification},
we have
$
\bigwedge_{j}{
    \psi\bigl(
        \dnn'\bigl( \idx{j}\point\ly{0} \bigr)
    \bigr)
    }
$
for all vertices $
    \idx{j}\point\ly{0}
    \in
    \polytope\ly{0}
$
implies that
$
    \psi\bigl(
        \dnn'\bigl( \point\ly{0} \bigr)
    \bigr)
$
for any
$
    \point\ly{0}
    \in \convexhull\bigl(
        \polytope\ly{0}
    \bigr)
$.
Hence,
$\dnn'$ satisfies the V-polytope specification
$
    ( \polytopeSet\ly{0}, \spec )
$.

\item[\ref{cond:Shift-And-Assert-C3}]
Because $\varphi$ is a linear formula and its size is polynomial in the total
number of vertices in $\polytopeSet\ly{0}$ and size of the $\dnn$,
$\foo$ runs in polynomial time in the total number of vertices in
$\polytopeSet\ly{0}$ and size of the $\dnn$.
\qedhere
\end{cs}
\end{proof}
  }{}

\end{document}